\documentclass[11pt]{article}
\usepackage[preprint]{acl}
\usepackage{times}
\usepackage{latexsym}
\usepackage[T1]{fontenc}
\usepackage[utf8]{inputenc}
\usepackage{microtype}
\usepackage{graphicx}
\usepackage{booktabs}
\usepackage{array}
\usepackage{amsmath}
\usepackage{amssymb}
\usepackage{amsthm}
\newtheorem{proposition}{Proposition}
\usepackage{xcolor}
\usepackage{placeins}
\usepackage{needspace}
\usepackage{tikz}
\usetikzlibrary{arrows.meta,positioning,fit,backgrounds,patterns}
\usepackage{pifont}

\newcommand{\Qs}{$Q^*$}

\title{Where Should a KV-Cache Repair Budget Go?\\Distractor Structure and Contextual Compatibility in Non-Prefix Reuse}

\author{Yuxing Peng\textsuperscript{1}, Zhiqing Tang\textsuperscript{2}\thanks{Corresponding author: \texttt{zhiqingtang@bnu.edu.cn}.}, Weijia Jia\textsuperscript{2,3} \\
  \textsuperscript{1}Faculty of Arts and Sciences, Beijing Normal University, Zhuhai 519087, China \\
  \textsuperscript{2}Institute of Artificial Intelligence and Future Networks, Faculty of Arts and Sciences, \\
  Beijing Normal University, Zhuhai 519087, China \\
  \textsuperscript{3}Guangdong Key Lab of AI and Multi-Modal Data Processing, \\
  Beijing Normal-Hong Kong Baptist University, Zhuhai 519087, China \\
  \texttt{pyx26@mail.bnu.edu.cn}, \texttt{zhiqingtang@bnu.edu.cn}, \texttt{jiawj@bnu.edu.cn}}

\begin{document}
\maketitle
\begin{abstract}
Non-prefix KV-cache reuse encodes retrieved chunks independently and concatenates their caches. To find where a small repair budget should be spent, we vary the chunks around the answer-bearing one (the target) across three 7B--8B models and four datasets of dialogue memory, open-domain QA and API documentation. Reuse loses 19--100\% of the context-dependent questions that full prefill answers, and it loses more when the other chunks are of the same kind as the target. Because the recoverable damage is spread over these other chunks, restoring a few of them recovers little, and the budget is better spent on the target. This works through contextual compatibility: recomputing the target against the reused cache beats inserting its full-prefill KV in all seven settings tested, although the recomputed KV stays far from the full-prefill KV. Question attention over reused caches, however, selects the target in at most 18.8\% of development questions. A retriever finds it more often, and recomputing its top chunk after moving it next to the question significantly beats InfoFlow-style reordering at the same 5\% budget in six of nine open-domain QA and API-documentation settings. On dialogue memory, though, recomputing chunks never significantly beats the token-level selector chosen for each setting.
\end{abstract}

\section{Introduction}
\label{sec:intro}

\begin{figure*}[t]
\centering
\definecolor{ink}{HTML}{1F2937}%
\definecolor{ink2}{HTML}{4B5563}%
\definecolor{rule}{HTML}{D1D5DB}%
\definecolor{bg}{HTML}{F8FAFC}%
\definecolor{tgt}{HTML}{047857}%
\definecolor{tgtL}{HTML}{D1FAE5}%
\definecolor{alk}{HTML}{C2410C}%
\definecolor{alkL}{HTML}{FFEDD5}%
\definecolor{qcol}{HTML}{1D4ED8}%
\definecolor{qL}{HTML}{DBEAFE}%
\definecolor{lost}{HTML}{DC2626}%
\definecolor{lostL}{HTML}{FEE2E2}%
\begin{tikzpicture}[x=1cm, y=1cm, >=Latex, font=\sffamily, every node/.append style={execute at begin node={\hyphenpenalty=10000\exhyphenpenalty=10000\relax}}]
\useasboundingbox (0,0) rectangle (15.88,4.78);

\draw[rounded corners=3pt, fill=bg, draw=rule, line width=0.6pt] (0,0) rectangle (5.50,4.78);
\node[anchor=north west, inner sep=0pt, font=\sffamily\footnotesize\bfseries, text=ink] at (0.16,4.62) {(a) What reuse loses};

\node[anchor=west, inner sep=0pt, font=\sffamily\scriptsize\bfseries, text=ink] at (0.16,4.10) {Full prefill};
\node[anchor=east, inner sep=0pt, font=\sffamily\scriptsize, text=tgt] at (5.34,4.10) {cross-chunk context intact};

\draw[rounded corners=1.5pt, fill=white, draw=ink2, line width=0.5pt] (0.16,3.12) rectangle (0.80,3.66);
\node[inner sep=0pt, font=\sffamily\fontsize{6}{7}\selectfont, text=ink] at (0.48,3.39) {Prefix};

\draw[rounded corners=1.5pt, fill=white, draw=ink2, line width=0.5pt] (0.88,3.12) rectangle (1.52,3.66);
\node[inner sep=0pt, font=\sffamily\scriptsize, text=ink2] at (1.20,3.39) {1--5};

\draw[rounded corners=1.5pt, fill=tgtL, draw=tgt, line width=0.7pt] (1.60,3.12) rectangle (2.44,3.66);
\node[inner sep=0pt, font=\sffamily\scriptsize\bfseries, text=tgt] at (2.02,3.39) {Doc 6};

\draw[rounded corners=1.5pt, fill=white, draw=ink2, line width=0.5pt] (2.52,3.12) rectangle (3.16,3.66);
\node[inner sep=0pt, font=\sffamily\scriptsize, text=ink2] at (2.84,3.39) {7--10};

\draw[rounded corners=1.5pt, fill=alkL, draw=alk, line width=0.7pt] (3.24,3.12) rectangle (4.12,3.66);
\node[inner sep=0pt, font=\sffamily\scriptsize\bfseries, text=alk] at (3.68,3.39) {Doc 11};

\draw[rounded corners=1.5pt, fill=white, draw=ink2, line width=0.5pt] (4.20,3.12) rectangle (4.90,3.66);
\node[inner sep=0pt, font=\sffamily\scriptsize, text=ink2] at (4.55,3.39) {12--30};

\draw[rounded corners=1.5pt, fill=qL, draw=qcol, line width=0.7pt] (4.98,3.12) rectangle (5.34,3.66);
\node[inner sep=0pt, font=\sffamily\scriptsize\bfseries, text=qcol] at (5.16,3.39) {Q};

\draw[->, tgt, line width=0.65pt] (0.48,3.68) to[out=35, in=145] (1.20,3.68);
\draw[->, tgt, line width=0.65pt] (1.20,3.68) to[out=35, in=145] (2.02,3.68);
\draw[->, tgt, line width=0.65pt] (2.02,3.68) to[out=35, in=145] (2.84,3.68);
\draw[->, tgt, line width=0.65pt] (2.84,3.68) to[out=35, in=145] (3.68,3.68);
\draw[->, tgt, line width=0.65pt] (3.68,3.68) to[out=35, in=145] (4.55,3.68);
\draw[->, qcol, line width=0.65pt] (4.55,3.68) to[out=35, in=145] (5.16,3.68);

\node[anchor=west, inner sep=0pt, font=\sffamily\scriptsize, text=ink2] at (0.16,2.85) {Each chunk attends to all earlier chunks};

\draw[rule, line width=0.5pt] (0.16,2.58) -- (5.34,2.58);

\node[anchor=west, inner sep=0pt, font=\sffamily\scriptsize\bfseries, text=lost] at (0.16,2.30) {Non-prefix KV reuse};
\node[anchor=east, inner sep=0pt, font=\sffamily\scriptsize\bfseries, text=lost] at (5.34,2.30) {loses 19--100\% of \Qs};

\draw[rounded corners=1.5pt, fill=white, draw=ink2, line width=0.5pt] (0.16,1.24) rectangle (0.80,1.78);
\node[inner sep=0pt, font=\sffamily\fontsize{6}{7}\selectfont, text=ink] at (0.48,1.51) {Prefix};

\draw[rounded corners=1.5pt, fill=white, draw=ink2, line width=0.5pt, densely dashed] (0.88,1.24) rectangle (1.52,1.78);
\node[inner sep=0pt, font=\sffamily\scriptsize, text=ink2] at (1.20,1.51) {1--5};

\draw[rounded corners=1.5pt, fill=tgtL, draw=tgt, line width=0.7pt, densely dashed] (1.60,1.24) rectangle (2.44,1.78);
\node[inner sep=0pt, font=\sffamily\scriptsize\bfseries, text=tgt] at (2.02,1.51) {Doc 6};

\draw[rounded corners=1.5pt, fill=white, draw=ink2, line width=0.5pt, densely dashed] (2.52,1.24) rectangle (3.16,1.78);
\node[inner sep=0pt, font=\sffamily\scriptsize, text=ink2] at (2.84,1.51) {7--10};

\draw[rounded corners=1.5pt, fill=alkL, draw=alk, line width=0.7pt, densely dashed] (3.24,1.24) rectangle (4.12,1.78);
\node[inner sep=0pt, font=\sffamily\scriptsize\bfseries, text=alk] at (3.68,1.51) {Doc 11};

\draw[rounded corners=1.5pt, fill=white, draw=ink2, line width=0.5pt, densely dashed] (4.20,1.24) rectangle (4.90,1.78);
\node[inner sep=0pt, font=\sffamily\scriptsize, text=ink2] at (4.55,1.51) {12--30};

\draw[rounded corners=1.5pt, fill=qL, draw=qcol, line width=0.7pt] (4.98,1.24) rectangle (5.34,1.78);
\node[inner sep=0pt, font=\sffamily\scriptsize\bfseries, text=qcol] at (5.16,1.51) {Q};

\draw[lost, line width=0.6pt, densely dashed] (1.20,1.80) to[out=40, in=140] node[pos=0.5, fill=bg, inner sep=0.2pt, font=\sffamily\scriptsize, text=lost]{\ding{55}} (2.02,1.80);
\draw[lost, line width=0.6pt, densely dashed] (2.02,1.80) to[out=40, in=140] node[pos=0.5, fill=bg, inner sep=0.2pt, font=\sffamily\scriptsize, text=lost]{\ding{55}} (2.84,1.80);
\draw[lost, line width=0.6pt, densely dashed] (2.84,1.80) to[out=40, in=140] node[pos=0.5, fill=bg, inner sep=0.2pt, font=\sffamily\scriptsize, text=lost]{\ding{55}} (3.68,1.80);
\draw[lost, line width=0.6pt, densely dashed] (3.68,1.80) to[out=40, in=140] node[pos=0.5, fill=bg, inner sep=0.2pt, font=\sffamily\scriptsize, text=lost]{\ding{55}} (4.55,1.80);

\draw[->, qcol, line width=0.65pt] (2.84,1.22) to[out=-20, in=-160] (5.16,1.22);
\node[anchor=west, inner sep=0pt, font=\sffamily\scriptsize, text=lost] at (0.16,0.68) {\ding{55} Chunks encoded alone, each after the prefix};
\node[anchor=west, inner sep=0pt, font=\sffamily\scriptsize, text=qcol] at (0.16,0.28) {$\rightarrow$ Only \textbf{Q} and the answer read all caches};

\draw[rounded corners=3pt, fill=bg, draw=rule, line width=0.6pt] (5.64,0) rectangle (10.30,4.78);
\node[anchor=north west, inner sep=0pt, font=\sffamily\footnotesize\bfseries, text=ink] at (5.80,4.62) {(b) A similar chunk misleads};

\fill[qcol] (5.80,3.52) rectangle (5.87,4.22);
\node[anchor=north west, inner sep=0pt, text width=4.16cm, font=\sffamily\scriptsize, text=ink] at (5.96,4.20) {\textcolor{qcol}{\textbf{Question}} (NQ-Open, Qwen3-8B)\\Who played Daisy Duke in the new Dukes of Hazzard?};

\fill[tgt] (5.80,2.50) rectangle (5.87,3.36);
\node[anchor=north west, inner sep=0pt, text width=4.16cm, font=\sffamily\scriptsize, text=ink] at (5.96,3.34) {\textcolor{tgt}{\textbf{Doc 6 (target)}}\\\emph{Dukes of Hazzard (film):} \ldots{} Daisy Duke (\textbf{Jessica Simpson})};

\fill[alk] (5.80,1.44) rectangle (5.87,2.34);
\node[anchor=north west, inner sep=0pt, text width=4.16cm, font=\sffamily\scriptsize, text=ink] at (5.96,2.32) {\textcolor{alk}{\textbf{Doc 11 (similar passage)}}\\\textbf{Catherine Bach} \ldots{} known for playing Daisy Duke (TV series)};

\draw[rule, line width=0.5pt] (5.80,1.22) -- (10.14,1.22);

\draw[rounded corners=2pt, fill=tgtL, draw=tgt, line width=0.5pt] (5.80,0.68) rectangle (10.14,1.10);
\node[anchor=west, inner sep=0pt, font=\sffamily\scriptsize, text=ink] at (5.90,0.89) {Full prefill $\rightarrow$ \textbf{Jessica Simpson}};
\node[anchor=east, inner sep=0pt, font=\sffamily\scriptsize\bfseries, text=tgt] at (10.04,0.89) {\ding{51}};

\draw[rounded corners=2pt, fill=lostL, draw=lost, line width=0.5pt] (5.80,0.14) rectangle (10.14,0.56);
\node[anchor=west, inner sep=0pt, font=\sffamily\scriptsize, text=ink] at (5.90,0.35) {KV reuse $\rightarrow$ \textbf{Catherine Bach}};
\node[anchor=east, inner sep=0pt, font=\sffamily\scriptsize\bfseries, text=lost] at (10.04,0.35) {\ding{55}};

\draw[rounded corners=3pt, fill=bg, draw=rule, line width=0.6pt] (10.44,0) rectangle (15.88,4.78);
\node[anchor=north west, inner sep=0pt, font=\sffamily\footnotesize\bfseries, text=ink] at (10.60,4.62) {(c) Where the budget goes};

\draw[rounded corners=2pt, fill=white, draw=lost!70, line width=0.6pt] (10.60,3.15) rectangle (15.72,4.22);
\node[anchor=north west, inner sep=0pt, text width=4.90cm, font=\sffamily\scriptsize, text=ink2] at (10.72,4.11) {\textcolor{lost}{\textbf{\ding{55} Restore a few distractors}}\\[1pt]Damage spread over all; restoring a single one changes retention little};

\draw[rounded corners=2pt, fill=white, draw=alk!80, line width=0.6pt] (10.60,2.31) rectangle (15.72,3.07);
\node[anchor=north west, inner sep=0pt, text width=4.90cm, font=\sffamily\scriptsize, text=ink2] at (10.72,2.96) {\textcolor{alk}{\textbf{\ding{55} Choose by question attention}}\\[1pt]Finds the target in $\le$18.8\%};

\draw[rounded corners=2pt, fill=tgtL!45, draw=tgt, line width=0.9pt] (10.60,0.14) rectangle (15.72,2.22);
\node[anchor=north west, inner sep=0pt, text width=4.90cm, font=\sffamily\scriptsize, text=ink] at (10.72,2.11) {\textcolor{tgt}{\textbf{\ding{51} Recompute the target with reused KV}}\\[1pt]Beats inserting its full-prefill KV in all seven settings};

\draw[rounded corners=1.2pt, fill=white, draw=ink2, line width=0.5pt, densely dashed] (10.74,0.24) rectangle (12.34,0.68);
\node[inner sep=0pt, font=\sffamily\scriptsize, text=ink2] at (11.54,0.46) {Reused KV};

\draw[rounded corners=1.2pt, fill=tgtL, draw=tgt, line width=0.7pt] (12.50,0.24) rectangle (14.60,0.68);
\node[inner sep=0pt, font=\sffamily\scriptsize\bfseries, text=tgt] at (13.55,0.46) {Retrieved top-1};

\draw[rounded corners=1.2pt, fill=qL, draw=qcol, line width=0.7pt] (14.76,0.24) rectangle (15.58,0.68);
\node[inner sep=0pt, font=\sffamily\scriptsize\bfseries, text=qcol] at (15.17,0.46) {Q};

\draw[->, tgt, line width=0.65pt] (11.90,0.70) to[out=20, in=160] node[pos=0.52, above=1pt, inner sep=0pt, font=\sffamily\scriptsize\bfseries, text=tgt]{move last, recompute} (13.55,0.70);
\end{tikzpicture}%
\caption{\textbf{Reuse cuts each chunk off from the chunks before it, and similar chunks then mislead the model.} (a) Full prefill encodes each chunk after the earlier ones, whereas non-prefix reuse encodes each chunk alone and joins the caches, so only the question and the answer attend across chunks. (b) An NQ-Open question answered by Qwen3-8B with 30 passages (actual outputs and passages, both shortened). (c) Where a small repair budget goes. The inset shows that, when the target is unknown, the retriever's top chunk is moved last (Sections~\ref{sec:anatomy}--\ref{sec:repair}).}
\label{fig:schematic}
\end{figure*}
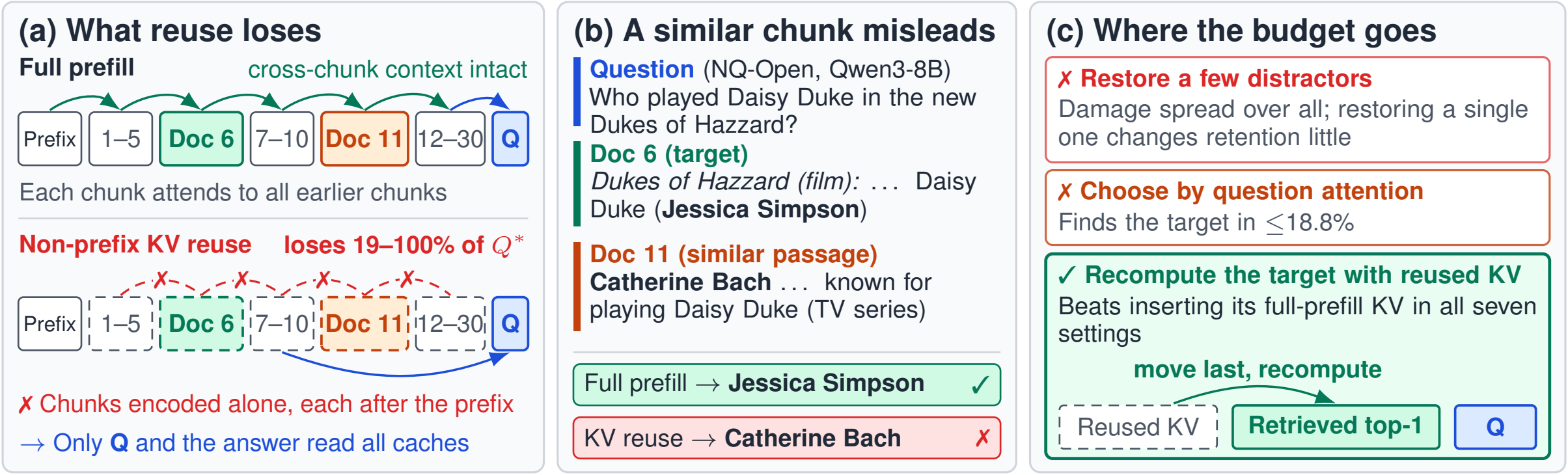

Retrieval-augmented generation, long-term conversational memory and tool-use agents build each prompt from retrieved units such as passages, dialogue sessions or API documents. Because the retrieved combination changes from request to request, prefix caching rarely applies. Non-prefix KV-cache reuse instead encodes each chunk once, stores its keys and values, and concatenates the stored caches at query time after assigning or correcting positions \citep{promptcache,turborag,blockattn,kvlink,ape}. This saves the prefill over the retrieved context, but no chunk is ever encoded in the presence of the others (Figure~\ref{fig:schematic}). Repair methods either select a small fraction of tokens and recompute them online \citep{cacheblend,epic,prophetkv,infoflowkv}, which we call token-level selectors, or encode chunks after a fixed surrogate prefix (SurroKV; \citealp{surrokv}). Most of them choose scattered tokens by KV deviation, by position or by the question's attention over the reused cache, and most use full prefill either as the reference that a repair should approach or to explain what a repair restores (Appendix Table~\ref{tab:assume}).

To find out where a small repair budget should be spent, we trace which parts of a retrieved context account for the accuracy that reuse loses. A retriever returns the chunks most similar to the query, and these chunks are often similar to one another: several passages about the same entity, sessions from the same conversation, or documentation for APIs of the same kind. In Figure~\ref{fig:schematic}, Qwen3-8B answers correctly from the full-prefill cache, but when the same passages are reused it names the actress from a similar passage. For each question we fix the answer-bearing chunk (the target) and its position, and we draw the other chunks (the distractors) from off-domain prose, from unrelated chunks of the same corpus, or from retrieved chunks similar to the target (Section~\ref{sec:setup}). We evaluate Qwen3-8B, Llama-3.1-8B-Instruct and Mistral-7B-Instruct-v0.2 on dialogue memory (LoCoMo), open-domain QA (NQ-Open and TriviaQA) and API documentation (Gorilla), and we call each model--dataset pair a setting. In each setting we use the questions that full prefill answers and that require the retrieved context (\Qs{}).

\paragraph{The loss grows as the other chunks become more like the target.} Position-corrected reuse loses 19--100\% of these questions. Part of the loss remains even when the distractors are off-domain prose. Unrelated chunks of the same kind add to the loss significantly in nine of twelve settings, and on open-domain QA, passages similar to the target reduce accuracy further in all six settings, significantly in five (Section~\ref{sec:anatomy}). A model that \citet{blockattn} fine-tune to encode documents as independent blocks still loses 60\% of these questions on NQ-Open, and most of this loss comes from the similar chunks (Appendix~\ref{app:blockft}).

\paragraph{The damage is spread over the distractors, and a small budget does not undo it.} Restoring the full-prefill caches of all distractors recovers more than restoring the target's cache, in all twelve settings and significantly in eleven. By contrast, restoring a single distractor, the most similar or a random one in each setting, changes retention significantly in only two of these 24 cases. The effects of the distractors add up. In a simple model, each reused distractor adds a weight to the odds of losing the answer. Calibrated only at its two measured end points, with no distractor restored and with every distractor restored, this model predicts how retention grows as random distractors are restored, and its prediction beats a linear one in five of the six open-domain QA settings (Section~\ref{sec:replace}). Restoring the distractors' caches needs the full prefill that reuse avoids, and a small budget spent on a few similar chunks does worse than the same budget spent on selected tokens (Section~\ref{sec:alloc}).

\paragraph{Spending the budget on the target works through contextual compatibility.} When the question is processed, it attends to the target together with the reused KV of the other chunks. Recomputing the target against the reused cache encodes it in that reused context. This recomputation beats inserting the target's full-prefill KV in all seven settings tested (four after Holm's correction), although the recomputed KV stays at a median relative distance of 0.23--0.30 from the full-prefill KV. Recomputing only the known target chunk, which makes up a median of 2.5--3.3\% of the context, beats the token-level selector chosen for each setting with a 5\% budget in all nine open-domain QA and API-documentation settings, by 0.10 to 0.49.

\paragraph{Question attention rarely finds the target; a retriever finds it more often.} In deployment the target is unknown. Over reused caches, question attention selects the target in at most 18.8\% of development questions, and the question's summed attention selects the last chunk before the question in 58--100\% of them (Section~\ref{sec:attention}). Because a recomputed chunk attends only to the chunks before it, we move the retriever's top-ranked chunk to the position just before the question and recompute it there. At the same 5\% budget, this significantly beats InfoFlow-style reordering \citep{infoflowkv} in six of the nine open-domain QA and API-documentation settings (five after Holm's correction). On the same reordered prompt, recomputing the moved chunk is never significantly worse than recomputing selected tokens, and it is significantly better in four settings (Section~\ref{sec:alloc}).
These gains are bounded by retrieval. On questions where the retriever's top chunk is not the target, recomputing the moved chunk does worse than recomputing selected tokens on the same prompt. Moving the top three chunks extends the gain to some of these questions, and with a 10\% budget its gain over moving only the top chunk matches a prediction from the retriever's hit rate in seven of the nine settings. With this budget, however, selected tokens recomputed on the same prompt remain competitive with the three chunks, which are significantly better in only three settings and significantly worse in two. On dialogue memory, chunk-level recomputation brings the 7B--8B models no significant gain over the token-level selector chosen for each setting.

Qwen3-14B and Qwen3-32B each reproduce all five results that we test for replication (Section~\ref{sec:repair} and Table~\ref{tab:larger_main}).

\section{Setup}
\label{sec:setup}

\paragraph{Datasets.} \looseness=-1 We use four datasets in which the answer is contained in one retrieved unit (the target chunk); Appendix~\ref{app:data} describes how they are built and gives the prompts.
\begin{itemize}\setlength{\itemsep}{1pt}
\item \looseness=-1 \textbf{LoCoMo} (dialogue memory; \citealp{locomo}): each chunk is one conversation session, with 13--16 sessions per prompt (about 12K tokens), and we use the questions whose evidence lies in a single session.
\item \textbf{NQ-Open} (open-domain QA; \citealp{nq,lostmiddle}): the multi-document version of \citet{lostmiddle}, with one gold passage among up to 30 Wikipedia passages (a median of about 4.6K tokens), scored by best-subspan exact match.
\item \textbf{TriviaQA} (open-domain QA; \citealp{triviaqa}): one gold passage together with 29 of the hard negatives released with DPR \citep{dpr}, also scored by best-subspan exact match.
\item \looseness=-1 \textbf{Gorilla} (API documentation; \citealp{gorilla}): HuggingFace APIBench requests, each with 31 API documents (a median of about 6.5K tokens), one of which documents the requested API.
\end{itemize}
Each dataset has 70 development questions and a disjoint test set (Appendix~\ref{app:data}). We evaluate Qwen3-8B \citep{qwen3} with thinking disabled, Llama-3.1-8B-Instruct \citep{llama3} and Mistral-7B-Instruct-v0.2 \citep{mistral} in bfloat16 with greedy decoding.

\paragraph{Distractor sources.} For each question, the target chunk and its position in the prompt are fixed. The other chunks come from one of three sources.
\begin{itemize}\setlength{\itemsep}{1pt}
\item \textbf{SIM}, retrieved chunks similar to the target: other sessions of the same conversation (LoCoMo), the retrieved hard negatives of the original datasets (NQ-Open, TriviaQA) and, for Gorilla, the API documents closest to the request and its reference solution under BGE-M3 \citep{bgem3}. A distractor list built from the request alone (SIM-q) gives the same near-zero retention under reuse (Appendix~\ref{app:data}), and the retriever used in Sections~\ref{sec:attention} and~\ref{sec:repair} is likewise given only the request.
\item \textbf{RAND}, unrelated chunks of the same kind: sessions of other conversations, passages retrieved for other questions, and API documents of other task categories.
\item \textbf{OTHER}, off-domain prose from Thoreau's \emph{Walden}, cut into pieces of the same lengths as the chunks they replace.
\end{itemize}

\paragraph{Conditions.} \looseness=-1 Full prefill encodes the whole prompt in one pass. Position-corrected reuse encodes every chunk alone after the shared instruction prefix, concatenates the caches with corrected rotary positions, and then appends the question. The no-context baseline omits all chunks. We write $C_0$ for the prefix of $P$ tokens, $C_1,\dots,C_m$ for the chunks in prompt order, $s_i$ for the first position of $C_i$ and $C_{<i}$ for $C_1\cup\dots\cup C_{i-1}$. In full prefill, a token of $C_i$ attends to $C_0$, $C_{<i}$ and the earlier tokens of $C_i$. Under reuse, it attends only to $C_0$ and the earlier tokens of $C_i$. The chunk is encoded starting at position $P$, and its keys are then rotated by $s_i-P$.

\begin{figure*}[t]
\centering
\includegraphics[width=\textwidth]{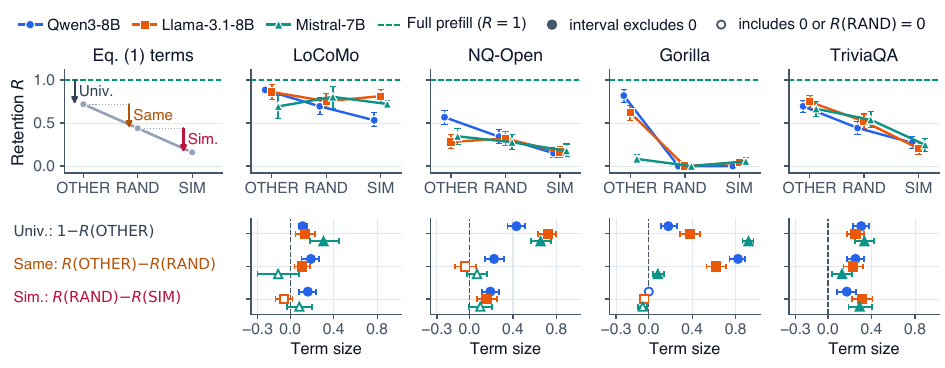}
\caption{Retention of position-corrected reuse on \Qs{} under each distractor source (top), and the three terms of Equation~\ref{eq:decomp} (bottom): universal $1-R(\mathrm{OTHER})$, same-category $R(\mathrm{OTHER})-R(\mathrm{RAND})$ and similarity $R(\mathrm{RAND})-R(\mathrm{SIM})$. Bars are 95\% paired bootstrap intervals, with questions resampled (conversations for LoCoMo). In the bottom row, open markers show two cases: intervals that include zero, and the similarity terms on Gorilla, which cannot be positive because retention under RAND is zero. Appendix Table~\ref{tab:main_decomposition} gives the values.}
\label{fig:retention}
\end{figure*}

\paragraph{Evaluation set and decomposition.} Following the meaningful subset of \citet{cestola2026}, we evaluate each setting on its context-dependent subset \Qs{}: the questions that the no-context baseline answers incorrectly and that full prefill answers correctly under SIM, RAND and OTHER (Appendix~\ref{app:impl} lists how our subset differs from theirs). Full-prefill accuracy on \Qs{} is therefore 1 by construction, and the accuracy of position-corrected reuse on \Qs{} is its retention $R$. The loss under SIM splits into three terms:
\begin{equation}
\label{eq:decomp}
\begin{split}
1-R(\mathrm{SIM}) ={}& \underbrace{1-R(\mathrm{OTHER})}_{\text{universal}}\\
&+ \underbrace{R(\mathrm{OTHER})-R(\mathrm{RAND})}_{\text{same-category}}\\
&+ \underbrace{R(\mathrm{RAND})-R(\mathrm{SIM})}_{\text{similarity}} .
\end{split}
\end{equation}
The universal term is the loss that remains when the distractors are off-domain prose. The same-category term is the extra loss from chunks of the same kind, and the similarity term is the further change when those chunks are also similar to the target. The terms are differences between ordered interventions, and they need not correspond to independent mechanisms, because the three sources can differ in more than one property. Intervals are 95\% paired bootstrap intervals with 2,000 resamples of questions (of conversations for LoCoMo), and we call a difference significant when its interval excludes zero.

\section{Anatomy of the Reuse Loss}
\label{sec:anatomy}

\subsection{Size and decomposition}
\label{sec:decomposition}

The loss of position-corrected reuse under SIM ranges from 0.19 to 1.00, and its interval excludes zero in all twelve settings (Figure~\ref{fig:retention} and Appendix Table~\ref{tab:main_decomposition}; Appendix Table~\ref{tab:allq} reports accuracy on all evaluated questions). The universal term, the loss that remains when every distractor is off-domain prose, ranges from 0.11 to 0.92.

Unrelated chunks of the same kind usually cost more than off-domain prose. The same-category term is significantly positive in nine of the twelve settings and never significantly negative. It is largest for Qwen3-8B and Llama-3.1-8B on Gorilla (0.82 and 0.62). There, retention under RAND is 0.00 for every model, while retention under OTHER reaches 0.82.

Whether chunks similar to the target add to the loss depends on the domain. On open-domain QA, the similarity term is positive in all six settings and significant in five. In dialogue memory, only Qwen3-8B has a significant similarity term (+0.16, [+0.08, +0.24]).

\paragraph{Competing answers.} On open-domain QA, the wrong answers point to the similar passages. Under SIM, the answer produced with reuse names an entity (a capitalised name or a number) that occurs in some distractor passage but not in the target passage on 40--63\% of the \Qs{} questions, compared with 8--40\% under RAND. The paired difference is significant in all six settings (Appendix Table~\ref{tab:competing}; Appendix~\ref{app:results} gives abstentions, examples and the corresponding counts of distractor APIs on Gorilla). These counts suggest that the loss beyond the universal term grows with the number of competing answers that the distractors offer. Replacing the competing entities of the answer's type with generic descriptors such as ``a person'' recovers only part of the loss (Appendix Table~\ref{tab:compctl}).

\subsection{Target versus distractor replacement}
\label{sec:replace}

To locate the recoverable damage, we take the position-corrected cache under SIM and replace the KV of selected chunks with their KV from the full-prefill cache, that is, the KV those chunks would have had if they had been encoded in context (Appendix Tables~\ref{tab:replacement} and~\ref{tab:replacement_ci}). Replacing the distractors recovers more than replacing the target. With all of them replaced, retention is higher than with only the target replaced in all twelve settings, significantly so in eleven, and 41--73\% of the loss is recovered on NQ-Open, TriviaQA and Gorilla. Compared with no replacement, replacing every distractor raises retention significantly in ten settings and lowers it significantly in one, while target replacement helps significantly in six and hurts significantly in one. By contrast, replacing a single distractor, the most similar one or a random one, changes retention significantly in only two of 24 cases (Appendix~\ref{app:results}).

\paragraph{Distractors add to the odds of losing the answer.} A simple model fits both the gain from restoring every distractor and the small change from restoring a single one. In this model, each reused distractor $j$ adds a weight $w_j$ to the odds of failure,
\begin{equation}
\label{eq:odds}
\frac{1-R}{R}=c_0+\sum_{j\neq t}w_j ,
\end{equation}
where $t$ is the target, restoring the full-prefill KV of $j$ removes $w_j$, and $c_0$ is what remains with every distractor restored. If a random set of $k$ of the $n$ distractors carries $k/n$ of the total weight, restoring it gives \mbox{$1/R(k)=(1-k/n)/R(0)+(k/n)/R(n)$}, with no parameter beyond the two measured end points $R(0)$ and $R(n)$. We test this prediction against two alternatives, linear in $R$ and linear in the probit of $R$, by restoring $k\in\{4,8,16,24\}$ random distractors on open-domain QA. At $k=16$, where the predictions differ most, the prediction of the odds model lies within the 95\% interval of the measured retention and the linear prediction lies outside it in five of the six settings; in the sixth setting, all three predictions lie within the interval (Appendix Table~\ref{tab:ksweep} and Figure~\ref{fig:ksweep}).

\paragraph{The target's KV depends on the chunks before it.} Target replacement inserts KV that was computed against the full-prefill KV of the earlier chunks, but the question then attends to it together with the reused KV of those chunks. We call a chunk's KV contextually compatible with a cache if it was computed against the KV of the earlier chunks in that cache. On 270 \Qs{} questions in seven settings, inserting the target's full-prefill KV raises retention by +0.11 [+0.06, +0.17] when the earlier chunks stay reused and by +0.25 [+0.20, +0.31] when their full-prefill KV is restored as well, so the interaction is +0.14 [+0.08, +0.20]. With the earlier chunks reused, the inserted KV helps significantly only in the three Mistral-7B settings (Appendix~\ref{app:compat}). Recomputing the target against the reused cache makes its KV compatible. The recomputed target significantly beats the inserted full-prefill KV in all seven settings, by 0.13 to 0.38, and four of these differences remain significant after Holm's correction (Appendix Tables~\ref{tab:compat} and~\ref{tab:compat2}). This holds although the recomputed KV stays at a median relative distance of 0.23--0.30 from the full-prefill KV, compared with 0.34--0.45 for the unrepaired KV.

\paragraph{Where the deviation starts.} Under reuse, every chunk except the first, whether a distractor or the target, differs from full prefill in two ways: it cannot attend to the chunks before it, and it is encoded at positions closer to the prefix. In the first layer, where every token enters with its embedding, these two differences account exactly for the change in each attention head.
\begin{proposition}\label{prop:layer}
Fix a head and a token $u$ of $C_i$, and suppose that $u$ and every token it attends to under either encoding have the same hidden states under both. Let $\alpha_u$ be the attention of $u$ on $C_{<i}$ under full prefill, and $o^{\mathrm{own}}_u$ and $o^{\mathrm{cross}}_u$ the head's outputs renormalised over $C_0$ and the tokens of $C_i$ up to $u$, and over $C_{<i}$, at the full-prefill positions. Then
\begin{equation}
\label{eq:layer}
\begin{split}
o^{\mathrm{full}}_u-o^{\mathrm{reuse}}_u={}&\alpha_u\bigl(o^{\mathrm{cross}}_u-o^{\mathrm{own}}_u\bigr)\\
&+\bigl(o^{\mathrm{own}}_u-o^{\mathrm{reuse}}_u\bigr),
\end{split}
\end{equation}
and the second term vanishes when $s_i=P$. Both terms vanish for $C_1$, whose KV is the same under the two encodings in every layer.
\end{proposition}
The proof splits the softmax normaliser (Appendix~\ref{app:prop}). In the first layer of the three models, $\alpha_u$ averaged over heads and the tokens of a chunk ($\bar\alpha$) rises with the chunk's position in all nine open-domain QA and API-documentation settings (Appendix Table~\ref{tab:layer0}). A chunk thus misses more of its context the more chunks precede it.

\section{Question Attention over Reused Caches}
\label{sec:attention}

Several online repair methods choose what to recompute from the attention of the question tokens over the reused cache \citep{prophetkv,surrokv}. If similar chunks compete with the target, that attention may be a poor guide to the target. We test three detectors of the answer-bearing chunk on development questions that full prefill answers correctly (Appendix Table~\ref{tab:detect}). The first, $d_1$, sums the question's attention over each chunk. The second, $d_2$, averages this attention over the upper half of the layers after dropping the first 16 tokens of each chunk, to discount the attention that independently encoded chunks place on their first tokens. The third, $d_3$, is the cosine similarity between the BGE-M3 embeddings of the question and of each chunk, computed without the model. For Gorilla, the question is the user request alone (Appendix~\ref{app:data}).

Over reused caches, the question's summed attention peaks most often on the chunk just before the question. Detector $d_1$ selects that chunk in 58--100\% of questions, and in every question in seven of the twelve settings. The target is selected in at most 12.5\% of questions by $d_1$ and in at most 18.8\% by $d_2$, while chance is 3.2--6.4\%. Over full-prefill caches, the same detectors do better in 20 of the 24 combinations of detector and setting, with scores of up to 89.3\%. The retriever $d_3$ finds the target in 50--64\% of open-domain QA questions, 50--52\% of dialogue-memory questions and 36--56\% of Gorilla questions.

\section{Implications for Budgeted Repair}
\label{sec:repair}

\paragraph{Spending a small budget on chunks.}
\label{sec:alloc}
Restoring the distractors costs almost a full prefill (Section~\ref{sec:replace}), and a small budget spent on a few of them does worse than the token-level selector. With a 10\% budget, recomputing the four chunks most similar to the target, without the target itself, is significantly worse than the selector in eleven of the twelve settings (Appendix Table~\ref{tab:alloc_controls}). We therefore spend a small budget on the chunk that holds the answer, which question attention rarely finds (Section~\ref{sec:attention}). For each setting, we use the development questions to choose a token-level rule from CacheBlend, EPIC, ProphetKV and question attention \citep{cacheblend,epic,prophetkv}. The token-level selector applies this rule to all chunks. Each chunk-level allocation spends the same budget on its chosen chunks and ranks the tokens within them by the same rule (Appendix~\ref{app:impl} describes where this selector differs from the published methods). \emph{Target} recomputes the known target chunk, an oracle reference that holds 2.5--3.3\% of the context on open-domain QA and API documentation; \emph{Target+4} also recomputes the four chunks most similar to the target, and \emph{Retrieved} and \emph{Retrieved+4} use the retriever's top chunk in place of the target. Table~\ref{tab:alloc} gives the paired gains, and Appendix Tables~\ref{tab:alloc_ci5}--\ref{tab:alloc_controls} give intervals and controls.

\begin{table*}[t]
\centering
\footnotesize
\setlength{\tabcolsep}{2pt}
\begin{tabular}{llccccccccccc}
\toprule
 & & & \multicolumn{5}{c}{5\% budget} & \multicolumn{5}{c}{10\% budget} \\ \cmidrule(lr){4-8}\cmidrule(lr){9-13}
 & & & Retention & \multicolumn{4}{c}{Gain over Token} & Retention & \multicolumn{4}{c}{Gain over Token} \\ \cmidrule(lr){5-8}\cmidrule(lr){10-13}
Dataset & Model & $N$ & Token & Target & \begin{tabular}[b]{@{}c@{}}Target\\+4\end{tabular} & Retrieved & \begin{tabular}[b]{@{}c@{}}Retrieved\\+4\end{tabular} & Token & Target & \begin{tabular}[b]{@{}c@{}}Target\\+4\end{tabular} & Retrieved & \begin{tabular}[b]{@{}c@{}}Retrieved\\+4\end{tabular} \\
\midrule
LoCoMo & Qwen3-8B & 105 & 0.71 & \textminus{}0.04 & \textminus{}0.06 & \textbf{\textminus{}0.10} & \textbf{\textminus{}0.10} & 0.72 & \textminus{}0.04 & \textminus{}0.04 & \textbf{\textminus{}0.10} & \textbf{\textminus{}0.09} \\
 & Llama-3.1-8B & 102 & 0.85 & +0.01 & +0.01 & \textminus{}0.01 & \textminus{}0.05 & 0.86 & +0.00 & \textminus{}0.02 & \textminus{}0.02 & \textminus{}0.05 \\
 & Mistral-7B & 36 & 0.78 & +0.03 & \textminus{}0.03 & \textminus{}0.03 & +0.03 & 0.75 & +0.00 & +0.00 & +0.00 & +0.08 \\
\midrule
NQ-Open & Qwen3-8B & 146 & 0.16 & \textbf{+0.35} & \textbf{+0.29} & \textbf{+0.22} & \textbf{+0.21} & 0.24 & \textbf{+0.27} & \textbf{+0.33} & \textbf{+0.14} & \textbf{+0.24} \\
 & Llama-3.1-8B & 128 & 0.21 & \textbf{+0.15} & \textbf{+0.18} & +0.08 & \textbf{+0.12} & 0.38 & \textminus{}0.02 & +0.00 & \textbf{\textminus{}0.09} & \textminus{}0.07 \\
 & Mistral-7B & 101 & 0.28 & \textbf{+0.31} & \textbf{+0.21} & \textbf{+0.14} & \textbf{+0.15} & 0.34 & \textbf{+0.25} & \textbf{+0.25} & +0.08 & \textbf{+0.14} \\
\midrule
Gorilla & Qwen3-8B & 89 & 0.06 & \textbf{+0.17} & \textbf{+0.08} & +0.09 & \textbf{+0.08} & 0.19 & +0.03 & +0.02 & \textminus{}0.04 & +0.00 \\
 & Llama-3.1-8B & 100 & 0.12 & \textbf{+0.10} & +0.06 & +0.02 & +0.01 & 0.31 & \textminus{}0.09 & \textbf{\textminus{}0.11} & \textbf{\textminus{}0.17} & \textbf{\textminus{}0.17} \\
 & Mistral-7B & 108 & 0.10 & \textbf{+0.49} & \textbf{+0.34} & \textbf{+0.19} & \textbf{+0.21} & 0.30 & \textbf{+0.30} & \textbf{+0.23} & +0.00 & +0.09 \\
\midrule
TriviaQA & Qwen3-8B & 151 & 0.32 & \textbf{+0.20} & \textbf{+0.23} & \textbf{+0.14} & \textbf{+0.17} & 0.49 & +0.03 & +0.08 & \textminus{}0.03 & +0.03 \\
 & Llama-3.1-8B & 127 & 0.46 & \textbf{+0.11} & \textbf{+0.09} & \textminus{}0.02 & +0.02 & 0.55 & +0.02 & +0.04 & \textbf{\textminus{}0.11} & \textminus{}0.06 \\
 & Mistral-7B & 117 & 0.44 & \textbf{+0.23} & \textbf{+0.18} & \textbf{+0.10} & \textbf{+0.11} & 0.52 & \textbf{+0.15} & \textbf{+0.13} & +0.03 & +0.07 \\
\bottomrule
\end{tabular}
\caption{Recomputing chosen chunks instead of tokens. Token: retention of the token-level selector at the budget. The other columns give paired gains over Token when the same budget is spent on the target chunk alone (Target), on the target and its four most similar chunks (Target+4), or on the BGE-M3 top-1 chunk alone or with its four most similar chunks (Retrieved, Retrieved+4). On open-domain QA and API documentation, a single chunk is about 3\% of the context and is recomputed in full, except on a few questions at 5\%. Target and Retrieved therefore score identically at both budgets outside LoCoMo, and their 10\% gains differ from their 5\% gains only because the 10\% Token baseline is stronger. Bold gains have 95\% intervals that exclude zero (Appendix Tables~\ref{tab:alloc_ci5} and~\ref{tab:alloc_ci10}). A LoCoMo session is longer than a passage, and a whole session fits within the 5\% budget on few questions (Section~\ref{sec:alloc}).}
\label{tab:alloc}
\end{table*}

\paragraph{The target chunk.} With a 5\% budget, recomputing the known target alone beats the token-level selector in all nine open-domain QA and API-documentation settings (Table~\ref{tab:alloc}), and it usually uses less than the budget. When the budget doubles to 10\%, the target remains significantly ahead of the selector in four of them and is never significantly behind. The gain depends on where the target sits. Averaged over the nine settings, recomputing the target raises retention by 0.13 when the target lies in the first third of the prompt, by 0.35 when it lies in the middle third and by 0.47 when it lies in the last third (Appendix Table~\ref{tab:position} and Figure~\ref{fig:position}). On average, therefore, the more chunks precede the target, the more its recomputation adds.

\paragraph{The retriever's choice.} \looseness=-1 The retriever ranks the target first on only 47--64\% of the \Qs{} questions in the nine open-domain QA and API-documentation settings. Even so, recomputing its top chunk with a 5\% budget significantly beats the token-level selector in five of these settings, and recomputing the four chunks most similar to it as well raises the count to seven.

\paragraph{Moving the retrieved chunk last.} \looseness=-1 Under reuse, the chunk order is a free choice, because every cache is re-rotated when the prompt is assembled. InfoFlow KV \citep{infoflowkv} uses this freedom to move the chunks in which question attention is densest toward the question. Since a recomputed chunk gains more on average the more chunks precede it, we move the retriever's top chunk to the position just before the question (Appendix Tables~\ref{tab:reorder} and~\ref{tab:reorder_tok}). Recomputing the chunk in its new position gives significantly higher retention than recomputing it in place in all nine open-domain QA and API-documentation settings, by 0.07 to 0.18. The move also significantly raises the retention of the token-level selector, by 0.11--0.25, in the seven of these settings whose rule ranks tokens by question attention, which over reused caches peaks on the last chunk (Section~\ref{sec:attention}). To separate the effect of the move from that of the recomputation, we compare the moved chunk with selected tokens recomputed on the same reordered prompt, and with InfoFlow-style reordering at the same budget. Against selected tokens, the moved chunk is never significantly worse, and in four settings it is significantly better, by 0.09 to 0.31. In the others the difference lies between \textminus{}0.07 and +0.04. Compared with InfoFlow-style reordering at the same 5\% budget, the moved chunk wins significantly in six of the nine and never loses significantly.

\paragraph{Moving the top three chunks.} Moving the retriever's top three chunks instead, with the top one last, extends the gain to questions on which the top chunk is not the target. With a 10\% budget, recomputing only the moved top chunk leaves the rest of the budget unspent, and recomputing all three chunks does better in all nine settings, by 0.06 to 0.16, significantly in seven. If the moved chunks help only when they include the target, the retriever's hit rates predict the size of this gain (Appendix~\ref{app:topk}). The prediction lies within the interval of the measured gain in seven of the nine and above it in the other two (Appendix Table~\ref{tab:topk}, and Table~\ref{tab:topkbm25} with BM25 as the retriever). Selected tokens recomputed on the same reordered prompt remain competitive at this budget. The three chunks beat them significantly in only three settings and lose significantly in two, both with Llama-3.1-8B.

\paragraph{Dialogue memory.} A whole LoCoMo session fits within the 5\% budget on only 11--21\% of the questions, but within the 10\% budget on 89--94\%. Relative to the token-level selector, recomputing the target session changes retention by \textminus{}0.04 to +0.03, never significantly, even with the 10\% budget, so the larger size of LoCoMo chunks does not explain the lack of gain. Against a ProphetKV selector, recomputing the target session gains significantly for Qwen3-14B (+0.14) but not for Qwen3-32B (+0.07; Appendix~\ref{app:larger}).

\begin{table}[t]
\centering
\small
\setlength{\tabcolsep}{2.9pt}
\begin{tabular}{llccccc}
\toprule
 & & \multicolumn{3}{c}{Loss and terms} & \multicolumn{2}{c}{Gain over Token} \\ \cmidrule(lr){3-5}\cmidrule(lr){6-7}
Dataset & Model & Loss & Same & Sim. & Target & \begin{tabular}[b]{@{}c@{}}Chunk\\moved\end{tabular} \\
\midrule
LoCoMo & 14B & \textbf{0.36} & \textbf{+0.20} & +0.03 & \textbf{+0.14} & -- \\
 & 32B & \textbf{0.46} & \textbf{+0.16} & \textbf{+0.18} & +0.07 & -- \\
\midrule
NQ-Open & 14B & \textbf{0.79} & \textbf{+0.17} & +0.08 & \textbf{+0.22} & \textbf{+0.34} \\
 & 32B & \textbf{0.78} & \textbf{+0.37} & +0.08 & \textbf{+0.38} & \textbf{+0.35} \\
\midrule
Gorilla & 14B & \textbf{0.90} & \textbf{+0.58} & \textminus{}0.10$^\dagger$ & \textbf{+0.45} & \textbf{+0.38} \\
 & 32B & \textbf{0.93} & \textbf{+0.90} & \textminus{}0.07$^\dagger$ & \textbf{+0.43} & \textbf{+0.30} \\
\midrule
TriviaQA & 14B & \textbf{0.77} & \textbf{+0.32} & \textbf{+0.20} & \textbf{+0.20} & \textbf{+0.24} \\
 & 32B & \textbf{0.72} & \textbf{+0.24} & \textbf{+0.25} & \textbf{+0.25} & \textbf{+0.22} \\
\bottomrule
\end{tabular}
\caption{Qwen3-14B and Qwen3-32B on \Qs{} (67--157 questions per setting): the five results tested for replication. Loss is $1-R(\mathrm{SIM})$, and Same and Sim.\ are the same-category and similarity terms of Equation~\ref{eq:decomp}. Target and Chunk moved are the gains over the token-level selector (ProphetKV) at 5\% when the budget recomputes the target or the retriever\textquoteright s top chunk moved to the last position. Bold: the 95\% interval excludes zero. $^\dagger$Retention under RAND is zero, so the term cannot be positive and is not set in bold. Appendix Table~\ref{tab:larger} gives the intervals.}
\label{tab:larger_main}
\end{table}

\paragraph{Larger models.} Qwen3-14B and Qwen3-32B, run on the main conditions with ProphetKV as the token-level rule, reproduce all five results that we test for replication (Table~\ref{tab:larger_main}; Appendix~\ref{app:larger} reports each result).

\section{Related Work}
\label{sec:related}

\paragraph{Non-prefix and parallel KV reuse.} \looseness=-1 Prompt Cache \citep{promptcache}, TurboRAG \citep{turborag}, Block-Attention \citep{blockattn}, KVLink \citep{kvlink} and APE \citep{ape} encode chunks independently or in parallel and concatenate their caches with position adjustments; some of them adapt the model to this layout with fine-tuning or modified attention. \citet{attnentropy} trace the loss of parallel encoding to unusually high attention entropy and reduce it with attention sinks and selective attention. Proposition~\ref{prop:layer} instead splits the first-layer deviation exactly into a cross-chunk term and a position term, and our distractor sources show that the loss also depends on what the other chunks contain.

\paragraph{Repairing reused caches.} \looseness=-1 Token-level repairs such as CacheBlend \citep{cacheblend}, EPIC \citep{epic}, ProphetKV \citep{prophetkv} and CacheTune \citep{cachetune} differ in how they choose the tokens to recompute (Appendix~\ref{app:related} and Table~\ref{tab:assume}). InfoFlow KV \citep{infoflowkv} selects tokens by question attention; before recomputing, it moves the chunks in which the selected tokens are densest to positions next to the question. Section~\ref{sec:alloc} compares with it at the same budget. SurroKV \citep{surrokv}, a concurrent submission, encodes chunks after a fixed surrogate prefix and then recomputes tokens selected by question attention; Appendix Table~\ref{tab:repair} evaluates surrogate prefixes, with and without online recomputation, in our settings.

\paragraph{Chunk and suffix repair.} \looseness=-1 SAFE-KV \citep{safekv} starts at a selected document and re-encodes the rest of the prompt over the retained reused cache of the earlier documents. It takes full prefill of the prompt in its original order as the reference and does not reorder documents. On a reordered prompt, recomputing the chunk moved last is the SAFE-KV suffix repair that starts at the last chunk; unlike SAFE-KV, we use the freedom of chunk order to shrink the suffix that needs repair to one chunk. Budgeted cache repair (BCR; \citealp{bcr}) recomputes cache rows or short slices rather than whole retrieved documents (Appendix~\ref{app:related}). Both in SAFE-KV and after small document edits \citep{contiguity}, copying full-prefill states into a cache and recomputing them in place give nearly the same result for a whole suffix or a contiguous window. The gap of Section~\ref{sec:replace} arises when the target alone is inserted while the chunks before it stay reused (Appendix~\ref{app:compat}). Appendix~\ref{app:related} covers further work.

\section{Conclusion}
\label{sec:conclusion}

\looseness=-1 Encoding retrieved chunks independently makes models lose a large share of the questions that need the retrieved context. The loss grows as the other chunks become more like the target, even in a model fine-tuned for block-wise encoding. Because the recoverable damage is spread over the distractors and accumulates in the odds of failure, restoring the distractors costs nearly a full prefill, and a small budget is better spent on the target. When the budget is spent on the target, our controls show that contextual compatibility, rather than closeness to full prefill, explains what helps: recomputing the target against the reused cache beats inserting its full-prefill KV while the chunks before it stay reused. Question attention over reused caches mostly goes to the chunk next to the question, so the target has to be found by other means. With a 5\% budget on open-domain QA and API documentation, recomputing the known target chunk beats token-level selectors. Moving the retriever's top chunk next to the question and recomputing it there beats InfoFlow-style reordering at the same budget in most settings, and it never does significantly worse than recomputing selected tokens on the same prompt.
These gains are bounded by retrieval: when the retriever does not rank the target first, the moved chunk does worse than selected tokens, and the gain from moving the top three chunks follows the retriever's hit rate in most settings. Two further limits do not depend on retrieval. With a 10\% budget, token-level selectors remain competitive, and for the 7B--8B models on dialogue memory, recomputing chunks never significantly improves on the token-level selector chosen for each setting. Evaluations of cache reuse should therefore use contexts built by retrieval, and on open-domain QA and API documentation, repair methods should locate the answer-bearing chunk rather than rely on attention over the reused cache.

\section*{Limitations}

Our experiments use 7B--8B instruction-tuned models on four English datasets. We repeat the main conditions with two larger Qwen3 models, but larger models and other languages are not covered. Models fine-tuned for chunked encoding are represented by a single released model, and we do not test repairs on it. The chunk-level allocation needs either the known target or a retriever, so its gains in deployment depend on retrieval quality. We study only small budgets and count them in recomputed tokens. Latency is measured only in our research engine, and in an optimised serving system, recomputing a contiguous chunk and recomputing scattered tokens may take different amounts of time. Finally, we compare with InfoFlow KV through our own re-implementation, on our models and datasets.

\bibliography{references}

\begin{thebibliography}{40}
\providecommand{\natexlab}[1]{#1}

\bibitem[{Anonymous(2026{\natexlab{a}})}]{surrokv}
Anonymous. 2026{\natexlab{a}}.
\newblock Construct before you repair: Surrogate contexts for non-prefix {KV}
  cache reuse.
\newblock Submitted to ICLR 2027, OpenReview
  \url{https://openreview.net/forum?id=OWX14VkewV}.

\bibitem[{Anonymous(2026{\natexlab{b}})}]{ocp}
Anonymous. 2026{\natexlab{b}}.
\newblock Predicting output preservation and first divergence under {KV}-cache
  compression.
\newblock Submitted to ICLR 2027, OpenReview
  \url{https://openreview.net/forum?id=zKxUf16XnW}.

\bibitem[{Anonymous(2026{\natexlab{c}})}]{stalespec}
Anonymous. 2026{\natexlab{c}}.
\newblock Speculating on the past: Lossless reuse of stale {KV} caches via
  self-speculative verification.
\newblock Submitted to ICLR 2027, OpenReview
  \url{https://openreview.net/forum?id=96mkAG7gh1}.

\bibitem[{Cao et~al.(2026)Cao, Mu, Fang, and Chen}]{cachereforge}
Yuhang Cao, Yanzhou Mu, Chunrong Fang, and Zhenyu Chen. 2026.
\newblock \href {https://arxiv.org/abs/2609.30884} {{CacheReforge}: Bounded
  recovery for stale {KV} caches under evolving adapters}.
\newblock \emph{Preprint}, arXiv:2609.30884.

\bibitem[{Cestola et~al.(2026)Cestola, Xia, Zheng, Zheng, Wang, Zhao, and
  Didona}]{cestola2026}
Samuel Cestola, Tianxiang Xia, Pengfei Zheng, Weiyan Zheng, Bo~Wang, Yi~Zhao,
  and Diego Didona. 2026.
\newblock \href {https://arxiv.org/abs/2609.31415} {Evaluating the accuracy of
  {KV} cache reuse techniques}.
\newblock \emph{Preprint}, arXiv:2609.31415.

\bibitem[{Chen et~al.(2024)Chen, Xiao, Zhang, Luo, Lian, and Liu}]{bgem3}
Jianlyu Chen, Shitao Xiao, Peitian Zhang, Kun Luo, Defu Lian, and Zheng Liu.
  2024.
\newblock \href {https://doi.org/10.18653/v1/2024.findings-acl.137}
  {{M3}-embedding: Multi-linguality, multi-functionality, multi-granularity
  text embeddings through self-knowledge distillation}.
\newblock In \emph{Findings of the Association for Computational Linguistics:
  ACL 2024}, pages 2318--2335.

\bibitem[{Chen et~al.(2026)Chen, Rao, Xiu, Zhang, Cui, Chen, and
  Zhao}]{latentremedy}
Zhiwei Chen, Zhongtao Rao, Waikit Xiu, Yuxiang Zhang, Shuting Cui, Jian Chen,
  and Tianxiang Zhao. 2026.
\newblock \href {https://arxiv.org/abs/2610.18014} {{LatentRemedy}: Efficient
  remedy for non-prefix {KV} cache reuse in multi-agent systems}.
\newblock \emph{Preprint}, arXiv:2610.18014.

\bibitem[{Du et~al.(2026)Du, Chen, Tang, Liu, Lan, Qu, Niu, Liu, Chen, and
  Wu}]{c2kv}
Chuheng Du, Junyi Chen, Hanlin Tang, Kan Liu, Tao Lan, Lin Qu, Chaoyue Niu,
  Shengzhong Liu, Guihai Chen, and Fan Wu. 2026.
\newblock \href {https://doi.org/10.1145/3770855.3817715} {{C$^2$KV}:
  Compressed and composable {KV} cache reuse for efficient {LLM} inference}.
\newblock In \emph{Proceedings of the 32nd ACM SIGKDD Conference on Knowledge
  Discovery and Data Mining (KDD)}, pages 991--1002.

\bibitem[{Gim et~al.(2024)Gim, Chen, Lee, Sarda, Khandelwal, and
  Zhong}]{promptcache}
In~Gim, Guojun Chen, Seung-seob Lee, Nikhil Sarda, Anurag Khandelwal, and Lin
  Zhong. 2024.
\newblock \href {https://arxiv.org/abs/2311.04934} {Prompt cache: Modular
  attention reuse for low-latency inference}.
\newblock In \emph{Proceedings of Machine Learning and Systems (MLSys)}.

\bibitem[{Grattafiori et~al.(2024)Grattafiori, Dubey, Jauhri, Pandey, Kadian,
  Al-Dahle, Letman, Mathur, Schelten, Vaughan, Yang et~al.}]{llama3}
Aaron Grattafiori, Abhimanyu Dubey, Abhinav Jauhri, Abhinav Pandey, Abhishek
  Kadian, Ahmad Al-Dahle, Aiesha Letman, Akhil Mathur, Alan Schelten, Alex
  Vaughan, Amy Yang, et~al. 2024.
\newblock \href {https://arxiv.org/abs/2407.21783} {The {Llama} 3 herd of
  models}.
\newblock \emph{Preprint}, arXiv:2407.21783.

\bibitem[{Hu et~al.(2025)Hu, Huang, Wang, Wang, Hu, Zhang, Feng, Chen, Shan,
  and Xie}]{epic}
Junhao Hu, Wenrui Huang, Weidong Wang, Haoyi Wang, Tiancheng Hu, Qin Zhang, Hao
  Feng, Xusheng Chen, Yizhou Shan, and Tao Xie. 2025.
\newblock \href {https://proceedings.mlr.press/v267/hu25j.html} {{EPIC}:
  Efficient position-independent caching for serving large language models}.
\newblock In \emph{Proceedings of the 42nd International Conference on Machine
  Learning}, volume 267 of \emph{Proceedings of Machine Learning Research},
  pages 24391--24402. PMLR.

\bibitem[{Ikeda et~al.(2026)Ikeda, Miyashita, and Deguchi}]{kvrepurpose}
Taiga Ikeda, Daisuke Miyashita, and Jun Deguchi. 2026.
\newblock \href {https://arxiv.org/abs/2610.13060} {{KV-Repurpose}: A testbed
  for evaluating non-prefix {KV}-cache reuse}.
\newblock \emph{Preprint}, arXiv:2610.13060.

\bibitem[{Izacard et~al.(2022)Izacard, Caron, Hosseini, Riedel, Bojanowski,
  Joulin, and Grave}]{contriever}
Gautier Izacard, Mathilde Caron, Lucas Hosseini, Sebastian Riedel, Piotr
  Bojanowski, Armand Joulin, and Edouard Grave. 2022.
\newblock \href {https://arxiv.org/abs/2112.09118} {Unsupervised dense
  information retrieval with contrastive learning}.
\newblock \emph{Transactions on Machine Learning Research}.

\bibitem[{Jiang et~al.(2023)Jiang, Sablayrolles, Mensch, Bamford, Chaplot,
  de~las Casas, Bressand, Lengyel, Lample, Saulnier, Lavaud, Lachaux, Stock,
  Le~Scao, Lavril, Wang, Lacroix, and El~Sayed}]{mistral}
Albert~Q. Jiang, Alexandre Sablayrolles, Arthur Mensch, Chris Bamford,
  Devendra~Singh Chaplot, Diego de~las Casas, Florian Bressand, Gianna Lengyel,
  Guillaume Lample, Lucile Saulnier, Lélio~Renard Lavaud, Marie-Anne Lachaux,
  Pierre Stock, Teven Le~Scao, Thibaut Lavril, Thomas Wang, Timothée Lacroix,
  and William El~Sayed. 2023.
\newblock \href {https://arxiv.org/abs/2310.06825} {{Mistral 7B}}.
\newblock \emph{Preprint}, arXiv:2310.06825.

\bibitem[{Joshi et~al.(2017)Joshi, Choi, Weld, and Zettlemoyer}]{triviaqa}
Mandar Joshi, Eunsol Choi, Daniel~S. Weld, and Luke Zettlemoyer. 2017.
\newblock \href {https://arxiv.org/abs/1705.03551} {{TriviaQA}: A large scale
  distantly supervised challenge dataset for reading comprehension}.
\newblock In \emph{Proceedings of ACL}, pages 1601--1611.

\bibitem[{Kang and Yoo(2026)}]{bcr}
Haeyong Kang and Chang~D. Yoo. 2026.
\newblock \href {https://arxiv.org/abs/2610.02233} {Budgeted cache repair for
  cross-context {KV}-cache reuse}.
\newblock \emph{Preprint}, arXiv:2610.02233.

\bibitem[{Karpukhin et~al.(2020)Karpukhin, Oğuz, Min, Lewis, Wu, Edunov, Chen,
  and Yih}]{dpr}
Vladimir Karpukhin, Barlas Oğuz, Sewon Min, Patrick Lewis, Ledell Wu, Sergey
  Edunov, Danqi Chen, and Wen-tau Yih. 2020.
\newblock \href {https://arxiv.org/abs/2004.04906} {Dense passage retrieval for
  open-domain question answering}.
\newblock In \emph{Proceedings of EMNLP}, pages 6769--6781.

\bibitem[{Kwiatkowski et~al.(2019)Kwiatkowski, Palomaki, Redfield, Collins,
  Parikh, Alberti, Epstein, Polosukhin, Devlin, Lee, Toutanova, Jones, Kelcey,
  Chang, Dai, Uszkoreit, Le, and Petrov}]{nq}
Tom Kwiatkowski, Jennimaria Palomaki, Olivia Redfield, Michael Collins, Ankur
  Parikh, Chris Alberti, Danielle Epstein, Illia Polosukhin, Jacob Devlin,
  Kenton Lee, Kristina Toutanova, Llion Jones, Matthew Kelcey, Ming-Wei Chang,
  Andrew~M. Dai, Jakob Uszkoreit, Quoc Le, and Slav Petrov. 2019.
\newblock \href {https://doi.org/10.1162/tacl_a_00276} {Natural questions: A
  benchmark for question answering research}.
\newblock \emph{Transactions of the Association for Computational Linguistics},
  7:453--466.

\bibitem[{Li et~al.(2026)Li, Liu, Liu, Cui, Nie, Wang, and Wu}]{cachetune}
Fei Li, Song Liu, Yan Liu, Jinhua Cui, Shiqiang Nie, Jinyu Wang, and Weiguo Wu.
  2026.
\newblock \href {https://arxiv.org/abs/2605.24022} {Adaptive {KV} cache reuse
  for fast long-context {LLM} serving}.
\newblock \emph{Preprint}, arXiv:2605.24022.

\bibitem[{Liu et~al.(2024)Liu, Lin, Hewitt, Paranjape, Bevilacqua, Petroni, and
  Liang}]{lostmiddle}
Nelson~F. Liu, Kevin Lin, John Hewitt, Ashwin Paranjape, Michele Bevilacqua,
  Fabio Petroni, and Percy Liang. 2024.
\newblock \href {https://doi.org/10.1162/tacl_a_00638} {Lost in the middle: How
  language models use long contexts}.
\newblock \emph{Transactions of the Association for Computational Linguistics},
  12:157--173.

\bibitem[{Lu et~al.(2025)Lu, Wang, Rong, Chen, and Tang}]{turborag}
Songshuo Lu, Hua Wang, Yutian Rong, Zhi Chen, and Yaohua Tang. 2025.
\newblock \href {https://doi.org/10.18653/v1/2025.emnlp-main.334} {{TurboRAG}:
  Accelerating retrieval-augmented generation with precomputed {KV} caches for
  chunked text}.
\newblock In \emph{Proceedings of the 2025 Conference on Empirical Methods in
  Natural Language Processing}, pages 6588--6601.

\bibitem[{Ma et~al.(2025)Ma, Wang, and Lan}]{blockattn}
Dongyang Ma, Yan Wang, and Tian Lan. 2025.
\newblock \href {https://arxiv.org/abs/2409.15355} {Block-attention for
  efficient prefilling}.
\newblock In \emph{International Conference on Learning Representations
  (ICLR)}.

\bibitem[{Maharana et~al.(2024)Maharana, Lee, Tulyakov, Bansal, Barbieri, and
  Fang}]{locomo}
Adyasha Maharana, Dong-Ho Lee, Sergey Tulyakov, Mohit Bansal, Francesco
  Barbieri, and Yuwei Fang. 2024.
\newblock \href {https://doi.org/10.18653/v1/2024.acl-long.747} {Evaluating
  very long-term conversational memory of {LLM} agents}.
\newblock In \emph{Proceedings of the 62nd Annual Meeting of the Association
  for Computational Linguistics (Volume 1: Long Papers)}, pages 13851--13870.

\bibitem[{Mao et~al.(2026)Mao, Mackey, and Lin}]{contiguity}
Mingyang Mao, Wyatt Mackey, and Xiaomin Lin. 2026.
\newblock \href {https://arxiv.org/abs/2609.17983} {Contiguity, not importance:
  Budgeted repair of stale {KV} caches after document edits}.
\newblock \emph{Preprint}, arXiv:2609.17983.

\bibitem[{Pandey et~al.(2026)Pandey, Kong, Hu, Zhao, Zhao, Gungor, Zhang, and
  Rosing}]{agentkvshift}
Nilesh~Prasad Pandey, Jason Kong, Lanxiang Hu, Quanling Zhao, Yujie Zhao, Onat
  Gungor, Hao Zhang, and Tajana Rosing. 2026.
\newblock \href {https://arxiv.org/abs/2607.21604} {{AgentKVShift}: Efficient
  {KV} cache reuse for agentic memory systems}.
\newblock \emph{Preprint}, arXiv:2607.21604.

\bibitem[{Patil et~al.(2024)Patil, Zhang, Wang, and Gonzalez}]{gorilla}
Shishir~G. Patil, Tianjun Zhang, Xin Wang, and Joseph~E. Gonzalez. 2024.
\newblock \href {https://arxiv.org/abs/2305.15334} {Gorilla: Large language
  model connected with massive {API}s}.
\newblock In \emph{Advances in Neural Information Processing Systems
  (NeurIPS)}.

\bibitem[{Qi et~al.(2026)Qi, Liu, Wu, Jin, Chen, Qin, Chen, and Shao}]{relaxkv}
Ruoling Qi, Yirui Liu, Xuaner Wu, Yuxin Jin, Jian Chen, Jiayu Qin, Yin Chen,
  and Jiawei Shao. 2026.
\newblock \href {https://arxiv.org/abs/2609.33503} {{RelaxKV}: Recomputation
  guided by the query with sparse context attention for efficient {KV} cache
  reuse}.
\newblock \emph{Preprint}, arXiv:2609.33503.

\bibitem[{Robertson and Zaragoza(2009)}]{robertson2009bm25}
Stephen Robertson and Hugo Zaragoza. 2009.
\newblock \href {https://doi.org/10.1561/1500000019} {The probabilistic
  relevance framework: {BM25} and beyond}.
\newblock \emph{Foundations and Trends in Information Retrieval}.

\bibitem[{Shi et~al.(2026)Shi, Zheng, and Lou}]{kvsharearena}
Xi~Shi, Mengxin Zheng, and Qian Lou. 2026.
\newblock \href {https://arxiv.org/abs/2609.10266} {{KVShareArena}: {KV}-cache
  reuse across contexts and model checkpoints}.
\newblock \emph{Preprint}, arXiv:2609.10266.

\bibitem[{Teng et~al.(2026)Teng, Zhang, Zheng, Zhuo, Zhou, and
  Wan}]{infoflowkv}
Xin Teng, Canyu Zhang, Shaoyi Zheng, Danyang Zhuo, Tianyi Zhou, and Shenji Wan.
  2026.
\newblock \href {https://proceedings.mlr.press/v306/teng26b.html} {{InfoFlow
  KV}: Information-flow-aware {KV} recomputation for long context}.
\newblock In \emph{Proceedings of the 43rd International Conference on Machine
  Learning}, volume 306 of \emph{Proceedings of Machine Learning Research},
  pages 120070--120088.

\bibitem[{Wang et~al.(2026)Wang, Chen, Pan, Huang, Hao, Zou, Xia, Qiu, Zhang,
  and Wang}]{prophetkv}
Shihao Wang, Jiahao Chen, Yanqi Pan, Hao Huang, Yichen Hao, Xiangyu Zou, Wen
  Xia, Chongyang Qiu, Wentao Zhang, and Pengfei Wang. 2026.
\newblock \href {https://proceedings.mlr.press/v306/wang26y.html} {{ProphetKV}:
  User-query-driven selective recomputation for efficient {KV} cache reuse in
  retrieval-augmented generation}.
\newblock In \emph{Proceedings of the 43rd International Conference on Machine
  Learning}, volume 306 of \emph{Proceedings of Machine Learning Research},
  pages 125415--125434.

\bibitem[{Xie et~al.(2026)Xie, Xiao, Deng, Dou, Yang, and Guo}]{sempic}
Hui Xie, Peng Xiao, Yutong Deng, Shuoran Dou, Jian Yang, and Jinyang Guo. 2026.
\newblock \href {https://arxiv.org/abs/2607.28069} {{SemPIC}: Learning semantic
  position-independent {KV} caches}.
\newblock \emph{Preprint}, arXiv:2607.28069.

\bibitem[{Xu et~al.(2026)Xu, Wang, and Tian}]{transitionoverlap}
Liang Xu, Zhichun Wang, and Xuetao Tian. 2026.
\newblock \href {https://doi.org/10.18653/2026.findings-emnlp.1520} {Transition
  overlap: Construction-time conditioning for position-independent {KV} cache
  reuse}.
\newblock In \emph{Findings of the Association for Computational Linguistics:
  EMNLP 2026}, pages 32366--32386.

\bibitem[{Yang et~al.(2025{\natexlab{a}})Yang, Li, Yang, Zhang, Hui, Zheng, Yu,
  Gao, Huang, Lv, Zheng, Liu, Zhou, Huang, Hu, Ge, Wei, Lin, Tang, Yang
  et~al.}]{qwen3}
An~Yang, Anfeng Li, Baosong Yang, Beichen Zhang, Binyuan Hui, Bo~Zheng, Bowen
  Yu, Chang Gao, Chengen Huang, Chenxu Lv, Chujie Zheng, Dayiheng Liu, Fan
  Zhou, Fei Huang, Feng Hu, Hao Ge, Haoran Wei, Huan Lin, Jialong Tang, et~al.
  2025{\natexlab{a}}.
\newblock \href {https://arxiv.org/abs/2505.09388} {{Qwen3} technical report}.
\newblock \emph{Preprint}, arXiv:2505.09388.

\bibitem[{Yang et~al.(2025{\natexlab{b}})Yang, Hou, Wei, Bao, and
  Chang}]{kvlink}
Jingbo Yang, Bairu Hou, Wei Wei, Yujia Bao, and Shiyu Chang.
  2025{\natexlab{b}}.
\newblock \href {https://arxiv.org/abs/2502.16002} {{KVLink}: Accelerating
  large language models via efficient {KV} cache reuse}.
\newblock In \emph{Advances in Neural Information Processing Systems
  (NeurIPS)}.

\bibitem[{Yang et~al.(2025{\natexlab{c}})Yang, Chen, and Chen}]{ape}
Xinyu Yang, Tianqi Chen, and Beidi Chen. 2025{\natexlab{c}}.
\newblock \href {https://arxiv.org/abs/2502.05431} {{APE}: Faster and longer
  context-augmented generation via adaptive parallel encoding}.
\newblock In \emph{International Conference on Learning Representations
  (ICLR)}.

\bibitem[{Yao et~al.(2025)Yao, Li, Liu, Ray, Cheng, Zhang, Du, Lu, and
  Jiang}]{cacheblend}
Jiayi Yao, Hanchen Li, Yuhan Liu, Siddhant Ray, Yihua Cheng, Qizheng Zhang,
  Kuntai Du, Shan Lu, and Junchen Jiang. 2025.
\newblock \href {https://doi.org/10.1145/3689031.3696098} {{CacheBlend}: Fast
  large language model serving for {RAG} with cached knowledge fusion}.
\newblock In \emph{Proceedings of the Twentieth European Conference on Computer
  Systems (EuroSys '25)}, pages 94--109.

\bibitem[{Ye et~al.(2025)Ye, Gao, Ma, Wang, Fu, Chung, Lin, Liu, Zhang, Zhuo,
  and Chen}]{kvcomm}
Hancheng Ye, Zhengqi Gao, Mingyuan Ma, Qinsi Wang, Yuzhe Fu, Ming-Yu Chung,
  Yueqian Lin, Zhijian Liu, Jianyi Zhang, Danyang Zhuo, and Yiran Chen. 2025.
\newblock \href {https://arxiv.org/abs/2510.12872} {{KVCOMM}: Online
  cross-context {KV}-cache communication for efficient {LLM}-based multi-agent
  systems}.
\newblock In \emph{Advances in Neural Information Processing Systems
  (NeurIPS)}.

\bibitem[{Zhang(2026)}]{safekv}
Xingjian Zhang. 2026.
\newblock \href {https://arxiv.org/abs/2610.15186} {{SAFE-KV}: Causal suffix
  repair for behavior-preserving non-prefix {KV} cache reuse in {RAG}}.
\newblock \emph{Preprint}, arXiv:2610.15186.
\newblock To appear in Findings of AACL-IJCNLP 2026.

\bibitem[{Zhang et~al.(2025)Zhang, Wang, Huang, Fang, Zhang, Deng, Li, and
  Yu}]{attnentropy}
Zhisong Zhang, Yan Wang, Xinting Huang, Tianqing Fang, Hongming Zhang, Chenlong
  Deng, Shuaiyi Li, and Dong Yu. 2025.
\newblock \href {https://doi.org/10.18653/v1/2025.acl-long.485} {Attention
  entropy is a key factor: An analysis of parallel context encoding with
  full-attention-based pre-trained language models}.
\newblock In \emph{Proceedings of the 63rd Annual Meeting of the Association
  for Computational Linguistics (Volume 1: Long Papers)}, pages 9840--9855,
  Vienna, Austria. Association for Computational Linguistics.

\end{thebibliography}

\section*{AI Usage Statement}
We used AI assistants, including OpenAI's GPT models and Anthropic's Claude, to help implement and check the experiment and analysis code, check the references, and edit the text. The authors take full responsibility for the content of this paper.

\appendix
\setcounter{dbltopnumber}{4}\renewcommand{\dbltopfraction}{0.95}\renewcommand{\dblfloatpagefraction}{0.7}
\setcounter{topnumber}{4}\renewcommand{\topfraction}{0.95}\renewcommand{\textfraction}{0.05}\renewcommand{\floatpagefraction}{0.7}
\let\origparagraph\paragraph
\renewcommand{\paragraph}{\Needspace{4\baselineskip}\origparagraph}

\section{Data and Prompts}
\label{app:data}

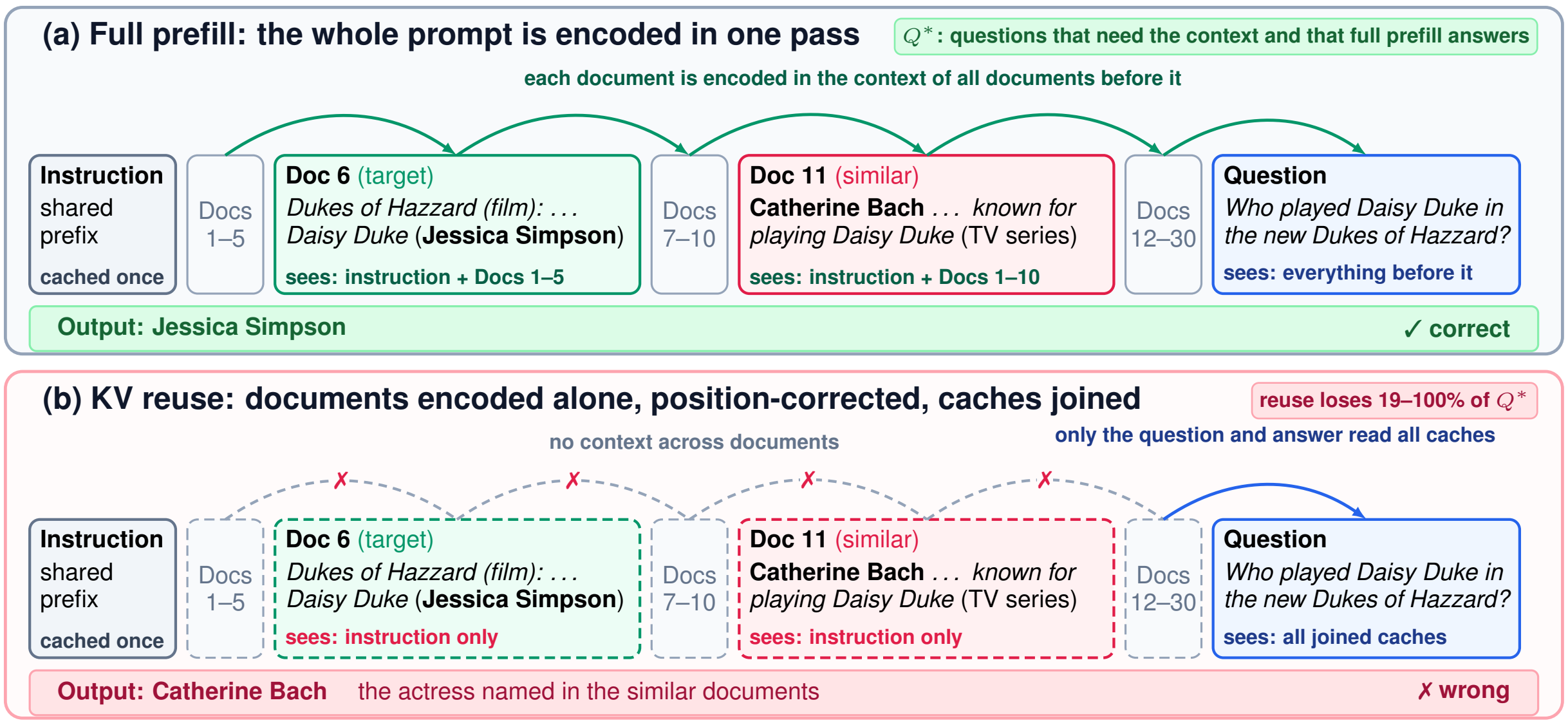
\begin{figure*}[t]
\centering
\definecolor{slA}{HTML}{F8FAFC}\definecolor{slB}{HTML}{F1F5F9}\definecolor{slD}{HTML}{94A3B8}\definecolor{slE}{HTML}{64748B}\definecolor{slF}{HTML}{334155}\definecolor{slG}{HTML}{0F172A}\definecolor{emA}{HTML}{ECFDF5}\definecolor{emB}{HTML}{059669}\definecolor{emC}{HTML}{065F46}\definecolor{grA}{HTML}{DCFCE7}\definecolor{grB}{HTML}{86EFAC}\definecolor{grC}{HTML}{166534}\definecolor{roA}{HTML}{FFF1F2}\definecolor{roB}{HTML}{E11D48}\definecolor{roD}{HTML}{FFE4E6}\definecolor{roE}{HTML}{FDA4AF}\definecolor{roF}{HTML}{9F1239}\definecolor{roG}{HTML}{FFFBFB}\definecolor{blA}{HTML}{EFF6FF}\definecolor{blB}{HTML}{2563EB}\definecolor{blC}{HTML}{1E3A8A}%
\begin{tikzpicture}[x=1cm, y=1cm]
\draw[rounded corners=5pt, fill=slA, draw=slD, line width=0.9pt] (0,0.00) rectangle (16.0,-3.56);
\node[anchor=west, font=\sffamily\footnotesize\bfseries, text=slG] at (0.25,-0.30) {(a) Full prefill: the whole prompt is encoded in one pass};
\node[anchor=east, rounded corners=2pt, fill=grA, draw=grB, line width=0.5pt, inner sep=2.2pt, font=\sffamily\tiny\bfseries, text=grC] at (15.75,-0.30) {$Q^*$: questions that need the context and that full prefill answers};
\draw[rounded corners=3pt, fill=slB, draw=slE, line width=0.8pt] (0.25,-1.52) rectangle ++(1.5,-1.42);
\node[anchor=north west, inner sep=0pt, text width=1.30cm, align=flush left, font=\sffamily\scriptsize] at (0.35,-1.62) {\textbf{Instruction}\\[1.5pt]shared prefix};
\node[anchor=south west, inner sep=0pt, font=\sffamily\tiny\bfseries, text=slF] at (0.35,-2.85) {cached once};
\coordinate (ai) at (1.00,-1.52);
\draw[rounded corners=3pt, fill=slA, draw=slD, line width=0.6pt] (1.87,-1.52) rectangle ++(0.78,-1.42);
\node[font=\sffamily\scriptsize, text=slE, align=center] at (2.26,-2.23) {Docs\\1--5};
\coordinate (am1) at (2.26,-1.52);
\draw[rounded corners=3pt, fill=emA, draw=emB, line width=0.8pt] (2.77,-1.52) rectangle ++(3.75,-1.42);
\node[anchor=north west, inner sep=0pt, text width=3.55cm, align=flush left, font=\sffamily\scriptsize] at (2.87,-1.62) {\textbf{Doc 6} \textcolor{emB}{(target)}\\[1.5pt]\textit{Dukes of Hazzard (film): \ldots{}}\\\textit{Daisy Duke} (\textbf{Jessica Simpson})};
\node[anchor=south west, inner sep=0pt, font=\sffamily\tiny\bfseries, text=emC] at (2.87,-2.85) {sees: instruction + Docs 1--5};
\coordinate (a6) at (4.64,-1.52);
\draw[rounded corners=3pt, fill=slA, draw=slD, line width=0.6pt] (6.64,-1.52) rectangle ++(0.78,-1.42);
\node[font=\sffamily\scriptsize, text=slE, align=center] at (7.03,-2.23) {Docs\\7--10};
\coordinate (am2) at (7.03,-1.52);
\draw[rounded corners=3pt, fill=roA, draw=roB, line width=0.8pt] (7.54,-1.52) rectangle ++(3.85,-1.42);
\node[anchor=north west, inner sep=0pt, text width=3.65cm, align=flush left, font=\sffamily\scriptsize] at (7.64,-1.62) {\textbf{Doc 11} \textcolor{roB}{(similar)}\\[1.5pt]\textbf{Catherine Bach} \textit{\ldots{} known for}\\\textit{playing Daisy Duke} (TV series)};
\node[anchor=south west, inner sep=0pt, font=\sffamily\tiny\bfseries, text=emC] at (7.64,-2.85) {sees: instruction + Docs 1--10};
\coordinate (a11) at (9.47,-1.52);
\draw[rounded corners=3pt, fill=slA, draw=slD, line width=0.6pt] (11.51,-1.52) rectangle ++(0.78,-1.42);
\node[font=\sffamily\scriptsize, text=slE, align=center] at (11.90,-2.23) {Docs\\12--30};
\coordinate (am3) at (11.90,-1.52);
\draw[rounded corners=3pt, fill=blA, draw=blB, line width=0.8pt] (12.41,-1.52) rectangle ++(3.15,-1.42);
\node[anchor=north west, inner sep=0pt, text width=2.95cm, align=flush left, font=\sffamily\scriptsize] at (12.51,-1.62) {\textbf{Question}\\[1.5pt]\textit{Who played Daisy Duke in}\\\textit{the new Dukes of Hazzard?}};
\node[anchor=south west, inner sep=0pt, font=\sffamily\tiny\bfseries, text=blC] at (12.51,-2.85) {sees: everything before it};
\coordinate (aq) at (13.99,-1.52);
\draw[-{Latex[length=1.5mm]}, line width=0.9pt, draw=emB] (am1) to[out=35,in=145] (a6);
\draw[-{Latex[length=1.5mm]}, line width=0.9pt, draw=emB] (a6) to[out=35,in=145] (am2);
\draw[-{Latex[length=1.5mm]}, line width=0.9pt, draw=emB] (am2) to[out=35,in=145] (a11);
\draw[-{Latex[length=1.5mm]}, line width=0.9pt, draw=emB] (a11) to[out=35,in=145] (am3);
\draw[-{Latex[length=1.5mm]}, line width=0.9pt, draw=emB] (am3) to[out=35,in=145] (aq);
\node[font=\sffamily\tiny\bfseries, text=emC] at (8.71,-0.72) {each document is encoded in the context of all documents before it};
\draw[rounded corners=2pt, fill=grA, draw=grB, line width=0.6pt] (0.25,-3.07) rectangle (15.75,-3.53);
\node[anchor=west, font=\sffamily\scriptsize, text=grC] at (0.4,-3.30) {\textbf{Output: Jessica Simpson}};
\node[anchor=east, font=\sffamily\scriptsize\bfseries, text=grC] at (15.6,-3.30) {\ding{51} correct};
\draw[rounded corners=5pt, fill=roG, draw=roE, line width=0.9pt] (0,-3.74) rectangle (16.0,-7.30);
\node[anchor=west, font=\sffamily\footnotesize\bfseries, text=slG] at (0.25,-4.04) {(b) KV reuse: documents encoded alone, position-corrected, caches joined};
\node[anchor=east, rounded corners=2pt, fill=roD, draw=roE, line width=0.5pt, inner sep=2.2pt, font=\sffamily\tiny\bfseries, text=roF] at (15.75,-4.04) {reuse loses 19--100\% of $Q^*$};
\draw[rounded corners=3pt, fill=slB, draw=slE, line width=0.8pt] (0.25,-5.26) rectangle ++(1.5,-1.42);
\node[anchor=north west, inner sep=0pt, text width=1.30cm, align=flush left, font=\sffamily\scriptsize] at (0.35,-5.36) {\textbf{Instruction}\\[1.5pt]shared prefix};
\node[anchor=south west, inner sep=0pt, font=\sffamily\tiny\bfseries, text=slF] at (0.35,-6.59) {cached once};
\coordinate (bi) at (1.00,-5.26);
\draw[rounded corners=3pt, fill=slA, draw=slD, line width=0.6pt, dash pattern=on 3pt off 1.6pt] (1.87,-5.26) rectangle ++(0.78,-1.42);
\node[font=\sffamily\scriptsize, text=slE, align=center] at (2.26,-5.97) {Docs\\1--5};
\coordinate (bm1) at (2.26,-5.26);
\draw[rounded corners=3pt, fill=emA, draw=emB, line width=0.8pt, dash pattern=on 3pt off 1.6pt] (2.77,-5.26) rectangle ++(3.75,-1.42);
\node[anchor=north west, inner sep=0pt, text width=3.55cm, align=flush left, font=\sffamily\scriptsize] at (2.87,-5.36) {\textbf{Doc 6} \textcolor{emB}{(target)}\\[1.5pt]\textit{Dukes of Hazzard (film): \ldots{}}\\\textit{Daisy Duke} (\textbf{Jessica Simpson})};
\node[anchor=south west, inner sep=0pt, font=\sffamily\tiny\bfseries, text=roB] at (2.87,-6.59) {sees: instruction only};
\coordinate (b6) at (4.64,-5.26);
\draw[rounded corners=3pt, fill=slA, draw=slD, line width=0.6pt, dash pattern=on 3pt off 1.6pt] (6.64,-5.26) rectangle ++(0.78,-1.42);
\node[font=\sffamily\scriptsize, text=slE, align=center] at (7.03,-5.97) {Docs\\7--10};
\coordinate (bm2) at (7.03,-5.26);
\draw[rounded corners=3pt, fill=roA, draw=roB, line width=0.8pt, dash pattern=on 3pt off 1.6pt] (7.54,-5.26) rectangle ++(3.85,-1.42);
\node[anchor=north west, inner sep=0pt, text width=3.65cm, align=flush left, font=\sffamily\scriptsize] at (7.64,-5.36) {\textbf{Doc 11} \textcolor{roB}{(similar)}\\[1.5pt]\textbf{Catherine Bach} \textit{\ldots{} known for}\\\textit{playing Daisy Duke} (TV series)};
\node[anchor=south west, inner sep=0pt, font=\sffamily\tiny\bfseries, text=roB] at (7.64,-6.59) {sees: instruction only};
\coordinate (b11) at (9.47,-5.26);
\draw[rounded corners=3pt, fill=slA, draw=slD, line width=0.6pt, dash pattern=on 3pt off 1.6pt] (11.51,-5.26) rectangle ++(0.78,-1.42);
\node[font=\sffamily\scriptsize, text=slE, align=center] at (11.90,-5.97) {Docs\\12--30};
\coordinate (bm3) at (11.90,-5.26);
\draw[rounded corners=3pt, fill=blA, draw=blB, line width=0.8pt] (12.41,-5.26) rectangle ++(3.15,-1.42);
\node[anchor=north west, inner sep=0pt, text width=2.95cm, align=flush left, font=\sffamily\scriptsize] at (12.51,-5.36) {\textbf{Question}\\[1.5pt]\textit{Who played Daisy Duke in}\\\textit{the new Dukes of Hazzard?}};
\node[anchor=south west, inner sep=0pt, font=\sffamily\tiny\bfseries, text=blC] at (12.51,-6.59) {sees: all joined caches};
\coordinate (bq) at (13.99,-5.26);
\draw[line width=0.8pt, draw=slD, dash pattern=on 2.5pt off 1.8pt] (bm1) to[out=35,in=145] node[pos=0.5, fill=roG, inner sep=0.3pt, font=\sffamily\scriptsize\bfseries, text=roB]{\ding{55}} (b6);
\draw[line width=0.8pt, draw=slD, dash pattern=on 2.5pt off 1.8pt] (b6) to[out=35,in=145] node[pos=0.5, fill=roG, inner sep=0.3pt, font=\sffamily\scriptsize\bfseries, text=roB]{\ding{55}} (bm2);
\draw[line width=0.8pt, draw=slD, dash pattern=on 2.5pt off 1.8pt] (bm2) to[out=35,in=145] node[pos=0.5, fill=roG, inner sep=0.3pt, font=\sffamily\scriptsize\bfseries, text=roB]{\ding{55}} (b11);
\draw[line width=0.8pt, draw=slD, dash pattern=on 2.5pt off 1.8pt] (b11) to[out=35,in=145] node[pos=0.5, fill=roG, inner sep=0.3pt, font=\sffamily\scriptsize\bfseries, text=roB]{\ding{55}} (bm3);
\draw[-{Latex[length=1.5mm]}, line width=0.9pt, draw=blB] (bm3) to[out=35,in=145] (bq);
\node[font=\sffamily\tiny\bfseries, text=slE] at (7.08,-4.46) {no context across documents};
\node[font=\sffamily\tiny\bfseries, text=blC, anchor=south] at (13.05,-4.64) {only the question and answer read all caches};
\draw[rounded corners=2pt, fill=roD, draw=roE, line width=0.6pt] (0.25,-6.81) rectangle (15.75,-7.27);
\node[anchor=west, font=\sffamily\scriptsize, text=roF] at (0.4,-7.04) {\textbf{Output: Catherine Bach} \quad the actress named in the similar documents};
\node[anchor=east, font=\sffamily\scriptsize\bfseries, text=roF] at (15.6,-7.04) {\ding{55} wrong};
\end{tikzpicture}
\caption{The example of Figure~\ref{fig:schematic}, shown chunk by chunk. The bottom line of each box states what that chunk attends to when its keys and values are computed; under reuse, every document after the first cannot attend to the documents before it.}
\label{fig:schematic_boxes}
\end{figure*}

Figure~\ref{fig:schematic_boxes} shows the example of Figure~\ref{fig:schematic} chunk by chunk.

\paragraph{Splits and sources.} LoCoMo answers are scored by token F1 and counted as correct when F1 $\ge 0.5$. Gorilla prompts contain the user request only, and an answer is correct when it contains the ground-truth \texttt{api\_name}. Each dataset is split by document or conversation into 70 development questions and a disjoint test set of 250 questions (800 for TriviaQA, where many questions are answerable without context). For NQ-Open and Gorilla, where every model has fewer than 80 context-dependent test questions, we add a second set of 250 questions built in the same way. In every question, the slot of the target is drawn uniformly at random from all slots with a fixed seed and is the same under SIM, RAND and OTHER. Replacement chunks are chosen to match the length of the SIM chunk in the same position (within 10\% for 92--100\% of chunks). In each open-domain QA or API dataset, a normalised gold answer occurs in some RAND or OTHER chunk for at most 1.9\% of the questions, and in some SIM distractor for at most 4.6\% of the questions. SIM is the kind of context that a retriever returns; Gorilla is discussed below.

\paragraph{LoCoMo.} Each chunk is one session of a conversation, headed by its date, for example ``Session 3 (9:27 am on 7 February, 2022):''. We keep questions whose evidence lies in a single session and drop unanswerable ones. SIM uses up to 15 other sessions of the same conversation, drawn in random order among those that do not contain the answer; RAND uses sessions of other conversations. Development and test questions come from different conversations; the test set covers eight conversations.

\paragraph{NQ-Open.} We use the 30-document version of \citet{lostmiddle}: one passage that contains the answer and up to 29 passages retrieved by Contriever \citep{contriever} that do not contain it. RAND uses passages retrieved for other questions.

\paragraph{TriviaQA.} We use the DPR release \citep{dpr}, which provides one positive passage and 29 hard negatives, and we remove any passage that contains an answer alias. Because the no-context baseline answers many TriviaQA questions, the test set has 800 questions.

\paragraph{Gorilla.} Requests come from the HuggingFace part of APIBench. Each document lists an API's name, domain, functionality, description and example code. The prompt contains the user request only, without the reference API call or code. SIM uses the 30 documents whose BGE-M3 embeddings are closest to that of the request together with its reference solution, whereas the detector $d_3$ (Section~\ref{sec:attention}) and the retriever of Section~\ref{sec:alloc} embed the request alone. RAND uses documents from other task categories. Retention on Gorilla is already zero under RAND, so building SIM with the reference solution cannot create a positive similarity term. Once the user request is extracted on its own, none of the request texts contains the reference call or code, but four requests still name their API, and we exclude them. Because SIM uses the reference solution, we also build the document list from the request alone (SIM-q), keeping the same corpus, filters, number of documents and target position. For each \Qs{} question, on average 40--42\% of the SIM-q distractor documents also appear in SIM. Full prefill answers 68--78\% of \Qs{} under SIM-q; on those questions, retention under reuse is 0.01--0.03, against 0.00--0.07 under SIM and 0.00 under RAND. The paired differences from SIM are +0.01 [+0.00, +0.05] with Qwen3-8B, \textminus{}0.03 [\textminus{}0.07, +0.00] with Llama-3.1-8B and \textminus{}0.04 [\textminus{}0.10, +0.01] with Mistral-7B. The differences from RAND (+0.01 to +0.03) are significant for none of the models.

\begin{table*}[t]
\centering
\footnotesize
\begin{tabular}{llcccccccc}
\toprule
Dataset & Model & $N$ & \begin{tabular}[b]{@{}c@{}}No\\context\end{tabular} & \begin{tabular}[b]{@{}c@{}}Full\\prefill\end{tabular} & Reuse & \begin{tabular}[b]{@{}c@{}}Full $-$\\reuse\end{tabular} & \begin{tabular}[b]{@{}c@{}}Reuse $-$\\no context\end{tabular} & \begin{tabular}[b]{@{}c@{}}$N$\\middle\end{tabular} & \begin{tabular}[b]{@{}c@{}}Retention\\middle\end{tabular} \\
\midrule
LoCoMo & Qwen3-8B & 250 & 0.11 & 0.61 & 0.36 & \begin{tabular}[t]{@{}c@{}}\textbf{+0.26}\\[-1pt]{\scriptsize [+0.19, +0.32]}\end{tabular} & \begin{tabular}[t]{@{}c@{}}\textbf{+0.24}\\[-1pt]{\scriptsize [+0.18, +0.32]}\end{tabular} & 129 & 0.49 \\
 & Llama-3.1-8B & 250 & 0.10 & 0.58 & 0.50 & \begin{tabular}[t]{@{}c@{}}\textbf{+0.08}\\[-1pt]{\scriptsize [+0.04, +0.13]}\end{tabular} & \begin{tabular}[t]{@{}c@{}}\textbf{+0.40}\\[-1pt]{\scriptsize [+0.32, +0.50]}\end{tabular} & 123 & 0.74 \\
 & Mistral-7B & 250 & 0.01 & 0.20 & 0.15 & \begin{tabular}[t]{@{}c@{}}+0.04\\[-1pt]{\scriptsize [+0.00, +0.09]}\end{tabular} & \begin{tabular}[t]{@{}c@{}}\textbf{+0.14}\\[-1pt]{\scriptsize [+0.08, +0.20]}\end{tabular} & 48 & 0.60 \\
\midrule
NQ-Open & Qwen3-8B & 500 & 0.29 & 0.56 & 0.22 & \begin{tabular}[t]{@{}c@{}}\textbf{+0.34}\\[-1pt]{\scriptsize [+0.30, +0.39]}\end{tabular} & \begin{tabular}[t]{@{}c@{}}\textbf{\textminus{}0.07}\\[-1pt]{\scriptsize [\textminus{}0.11, \textminus{}0.03]}\end{tabular} & 164 & 0.15 \\
 & Llama-3.1-8B & 500 & 0.41 & 0.61 & 0.15 & \begin{tabular}[t]{@{}c@{}}\textbf{+0.46}\\[-1pt]{\scriptsize [+0.41, +0.51]}\end{tabular} & \begin{tabular}[t]{@{}c@{}}\textbf{\textminus{}0.26}\\[-1pt]{\scriptsize [\textminus{}0.30, \textminus{}0.21]}\end{tabular} & 136 & 0.15 \\
 & Mistral-7B & 500 & 0.36 & 0.53 & 0.20 & \begin{tabular}[t]{@{}c@{}}\textbf{+0.33}\\[-1pt]{\scriptsize [+0.28, +0.37]}\end{tabular} & \begin{tabular}[t]{@{}c@{}}\textbf{\textminus{}0.15}\\[-1pt]{\scriptsize [\textminus{}0.20, \textminus{}0.10]}\end{tabular} & 128 & 0.18 \\
\midrule
Gorilla & Qwen3-8B & 496 & 0.00 & 0.19 & 0.02 & \begin{tabular}[t]{@{}c@{}}\textbf{+0.18}\\[-1pt]{\scriptsize [+0.14, +0.21]}\end{tabular} & \begin{tabular}[t]{@{}c@{}}\textbf{+0.02}\\[-1pt]{\scriptsize [+0.01, +0.03]}\end{tabular} & 95 & 0.01 \\
 & Llama-3.1-8B & 496 & 0.00 & 0.23 & 0.03 & \begin{tabular}[t]{@{}c@{}}\textbf{+0.19}\\[-1pt]{\scriptsize [+0.15, +0.23]}\end{tabular} & \begin{tabular}[t]{@{}c@{}}\textbf{+0.03}\\[-1pt]{\scriptsize [+0.02, +0.05]}\end{tabular} & 112 & 0.04 \\
 & Mistral-7B & 496 & 0.00 & 0.30 & 0.02 & \begin{tabular}[t]{@{}c@{}}\textbf{+0.28}\\[-1pt]{\scriptsize [+0.24, +0.32]}\end{tabular} & \begin{tabular}[t]{@{}c@{}}\textbf{+0.02}\\[-1pt]{\scriptsize [+0.01, +0.03]}\end{tabular} & 147 & 0.05 \\
\midrule
TriviaQA & Qwen3-8B & 800 & 0.63 & 0.75 & 0.39 & \begin{tabular}[t]{@{}c@{}}\textbf{+0.36}\\[-1pt]{\scriptsize [+0.33, +0.40]}\end{tabular} & \begin{tabular}[t]{@{}c@{}}\textbf{\textminus{}0.24}\\[-1pt]{\scriptsize [\textminus{}0.28, \textminus{}0.20]}\end{tabular} & 168 & 0.25 \\
 & Llama-3.1-8B & 800 & 0.72 & 0.79 & 0.33 & \begin{tabular}[t]{@{}c@{}}\textbf{+0.45}\\[-1pt]{\scriptsize [+0.42, +0.49]}\end{tabular} & \begin{tabular}[t]{@{}c@{}}\textbf{\textminus{}0.39}\\[-1pt]{\scriptsize [\textminus{}0.42, \textminus{}0.35]}\end{tabular} & 138 & 0.20 \\
 & Mistral-7B & 800 & 0.68 & 0.72 & 0.36 & \begin{tabular}[t]{@{}c@{}}\textbf{+0.36}\\[-1pt]{\scriptsize [+0.33, +0.40]}\end{tabular} & \begin{tabular}[t]{@{}c@{}}\textbf{\textminus{}0.33}\\[-1pt]{\scriptsize [\textminus{}0.36, \textminus{}0.29]}\end{tabular} & 130 & 0.25 \\
\bottomrule
\end{tabular}
\caption{Accuracy on every evaluated question under SIM (the Questions column of Table~\ref{tab:funnel}): without context, with full prefill and with position-corrected reuse; the differences have paired 95\% bootstrap intervals. $N$ middle: the number of questions answered wrongly without context and correctly with full prefill under SIM, before the RAND and OTHER conditions of \Qs{} are applied. Retention middle: the share of these questions that reuse still answers correctly.}
\label{tab:allq}
\end{table*}

\paragraph{All evaluated questions.} Table~\ref{tab:allq} gives accuracy on all evaluated questions. On open-domain QA, reuse lowers accuracy by 0.33--0.46 relative to full prefill and answers significantly fewer questions than the model without context in all six settings, because it also loses 47--75\% of the questions that the model answers without context. On LoCoMo and Gorilla, where the no-context baseline is low, reuse stays above it.

\paragraph{Context-dependent subset.} Table~\ref{tab:funnel} shows how the evaluated questions are narrowed down to \Qs{}. Retention on \Qs{} is the share of these questions that reuse still answers correctly.

\begin{table*}[t]
\centering
\footnotesize
\begin{tabular}{llcccc}
\toprule
Dataset & Model & Questions & \begin{tabular}[b]{@{}c@{}}No context\\wrong\end{tabular} & \begin{tabular}[b]{@{}c@{}}+ full prefill\\right (SIM)\end{tabular} & \begin{tabular}[b]{@{}c@{}}+ right under\\RAND, OTHER (\Qs{})\end{tabular} \\
\midrule
LoCoMo & Qwen3-8B & 250 & 222 & 129 & 105 \\
 & Llama-3.1-8B & 250 & 225 & 123 & 102 \\
 & Mistral-7B & 250 & 247 & 48 & 36 \\
\midrule
NQ-Open & Qwen3-8B & 500 & 354 & 164 & 146 \\
 & Llama-3.1-8B & 500 & 296 & 136 & 128 \\
 & Mistral-7B & 500 & 322 & 128 & 101 \\
\midrule
Gorilla & Qwen3-8B & 496 & 496 & 95 & 89 \\
 & Llama-3.1-8B & 496 & 496 & 112 & 100 \\
 & Mistral-7B & 496 & 496 & 147 & 108 \\
\midrule
TriviaQA & Qwen3-8B & 800 & 299 & 168 & 151 \\
 & Llama-3.1-8B & 800 & 226 & 138 & 127 \\
 & Mistral-7B & 800 & 254 & 130 & 117 \\
\bottomrule
\end{tabular}
\caption{Number of questions left after each condition, from the evaluated questions to \Qs{} (test set and, where used, the second question set).}
\label{tab:funnel}
\end{table*}

\paragraph{Prompts.} The user message consists of the instruction, the chunks separated by blank lines, and the question. It is wrapped in each model's chat template with the generation prompt, and Qwen3 runs with thinking disabled. The instructions are:
\begin{itemize}\setlength{\itemsep}{1pt}
\item LoCoMo: ``Answer the question based on the following conversation sessions. Only provide the short, concise answer without any other words.''
\item NQ-Open and TriviaQA: ``Write a high-quality answer for the given question using only the provided search results (some of which might be irrelevant). Only output the concise answer without any other words.''
\item Gorilla: ``Below is a list of API documentations. Select and output the exact api\_name that best matches the instruction.''
\end{itemize}
The question follows as ``Question: \ldots\ Answer:'' (LoCoMo, NQ-Open, TriviaQA) or ``Instruction: \ldots\ Answer with api\_name:'' (Gorilla). Passages are headed ``Document [1] (Title: \ldots):'', ``Document [2] (Title: \ldots):'' and so on, and API documents are headed ``API Document [1]:'', ``API Document [2]:'' and so on. The number in square brackets is the position of the chunk in the prompt, counting the target chunk.

\section{Implementation and Scoring}
\label{app:impl}

\paragraph{Reuse.} The shared instruction prefix is encoded once. Each chunk is encoded directly after the prefix, its keys are re-rotated to the chunk's position in the full prompt, and the caches are concatenated; the question tokens and the answer are then processed against this cache. Full prefill encodes the same tokens in one pass. When no chunk is replaced, the cache of the prefix is the same under reuse and under full prefill. Surrogate repairs encode each chunk after the prefix and a 512- or 1,024-token surrogate, and then discard the surrogate's KV. SurroKV full is our re-implementation of the full method from its description in \citet{surrokv}: it uses the 1,024-token Wikipedia surrogate and then recomputes the 10\% of tokens ranked highest by question attention. Online methods and chunk-level allocations recompute the selected token positions through all layers. Each recomputed token attends to the prefix and to all earlier positions of the current cache (reused, or already recomputed in this pass); the KV of later chunks is not updated; and the question and answer tokens then attend to the whole cache. Our CacheBlend implementation selects the tokens to recompute by the deviation of their keys and values at the second layer, a one-step version of the gradual filtering of \citet{cacheblend}. Budgets are fractions of the context tokens.

\paragraph{Scoring.} Each output has at most 32 new tokens (64 on Gorilla). LoCoMo is scored by the token F1 of the first output line against the gold answer; NQ-Open and TriviaQA by whether a normalised gold answer occurs in the normalised first output line; and Gorilla by whether the gold \texttt{api\_name} occurs in the output. Answers on LoCoMo, NQ-Open and TriviaQA are scored on their first line, as in the original setups. When Qwen3-8B begins with a copied document header, the answer is therefore scored wrong even if the correct answer follows; this happens on at most 4.6\% of the questions in any condition. All scores are binarised at 0.5 before \Qs{}, retention and every difference are computed. If the \texttt{api\_name} must match exactly instead of only being contained in the output, the same-category term on Gorilla remains significant for three models, and the similarity term is significantly positive for none of them.

\paragraph{Difference from the meaningful subset.} \Qs{} differs from the meaningful subset of \citet{cestola2026} in three ways: correctness is a binarised score rather than F1 $>0$ for full prefill and F1 $\le 0.2$ without context; full prefill must be correct under all three sources; and we do not drop yes/no or either-or questions. Yes/no questions account for at most 2 LoCoMo questions per model, and none of the questions in the other datasets are yes/no questions.

\paragraph{Token-level selectors.} The selector chosen on development questions is ProphetKV for LoCoMo with Qwen3-8B and Llama-3.1-8B, for NQ-Open, and for TriviaQA with Qwen3-8B and Llama-3.1-8B; question attention for LoCoMo and TriviaQA with Mistral-7B; EPIC for Gorilla with Qwen3-8B and Mistral-7B; and CacheBlend for Gorilla with Llama-3.1-8B. Question attention scores each chunk token by the attention that the question tokens place on it over the reused cache: in every layer, this attention is averaged over heads and question tokens and normalised over the chunk tokens, and the scores are then averaged over all layers. In the Token 10\% column of Table~\ref{tab:repair}, EPIC recomputes the first $\lceil b\,n_i\rceil$ tokens of every chunk $i$ of $n_i$ tokens at budget $b$ (the original EPIC recomputes a fixed number of initial tokens of every chunk except the first), and CacheBlend selects tokens by their KV deviation. The allocation tables instead apply the setting's ranking rule to all chunks jointly. For EPIC, the tokens of all chunks are ranked by their offset within their chunk, so every chunk receives the same number of initial tokens and the rest of the budget goes to the earliest chunks. For Gorilla with Llama-3.1-8B, question attention is used in place of CacheBlend. The token-level baselines of all three Gorilla settings therefore differ between Table~\ref{tab:repair} and the allocation tables. The in-domain pad of Table~\ref{tab:repair}, the text placed before every chunk, is built from the first SIM chunks of two development questions and cut to 512 tokens.

\paragraph{Allocation rule.} Within the chosen chunks, tokens are ranked by the setting's token-level rule (EPIC's first-token rule, ProphetKV or question attention), and the token-level baseline applies the same rule to all chunks, so that an allocation differs from its baseline only in which chunks may be recomputed. CacheBlend selects tokens by comparing recomputed and reused KV and has no within-chunk form; in the one setting where it is the best selector on development questions (Gorilla with Llama-3.1-8B), question attention is used instead for the baseline and the allocations. When the chosen chunks hold fewer tokens than the budget, all their tokens are recomputed.

\paragraph{Statistics.} Intervals are percentile intervals of 2,000 paired bootstrap resamples with a fixed seed. A resample draws questions (conversations for LoCoMo) with replacement and is shared by all conditions of a setting and by every table that reports the same questions. For the competing-answer comparison (Table~\ref{tab:competing}), the same resample of questions is used for SIM and RAND. The abstention comparison (Table~\ref{tab:abstain}) contrasts two different sets of questions and resamples them independently. Intervals are reported per comparison without correction for multiple comparisons.

\paragraph{Holm's correction.} We also apply Holm's correction within each family of settings whose count the text reports, with two-sided $p$-values computed from the bootstrap standard error with a normal approximation. Table~\ref{tab:holm} lists the 23 families, which contain 196 comparisons in total, with their counts before and after the correction. The counts of 13 families do not change.

\begin{table*}[t]
\centering
\footnotesize
\begin{tabular}{@{}llccc@{}}
\toprule
Comparison & Reported in & $m$ & Uncorrected & Holm \\
\midrule
Loss, $1-R(\mathrm{SIM})$ & Table~\ref{tab:main_decomposition} & 12 & 12\,/\,0 & 12\,/\,0 \\
Same-category term & Table~\ref{tab:main_decomposition} & 12 & 9\,/\,0 & 9\,/\,0 \\
Similarity term, open-domain QA & Table~\ref{tab:main_decomposition} & 6 & 5\,/\,0 & 5\,/\,0 \\
Competing answers, SIM $-$ RAND & Table~\ref{tab:competing} & 6 & 6\,/\,0 & 6\,/\,0 \\
All distractors replaced $-$ target replaced & Table~\ref{tab:replacement} & 12 & 11\,/\,0 & 11\,/\,0 \\
Reuse $-$ no context, all questions, open-domain QA & Table~\ref{tab:allq} & 6 & 0\,/\,6 & 0\,/\,6 \\
Target $-$ token-level selector, 5\% & Table~\ref{tab:alloc} & 9 & 9\,/\,0 & 9\,/\,0 \\
Target $-$ token-level selector, 10\% & Table~\ref{tab:alloc} & 9 & 4\,/\,0 & 4\,/\,0 \\
Retrieved $-$ token-level selector, 5\% & Table~\ref{tab:alloc} & 9 & 5\,/\,0 & 3\,/\,0 \\
Retrieved+4 $-$ token-level selector, 5\% & Table~\ref{tab:alloc} & 9 & 7\,/\,0 & 5\,/\,0 \\
Recomputed target $-$ inserted full-prefill KV & Tables~\ref{tab:compat}, \ref{tab:compat2} & 7 & 7\,/\,0 & 4\,/\,0 \\
Moved chunk $-$ token-level selector & Table~\ref{tab:reorder} & 9 & 9\,/\,0 & 9\,/\,0 \\
Moved chunk $-$ the same chunk recomputed in place & Table~\ref{tab:reorder} & 9 & 9\,/\,0 & 9\,/\,0 \\
Selector on the reordered $-$ original prompt & Table~\ref{tab:reorder_tok} & 9 & 7\,/\,0 & 7\,/\,0 \\
Moved chunk $-$ selected tokens, reordered prompt & Table~\ref{tab:reorder_tok} & 9 & 4\,/\,0 & 3\,/\,0 \\
Moved chunk $-$ InfoFlow-style reordering, 5\% & Table~\ref{tab:infoflow} & 9 & 6\,/\,0 & 5\,/\,0 \\
Moved chunk at 5\% $-$ InfoFlow-style reordering at 15\% & Table~\ref{tab:infoflow} & 9 & 1\,/\,2 & 1\,/\,1 \\
Three moved chunks $-$ one moved chunk, 10\% & Table~\ref{tab:topk} & 9 & 7\,/\,0 & 6\,/\,0 \\
Three moved chunks $-$ selected tokens, 10\% & Table~\ref{tab:topk} & 9 & 3\,/\,2 & 3\,/\,2 \\
One moved chunk $-$ selected tokens, 10\% & Appendix~\ref{app:topk} & 9 & 2\,/\,4 & 0\,/\,1 \\
Gap closed by removing competing entities & Table~\ref{tab:compctl} & 6 & 4\,/\,0 & 3\,/\,0 \\
Gap closed by removing other entities & Table~\ref{tab:compctl} & 6 & 1\,/\,0 & 1\,/\,0 \\
Competing $-$ other entities removed & Table~\ref{tab:compctl} & 6 & 3\,/\,0 & 2\,/\,0 \\
\bottomrule
\end{tabular}
\caption{Families of settings whose count of significant differences the text reports. $m$: number of settings in the family. Uncorrected and Holm: settings in which the difference is significantly positive\,/\,significantly negative, before and after Holm's correction within the family. Each difference is the first condition named minus the second.}
\label{tab:holm}
\end{table*}

\paragraph{Leaving out conversations.} On LoCoMo, we also leave out each of the eight conversations in turn and recompute the bootstrap intervals. Whichever conversation is left out, 10 of the 12 terms of Table~\ref{tab:main_decomposition} keep their significance, and the other two change in at most one of these analyses. None of the six comparisons of Target with the selector at 5\% and 10\% becomes significant (point estimates \textminus{}0.08 to +0.07), and the loss from recomputing the retrieved chunk for Qwen3-8B stays significant in all eight analyses. Against a ProphetKV selector, the gain of Target stays significant in eight of the eight analyses for Qwen3-14B (+0.11 to +0.16), but in only five of them for Mistral-7B (+0.06 to +0.13).

\paragraph{Determinism.} Running the first 60 NQ-Open test questions a second time with Qwen3-8B reproduced all 787 generated outputs exactly. The LoCoMo rows of Table~\ref{tab:alloc} and Tables~\ref{tab:allq}, \ref{tab:ksweep}, \ref{tab:uniform}, \ref{tab:reorder}--\ref{tab:topk}, \ref{tab:compat} and~\ref{tab:compat2} come from separate runs. Wherever one of these runs repeats a condition of the main experiments or of another of these runs, we compared the two on every shared question. The scores are identical, and so are the generated outputs wherever both runs store them, except in Table~\ref{tab:uniform} when the chosen chunks exceed the budget and their tokens are ranked by a different rule than in the main experiments.

\paragraph{Hardware and software.} Model runs use bfloat16 weights and greedy decoding. Each process runs on one NVIDIA RTX 4090 D (24 GB), except that Qwen3-14B uses two of these GPUs and Qwen3-32B uses one NVIDIA RTX PRO 6000 (96 GB). Most runs use PyTorch 2.8 and Transformers 4.56. A few supplementary experiments ran on a second server with the same GPUs, PyTorch 2.7 and Transformers 4.52.

\Needspace{8\baselineskip}
\section{Proposition~\ref{prop:layer}}
\label{app:prop}

\paragraph{Proof.} Fix a head, write $\ell_{uv}$ for the attention logit of $u$ on key $v$ under full prefill, and $Z_A=\sum_{v\in A}e^{\ell_{uv}}$ for a set of keys $A$. The keys that $u$ sees under full prefill split into $A=C_0\cup C_i^{\le u}$, where $C_i^{\le u}$ are the tokens of $C_i$ up to $u$, and $B=C_{<i}$. With $\alpha_u=Z_B/(Z_A+Z_B)$,
\begin{align*}
o^{\mathrm{full}}_u&=\frac{Z_A\,o^{\mathrm{own}}_u+Z_B\,o^{\mathrm{cross}}_u}{Z_A+Z_B}\\
&=(1-\alpha_u)\,o^{\mathrm{own}}_u+\alpha_u\,o^{\mathrm{cross}}_u,
\end{align*}
and subtracting $o^{\mathrm{reuse}}_u$ gives Equation~\ref{eq:layer}. Under reuse, $u$ sees the keys of $A$ only. If the hidden states are the same, the queries, keys and values coincide up to the rotary position. Values carry no rotation, and logits depend on positions only through their difference. Within $C_i$ the differences are the same under both encodings, so the logits on $C_i^{\le u}$ agree. On $C_0$ they differ, because reuse shortens the distance between $u$ and every prefix token by $s_i-P$. The second term of Equation~\ref{eq:layer} therefore vanishes when $s_i=P$. For $i=1$ we also have $B=\emptyset$, so the two encodings give the same output for every token of $C_1$ in the first layer. By induction over layers, the hidden states, keys and values of $C_1$ then agree in every layer. In the first layer every token enters with its embedding, so the assumption of equal hidden states holds for all tokens. In later layers the hidden states of $C_{<i}$ and $C_i$ differ between the two encodings, and the second term of Equation~\ref{eq:layer} is then no longer only a position effect.

\begin{table*}[t]
\centering
\footnotesize
\setlength{\tabcolsep}{3pt}
\begin{tabular}{llcccccccc}
\toprule
Dataset & Model & $N$ & \begin{tabular}[b]{@{}c@{}}$\bar\alpha$\\2nd chunk\end{tabular} & \begin{tabular}[b]{@{}c@{}}$\bar\alpha$\\first third\end{tabular} & \begin{tabular}[b]{@{}c@{}}$\bar\alpha$\\last third\end{tabular} & \begin{tabular}[b]{@{}c@{}}$\bar\alpha$\\last chunk\end{tabular} & \begin{tabular}[b]{@{}c@{}}Spearman\\slot, $\bar\alpha$\end{tabular} & \begin{tabular}[b]{@{}c@{}}Position term\\/ cross term\end{tabular} & \begin{tabular}[b]{@{}c@{}}Max.\\residual\end{tabular} \\
\midrule
NQ-Open & Qwen3-8B & 30 & 0.19 & 0.27 & 0.36 & 0.36 & 0.99 & 0.45 & 2e-7 \\
 & Llama-3.1-8B & 30 & 0.24 & 0.40 & 0.59 & 0.60 & 1.00 & 0.63 & 1e-7 \\
 & Mistral-7B & 30 & 0.21 & 0.38 & 0.60 & 0.62 & 1.00 & 0.38 & 5e-7 \\
\midrule
Gorilla & Qwen3-8B & 30 & 0.17 & 0.26 & 0.34 & 0.34 & 0.94 & 0.39 & 2e-7 \\
 & Llama-3.1-8B & 30 & 0.20 & 0.39 & 0.58 & 0.58 & 0.98 & 0.61 & 2e-7 \\
 & Mistral-7B & 30 & 0.21 & 0.38 & 0.58 & 0.60 & 1.00 & 0.31 & 6e-7 \\
\midrule
TriviaQA & Qwen3-8B & 30 & 0.19 & 0.27 & 0.36 & 0.36 & 1.00 & 0.45 & 2e-7 \\
 & Llama-3.1-8B & 30 & 0.24 & 0.40 & 0.59 & 0.60 & 1.00 & 0.63 & 2e-7 \\
 & Mistral-7B & 30 & 0.21 & 0.38 & 0.60 & 0.62 & 1.00 & 0.37 & 4e-7 \\
\bottomrule
\end{tabular}
\caption{First-layer decomposition of Proposition~\ref{prop:layer} under SIM, on the first 30 \Qs{} questions of each setting (CPU, float32). $\bar\alpha$: full-prefill attention mass on the earlier chunks, averaged over heads and the tokens of a chunk, then over questions, for the second chunk, the chunks in the first and last thirds of the prompt, and the last chunk. Spearman: rank correlation between the slot of a chunk and its $\bar\alpha$. Position term / cross term: the ratio of the norm of $o^{\mathrm{own}}_u-o^{\mathrm{reuse}}_u$ to the norm of $\alpha_u(o^{\mathrm{cross}}_u-o^{\mathrm{own}}_u)$, averaged over the chunks after the first. Residual: largest absolute difference between the two sides of the identity.}
\label{tab:layer0}
\end{table*}

\paragraph{First-layer measurement.} For Table~\ref{tab:layer0}, on the first 30 \Qs{} questions of each open-domain QA and API setting (ordered by identifier), we compute both sides of Equation~\ref{eq:layer} in the first layer for every head and every chunk token under SIM, in float32 on CPU. The largest difference between the two sides is $6\times 10^{-{}7}$. Both terms are zero for the first chunk in every question. The attention mass $\bar\alpha$ that a chunk places on the earlier chunks rises with the chunk's position in all nine settings (Spearman correlation 0.94--1.00), from 0.17--0.24 for the second chunk to 0.34--0.62 for the last. The position term is 0.31--0.63 times the size of the cross-chunk term, and the two terms partly cancel.

\begin{table*}[t]
\centering
\footnotesize
\setlength{\tabcolsep}{2.5pt}
\begin{tabular}{llccccccccccc}
\toprule
 & & & \multicolumn{6}{c}{Measured $R(k)$} & \multicolumn{3}{c}{Predicted $R(16)$} & \\ \cmidrule(lr){4-9}\cmidrule(lr){10-12}
Dataset & Model & $N$ & $k=0$ & 4 & 8 & 16 & 24 & $n$ & Odds & Probit & Linear & \begin{tabular}[b]{@{}c@{}}Most similar\\first, 16\end{tabular} \\
\midrule
NQ-Open & Qwen3-8B & 146 & 0.15 & 0.19 & 0.23 & 0.32 [+0.25, +0.40] & 0.51 & 0.71 & 0.27 & 0.44 & 0.46 & 0.27 \\
 & Llama-3.1-8B & 128 & 0.16 & 0.20 & 0.20 & 0.30 [+0.22, +0.38] & 0.55 & 0.70 & 0.28 & 0.44 & 0.46 & 0.29 \\
 & Mistral-7B & 101 & 0.18 & 0.25 & 0.29 & 0.32 [+0.23, +0.42] & 0.49 & 0.66 & 0.30 & 0.43 & 0.45 & 0.29 \\
\midrule
TriviaQA & Qwen3-8B & 151 & 0.27 & 0.31 & 0.30 & 0.43 [+0.35, +0.51] & 0.60 & 0.80 & 0.43 & 0.58 & 0.56 & 0.40 \\
 & Llama-3.1-8B & 127 & 0.20 & 0.29 & 0.35 & 0.43 [+0.34, +0.51] & 0.60 & 0.74 & 0.34 & 0.49 & 0.50 & 0.46 \\
 & Mistral-7B & 117 & 0.25 & 0.27 & 0.31 & 0.35 [+0.26, +0.44] & 0.50 & 0.71 & 0.39 & 0.50 & 0.50 & 0.31 \\
\bottomrule
\end{tabular}
\caption{Replacing the KV of $k$ distractors with their full-prefill KV under SIM, with the target left reused and the distractors taken in a random order fixed per question. The columns $k=0$ and $k=n$ (every distractor replaced) come from the main experiments. Predicted $R(16)$ is computed from $R(0)$ and $R(n)$ alone in three ways: additive odds of failure (Equation~\ref{eq:odds}), linear in the probit of $R$, and linear in $R$. Last column: retention when the 16 distractors most similar to the target are replaced first.}
\label{tab:ksweep}
\end{table*}

\begin{figure*}[t]
\centering
\includegraphics[width=\textwidth]{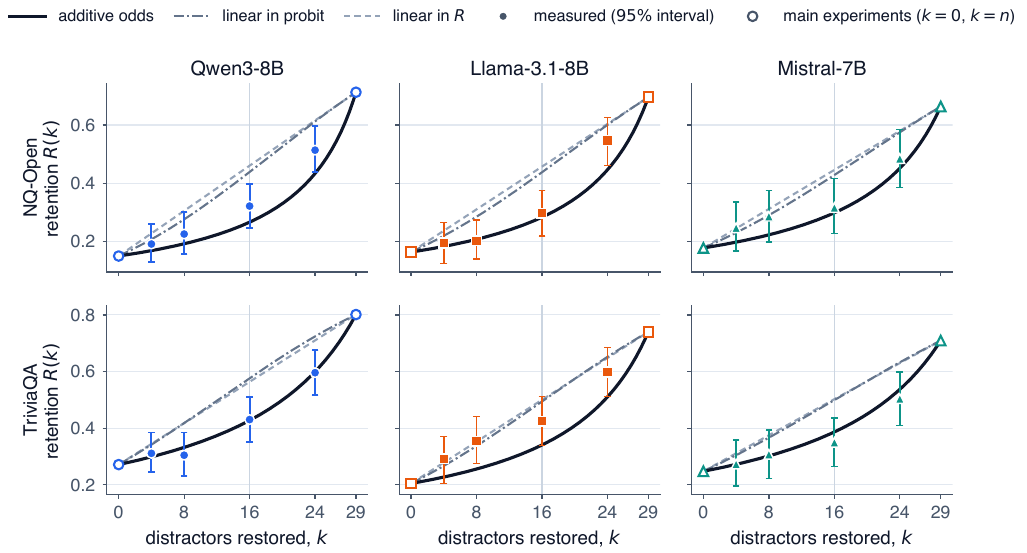}
\caption{Retention after restoring the full-prefill KV of $k$ random distractors under SIM. Filled markers show the measured retention with 95\% paired bootstrap intervals. Open markers show the two end points, $k=0$ and $k=n$ (every distractor restored), taken from the main experiments. The lines show the three predictions made from these two end points: additive odds of failure (Equation~\ref{eq:odds}), linear in the probit of $R$, and linear in $R$. The vertical line marks $k=16$, where the predictions differ most and are tested (Table~\ref{tab:ksweep}).}
\label{fig:ksweep}
\end{figure*}

\paragraph{Restoring $k$ distractors.} Figure~\ref{fig:ksweep} plots the results of Table~\ref{tab:ksweep}. All three predictions use only $R(0)$ and $R(n)$ from the main experiments, and the odds prediction applies Equation~\ref{eq:odds} to the retention of a setting rather than to each question. For each question, the distractors are put in a random order fixed by a seed and the first $k$ are restored, so the restored sets are nested; the target chunk itself stays reused. Over $k=4$ to 24, the odds model has the smallest squared error in five of the six settings, and summed over them its squared error is 0.06, against 0.18 for the probit model and 0.20 for the linear model. At $k=24$, the odds prediction falls below the interval of the measured retention in three of them, and the measured retention exceeds it by up to 0.10, so the curve is convex as predicted but rises faster near the end. Restoring the 16 distractors most similar to the target first, instead of 16 random distractors, changes retention by \textminus{}0.05 to +0.03, and the change is never significant. Gorilla is left out because $R(0)$ is 0.00--0.06 there, and at such values the odds are undefined or extreme.

\Needspace{8\baselineskip}
\section{Models Fine-Tuned for Chunked Encoding}
\label{app:blockft}

\begin{table*}[t]
\centering
\footnotesize
\setlength{\tabcolsep}{1.6pt}
\begin{tabular}{lllccccccc}
\toprule
 & & & \multicolumn{3}{c}{Retention $R$} & & & & \\ \cmidrule(lr){4-6}
Dataset & Model ($N$) & Reuse & OTHER & RAND & SIM & Loss & Universal & \begin{tabular}[b]{@{}c@{}}Same-\\category\end{tabular} & Similarity \\
\midrule
NQ-Open & Tulu3-RAG (83) & Pos.-corr. & 0.39 & 0.40 & 0.18 & \begin{tabular}[t]{@{}c@{}}\textbf{+0.82}\\[-1pt]{\scriptsize [+0.73, +0.90]}\end{tabular} & \begin{tabular}[t]{@{}c@{}}\textbf{+0.61}\\[-1pt]{\scriptsize [+0.51, +0.71]}\end{tabular} & \begin{tabular}[t]{@{}c@{}}\textminus{}0.01\\[-1pt]{\scriptsize [\textminus{}0.13, +0.10]}\end{tabular} & \begin{tabular}[t]{@{}c@{}}\textbf{+0.22}\\[-1pt]{\scriptsize [+0.08, +0.35]}\end{tabular} \\
 &  & Block & 0.07 & 0.08 & 0.10 & \begin{tabular}[t]{@{}c@{}}\textbf{+0.90}\\[-1pt]{\scriptsize [+0.83, +0.96]}\end{tabular} & \begin{tabular}[t]{@{}c@{}}\textbf{+0.93}\\[-1pt]{\scriptsize [+0.87, +0.98]}\end{tabular} & \begin{tabular}[t]{@{}c@{}}\textminus{}0.01\\[-1pt]{\scriptsize [\textminus{}0.08, +0.06]}\end{tabular} & \begin{tabular}[t]{@{}c@{}}\textminus{}0.01\\[-1pt]{\scriptsize [\textminus{}0.10, +0.06]}\end{tabular} \\
 & Tulu3-Block-FT (111) & Pos.-corr. & 0.74 & 0.60 & 0.20 & \begin{tabular}[t]{@{}c@{}}\textbf{+0.80}\\[-1pt]{\scriptsize [+0.73, +0.87]}\end{tabular} & \begin{tabular}[t]{@{}c@{}}\textbf{+0.26}\\[-1pt]{\scriptsize [+0.18, +0.34]}\end{tabular} & \begin{tabular}[t]{@{}c@{}}\textbf{+0.14}\\[-1pt]{\scriptsize [+0.03, +0.24]}\end{tabular} & \begin{tabular}[t]{@{}c@{}}\textbf{+0.41}\\[-1pt]{\scriptsize [+0.31, +0.51]}\end{tabular} \\
 &  & Block & 0.77 & 0.79 & 0.40 & \begin{tabular}[t]{@{}c@{}}\textbf{+0.60}\\[-1pt]{\scriptsize [+0.51, +0.69]}\end{tabular} & \begin{tabular}[t]{@{}c@{}}\textbf{+0.23}\\[-1pt]{\scriptsize [+0.16, +0.32]}\end{tabular} & \begin{tabular}[t]{@{}c@{}}\textminus{}0.03\\[-1pt]{\scriptsize [\textminus{}0.12, +0.05]}\end{tabular} & \begin{tabular}[t]{@{}c@{}}\textbf{+0.40}\\[-1pt]{\scriptsize [+0.29, +0.50]}\end{tabular} \\
\midrule
Gorilla & Tulu3-RAG (102) & Pos.-corr. & 0.44 & 0.00 & 0.04 & \begin{tabular}[t]{@{}c@{}}\textbf{+0.96}\\[-1pt]{\scriptsize [+0.92, +0.99]}\end{tabular} & \begin{tabular}[t]{@{}c@{}}\textbf{+0.56}\\[-1pt]{\scriptsize [+0.46, +0.66]}\end{tabular} & \begin{tabular}[t]{@{}c@{}}\textbf{+0.44}\\[-1pt]{\scriptsize [+0.34, +0.54]}\end{tabular} & \begin{tabular}[t]{@{}c@{}}\textminus{}0.04$^\dagger$\\[-1pt]{\scriptsize [\textminus{}0.08, \textminus{}0.01]}\end{tabular} \\
 &  & Block & 0.03 & 0.00 & 0.01 & \begin{tabular}[t]{@{}c@{}}\textbf{+0.99}\\[-1pt]{\scriptsize [+0.97, +1.00]}\end{tabular} & \begin{tabular}[t]{@{}c@{}}\textbf{+0.97}\\[-1pt]{\scriptsize [+0.93, +1.00]}\end{tabular} & \begin{tabular}[t]{@{}c@{}}+0.03\\[-1pt]{\scriptsize [+0.00, +0.07]}\end{tabular} & \begin{tabular}[t]{@{}c@{}}\textminus{}0.01$^\dagger$\\[-1pt]{\scriptsize [\textminus{}0.03, +0.00]}\end{tabular} \\
 & Tulu3-Block-FT (102) & Pos.-corr. & 0.76 & 0.00 & 0.06 & \begin{tabular}[t]{@{}c@{}}\textbf{+0.94}\\[-1pt]{\scriptsize [+0.89, +0.98]}\end{tabular} & \begin{tabular}[t]{@{}c@{}}\textbf{+0.24}\\[-1pt]{\scriptsize [+0.16, +0.32]}\end{tabular} & \begin{tabular}[t]{@{}c@{}}\textbf{+0.76}\\[-1pt]{\scriptsize [+0.68, +0.84]}\end{tabular} & \begin{tabular}[t]{@{}c@{}}\textminus{}0.06$^\dagger$\\[-1pt]{\scriptsize [\textminus{}0.11, \textminus{}0.02]}\end{tabular} \\
 &  & Block & 0.45 & 0.02 & 0.14 & \begin{tabular}[t]{@{}c@{}}\textbf{+0.86}\\[-1pt]{\scriptsize [+0.79, +0.92]}\end{tabular} & \begin{tabular}[t]{@{}c@{}}\textbf{+0.55}\\[-1pt]{\scriptsize [+0.45, +0.65]}\end{tabular} & \begin{tabular}[t]{@{}c@{}}\textbf{+0.43}\\[-1pt]{\scriptsize [+0.34, +0.53]}\end{tabular} & \begin{tabular}[t]{@{}c@{}}\textbf{\textminus{}0.12}\\[-1pt]{\scriptsize [\textminus{}0.20, \textminus{}0.05]}\end{tabular} \\
\bottomrule
\end{tabular}
\caption{Models fine-tuned by \citet{blockattn} on the same RAG data, one with full attention (Tulu3-RAG) and one with block attention (Tulu3-Block-FT), both starting from Tulu3-SFT (Llama-3.1-8B). Each model is evaluated on its own \Qs{} with our prompts, which use the Tulu chat tags without BOS, as in its training data. Pos.-corr.: position-corrected reuse, with each chunk encoded after the prefix as in the rest of the paper. Block: each chunk encoded alone from position 0, as in the released inference code. Retention $R$: $R(\mathrm{OTHER})$, $R(\mathrm{RAND})$ and $R(\mathrm{SIM})$, the retention under each distractor source. Terms of Equation~\ref{eq:decomp} with 95\% paired bootstrap intervals. Bold: the interval excludes zero. $^\dagger$Retention under RAND is zero, so the term cannot be positive; these intervals are not set in bold.}
\label{tab:blockft}
\end{table*}

\paragraph{Results.} The Tulu3 model that \citet{blockattn} fine-tune to encode documents as independent blocks loses 0.60 of its \Qs{} questions on NQ-Open in block mode, compared with 0.82 for position-corrected reuse with its full-attention counterpart. Its universal term falls to 0.23 (0.61 for the counterpart), but its similarity term is +0.40 [+0.29, +0.50] (Table~\ref{tab:blockft}, which also covers Gorilla), so the similarity term accounts for most of its loss on NQ-Open.

\paragraph{Models and prompts.} For Table~\ref{tab:blockft} we use the released Tulu3-RAG and Tulu3-Block-FT checkpoints of \citet{blockattn} and their inference procedure for block mode, in which each block is encoded alone from position 0 without BOS, its rotation is removed and reapplied at its final position, and the question attends to all blocks. Our instruction, chunks and question are wrapped in the Tulu tags without BOS, as in the models' training data. The settings use the NQ-Open and Gorilla test sets and second question sets, with the distractor sources of Section~\ref{sec:setup}. None of our NQ-Open questions occurs in their training data (0 matches among the questions extracted from its RAG part). The two models are evaluated on their own \Qs{} and are not compared question by question.

\paragraph{Prompt format and number of documents.} We also rebuild the prompts of the 111 NQ-Open \Qs{} questions of Tulu3-Block-FT in the format of its training data (their instruction, documents as ``- Title:'' entries, their question block), once with the same 30 documents and once with the 10-document version of the same prompt. With the 30-document and the 10-document prompt, full prefill answers 68\% and 67\% of these questions, respectively, and block mode answers 35\% and 50\%; the gaps are 0.32 [+0.21, +0.44] and 0.16 [+0.05, +0.27]. Both gaps are significant, and the 10-document gap is half the 30-document one: the loss persists when the prompt follows the training format, and it shrinks with fewer documents.

\Needspace{8\baselineskip}
\section{Additional Related Work}
\label{app:related}

\paragraph{Repair methods.} CacheBlend \citep{cacheblend} recomputes the tokens whose KV deviates most, EPIC \citep{epic} recomputes the first tokens of each chunk, ProphetKV \citep{prophetkv} selects tokens by question attention, and CacheTune \citep{cachetune} selects them offline from a frequency-domain analysis of each chunk; Table~\ref{tab:assume} compares their design choices. C$^2$KV \citep{c2kv} learns compressed caches that can be concatenated at new positions. Budgeted cache repair (BCR; \citealp{bcr}) ranks cache rows by the attention of two draft tokens. On multi-agent GSM8K, its task accuracy does not differ significantly between repairing 64-token slices and repairing single rows (Table~\ref{tab:assume} gives its attention-output errors). In the diagnostic of SAFE-KV \citep{safekv} on 2Wiki-100, copying full-prefill states into the selected document alone keeps 38.0\% of the full-prefill answers, because the documents after it keep states that were computed without it.
\paragraph{Other work.} \citet{cestola2026} show that aggregate scores overstate how well reuse preserves full-prefill accuracy, and they evaluate on a meaningful subset of questions that full prefill answers, that need the context and whose answers are hard to guess; our \Qs{} follows this design (Section~\ref{sec:setup}). KVCOMM \citep{kvcomm} corrects the offsets of caches reused across agents, AgentKVShift \citep{agentkvshift} corrects the retrieved memory chunks of an agent by an offset estimated from a few recomputed probe tokens, and CacheReforge \citep{cachereforge} bounds the recomputation that recovers an updated adapter's outputs from stale caches. StaleSpec \citep{stalespec} guarantees exact outputs after context edits through self-speculative verification. KVShareArena \citep{kvsharearena} benchmarks repairs across contexts and checkpoints, KV-Repurpose \citep{kvrepurpose} reports that segment reuse struggles on cross-document reasoning, and decision margins have been used to model output preservation under KV compression \citep{ocp}. RelaxKV \citep{relaxkv} chooses both the tokens to recompute and the context against which they are recomputed, and it restricts that context to a query-relevant part. Transition Overlap \citep{transitionoverlap} reports that recomputing non-contiguous tokens can add discontinuities between neighbouring states, and it instead encodes each segment after a short temporary overlap whose KV is not stored. SemPIC \citep{sempic} trains a LoRA writer that compiles document caches for the unchanged model. LatentRemedy \citep{latentremedy} attributes part of the loss of reuse across agents to the agent's difficulty in reading the reused context under its own instructions, and it places a few learned latent tokens after the reused block. SurroKV, StaleSpec and the decision-margin model are concurrent anonymous submissions under review at ICLR 2027.

\begin{table}[t]
\centering
\footnotesize
\setlength{\tabcolsep}{4pt}
\begin{tabular}{lccc}
\toprule
Method & \begin{tabular}[b]{@{}c@{}}Sparse\\tokens\end{tabular} & \begin{tabular}[b]{@{}c@{}}Full prefill as\\reference or\\explanation\end{tabular} & \begin{tabular}[b]{@{}c@{}}Attention\\locates\end{tabular} \\
\midrule
CacheBlend & \ding{51} & \ding{51} & \ding{55} \\
EPIC & \ding{51} & \ding{55} & \ding{55} \\
ProphetKV & \ding{51} & partly$^{h}$ & \ding{51} \\
InfoFlow KV & \ding{51} & partly$^{a}$ & \ding{51} \\
CacheTune & \ding{51} & \ding{51}$^{b}$ & \ding{55} \\
SurroKV & \ding{51}$^{c}$ & \ding{51}$^{d}$ & \ding{51}$^{c}$ \\
KVCOMM & --$^{e}$ & \ding{51} & \ding{55} \\
BCR & partly$^{f}$ & \ding{51}$^{g}$ & \ding{51} \\
SAFE-KV & \ding{55} & \ding{51} & \ding{55} \\
\bottomrule
\end{tabular}
\caption{Three choices of repair methods, as stated in their papers: recomputing a sparse set of tokens (high-deviation, boundary, high-attention or low-frequency tokens); taking full prefill as the reference or the explanation of what a repair restores; locating what to recompute by the attention of the question or of draft tokens over the reused cache. Methods: CacheBlend \citep{cacheblend}, EPIC \citep{epic}, ProphetKV \citep{prophetkv}, InfoFlow KV \citep{infoflowkv}, CacheTune \citep{cachetune}, SurroKV \citep{surrokv}, KVCOMM \citep{kvcomm}, BCR \citep{bcr}, SAFE-KV \citep{safekv}. $^{a}$Explains repair as restoring information flow and cross-chunk interaction. $^{b}$The reference is the attention distribution of full recomputation. $^{c}$In its online step, which uses an existing selector. $^{d}$The reference is full prefill; the surrogate prefix brings attention routing, not the values, closer to it. $^{e}$Predicts offsets instead of recomputing. $^{f}$Its selection outperforms random selection by far more for single rows than for 64-token slices: on one 2048-token segment, it removes 49.5\% of the attention-output error beyond chance for single rows and 10.6\% for 64-token slices. $^{g}$Repair is measured by the distance of attention outputs from those of the dense cache. $^{h}$The reference is the full-prefill attention of the question tokens, not the whole attention map.}
\label{tab:assume}
\end{table}

\needspace{6\baselineskip}
\section{Additional Results}
\label{app:results}

\begin{table*}[t]
\centering
\small
\begin{tabular}{llccccc}
\toprule
Dataset & Model & $N$ & Loss & Universal & Same-category & Similarity \\
\midrule
LoCoMo & Qwen3-8B & 105 & 0.47 & 0.11 & \textbf{+0.19 [+0.10, +0.26]} & \textbf{+0.16 [+0.08, +0.24]} \\
 & Llama-3.1-8B & 102 & 0.19 & 0.14 & \textbf{+0.11 [+0.04, +0.18]} & \textminus{}0.06 [\textminus{}0.13, +0.02] \\
 & Mistral-7B & 36 & 0.28 & 0.31 & \textminus{}0.11 [\textminus{}0.30, +0.08] & +0.08 [\textminus{}0.03, +0.20] \\
\midrule
NQ-Open & Qwen3-8B & 146 & 0.85 & 0.43 & \textbf{+0.23 [+0.14, +0.32]} & \textbf{+0.19 [+0.12, +0.27]} \\
 & Llama-3.1-8B & 128 & 0.84 & 0.72 & \textminus{}0.04 [\textminus{}0.13, +0.06] & \textbf{+0.16 [+0.07, +0.25]} \\
 & Mistral-7B & 101 & 0.82 & 0.65 & +0.07 [\textminus{}0.02, +0.16] & +0.10 [\textminus{}0.01, +0.21] \\
\midrule
Gorilla & Qwen3-8B & 89 & 1.00 & 0.18 & \textbf{+0.82 [+0.74, +0.89]} & +0.00 [+0.00, +0.00]$^\dagger$ \\
 & Llama-3.1-8B & 100 & 0.96 & 0.38 & \textbf{+0.62 [+0.53, +0.71]} & \textminus{}0.04 [\textminus{}0.08, \textminus{}0.01]$^\dagger$ \\
 & Mistral-7B & 108 & 0.94 & 0.92 & \textbf{+0.08 [+0.04, +0.14]} & \textminus{}0.06 [\textminus{}0.10, \textminus{}0.02]$^\dagger$ \\
\midrule
TriviaQA & Qwen3-8B & 151 & 0.73 & 0.30 & \textbf{+0.25 [+0.17, +0.33]} & \textbf{+0.17 [+0.08, +0.26]} \\
 & Llama-3.1-8B & 127 & 0.80 & 0.25 & \textbf{+0.23 [+0.14, +0.32]} & \textbf{+0.31 [+0.22, +0.41]} \\
 & Mistral-7B & 117 & 0.75 & 0.33 & \textbf{+0.13 [+0.03, +0.22]} & \textbf{+0.29 [+0.19, +0.40]} \\
\bottomrule
\end{tabular}
\caption{Reuse loss and its decomposition on $Q^*$. Loss is $1-R(\mathrm{SIM})$, the universal term is $1-R(\mathrm{OTHER})$, the same-category term is $R(\mathrm{OTHER})-R(\mathrm{RAND})$ and the similarity term is $R(\mathrm{RAND})-R(\mathrm{SIM})$, all for position-corrected reuse. Intervals are 95\% paired bootstrap intervals, with questions resampled (conversations for LoCoMo); bold intervals exclude zero. $^\dagger$Retention under RAND is zero, so the term cannot be positive; these intervals are not set in bold.}
\label{tab:main_decomposition}
\end{table*}

\begin{table*}[t]
\centering
\footnotesize
\begin{tabular}{llcccccccc}
\toprule
Dataset & Model & $N$ & None & \begin{tabular}[b]{@{}c@{}}Pad\\in-domain\end{tabular} & \begin{tabular}[b]{@{}c@{}}Pad\\SurroKV\end{tabular} & \begin{tabular}[b]{@{}c@{}}Pad\\neutral\end{tabular} & \begin{tabular}[b]{@{}c@{}}SurroKV\\full\end{tabular} & \begin{tabular}[b]{@{}c@{}}Token\\10\%\end{tabular} & \begin{tabular}[b]{@{}c@{}}Token 10\%\\+ pad\end{tabular} \\
\midrule
LoCoMo & Qwen3-8B & 105 & 0.53 & 0.72 & 0.66 & 0.59 & 0.90 & 0.72 & 0.91 \\
 & Llama-3.1-8B & 102 & 0.81 & 0.85 & 0.80 & 0.85 & 0.95 & 0.86 & 0.97 \\
 & Mistral-7B & 36 & 0.72 & 0.78 & 0.81 & 0.81 & 0.83 & 0.75 & 0.92 \\
\midrule
NQ-Open & Qwen3-8B & 146 & 0.15 & 0.22 & 0.17 & 0.18 & 0.41 & 0.24 & 0.38 \\
 & Llama-3.1-8B & 128 & 0.16 & 0.17 & 0.13 & 0.15 & 0.63 & 0.38 & 0.55 \\
 & Mistral-7B & 101 & 0.18 & 0.22 & 0.23 & 0.24 & 0.48 & 0.34 & 0.42 \\
\midrule
Gorilla & Qwen3-8B & 89 & 0.00 & 0.13 & 0.00 & 0.00 & 0.34 & 0.12 & 0.39 \\
 & Llama-3.1-8B & 100 & 0.04 & 0.06 & 0.08 & 0.04 & 0.33 & 0.16 & 0.25 \\
 & Mistral-7B & 108 & 0.06 & 0.24 & 0.03 & 0.03 & 0.23 & 0.24 & 0.33 \\
\midrule
TriviaQA & Qwen3-8B & 151 & 0.27 & 0.34 & 0.36 & 0.36 & 0.68 & 0.49 & 0.64 \\
 & Llama-3.1-8B & 127 & 0.20 & 0.42 & 0.42 & 0.38 & 0.78 & 0.55 & 0.76 \\
 & Mistral-7B & 117 & 0.25 & 0.40 & 0.37 & 0.36 & 0.73 & 0.52 & 0.68 \\
\bottomrule
\end{tabular}
\caption{Retention on $Q^*$ under existing repairs, under SIM. None: no repair. Pad: each chunk encoded after 512 tokens of in-domain text, of the SurroKV Wikipedia surrogate, or of neutral prose. SurroKV full: a 1,024-token surrogate followed by recomputing 10\% of the tokens ranked by question attention. Token 10\%: the online method chosen for the setting on development questions, with EPIC applied to each chunk and CacheBlend in a one-step version (Appendix~\ref{app:impl}). Token 10\% + pad: this method combined with the in-domain pad.}
\label{tab:repair}
\end{table*}

\begin{table*}[t]
\centering
\footnotesize
\begin{tabular}{llcccccccc}
\toprule
Dataset & Model & $N$ & Chance & \begin{tabular}[b]{@{}c@{}}$d_1$\\reuse\end{tabular} & \begin{tabular}[b]{@{}c@{}}$d_2$\\reuse\end{tabular} & \begin{tabular}[b]{@{}c@{}}$d_1$\\full\end{tabular} & \begin{tabular}[b]{@{}c@{}}$d_2$\\full\end{tabular} & $d_3$ & \begin{tabular}[b]{@{}c@{}}$d_1$ reuse:\\last chunk\end{tabular} \\
\midrule
LoCoMo & Qwen3-8B & 29 & 6.4 & 10.3 & 17.2 & 20.7 & 31.0 & 51.7 & 86.2 \\
 & Llama-3.1-8B & 28 & 6.4 & 0.0 & 14.3 & 46.4 & 89.3 & 50.0 & 100.0 \\
 & Mistral-7B & 14 & 6.3 & 7.1 & 7.1 & 14.3 & 42.9 & 50.0 & 92.9 \\
\midrule
NQ-Open & Qwen3-8B & 36 & 3.4 & 11.1 & 16.7 & 11.1 & 27.8 & 50.0 & 77.8 \\
 & Llama-3.1-8B & 38 & 3.3 & 10.5 & 13.2 & 23.7 & 55.3 & 55.3 & 100.0 \\
 & Mistral-7B & 33 & 3.4 & 6.1 & 6.1 & 9.1 & 54.5 & 63.6 & 100.0 \\
\midrule
Gorilla & Qwen3-8B & 16 & 3.2 & 12.5 & 18.8 & 18.8 & 12.5 & 56.2 & 68.8 \\
 & Llama-3.1-8B & 11 & 3.2 & 0.0 & 9.1 & 27.3 & 36.4 & 36.4 & 100.0 \\
 & Mistral-7B & 13 & 3.2 & 0.0 & 0.0 & 0.0 & 7.7 & 38.5 & 100.0 \\
\midrule
TriviaQA & Qwen3-8B & 50 & 3.3 & 6.0 & 14.0 & 16.0 & 6.0 & 54.0 & 58.0 \\
 & Llama-3.1-8B & 52 & 3.3 & 1.9 & 7.7 & 50.0 & 63.5 & 51.9 & 100.0 \\
 & Mistral-7B & 49 & 3.3 & 2.0 & 10.2 & 26.5 & 55.1 & 55.1 & 100.0 \\
\bottomrule
\end{tabular}
\caption{Top-1 accuracy (\%) of target-chunk detectors on development questions answered correctly by full prefill. Chance: the accuracy of choosing a chunk at random. $d_1$: the sum of question attention over a chunk; $d_2$: its mean over the upper half of the layers, with the first 16 tokens of each chunk excluded. Both are computed on the reused (position-corrected) KV and on the full-prefill KV. $d_3$: the cosine similarity between the BGE-M3 embeddings of the question and the chunk. Last column: the share of questions on which $d_1$ over the reused cache selects the chunk next to the question.}
\label{tab:detect}
\end{table*}

\begin{table*}[t]
\centering
\footnotesize
\setlength{\tabcolsep}{4pt}
\begin{tabular}{llcccccccc}
\toprule
Dataset & Model & $N$ & $R$ & \begin{tabular}[b]{@{}c@{}}Target\\KV\end{tabular} & \begin{tabular}[b]{@{}c@{}}All\\distr.\end{tabular} & \begin{tabular}[b]{@{}c@{}}1\\similar\end{tabular} & \begin{tabular}[b]{@{}c@{}}1\\random\end{tabular} & \begin{tabular}[b]{@{}c@{}}All distr.\\$-$ Target\end{tabular} & \begin{tabular}[b]{@{}c@{}}1 similar\\$-$ Target\end{tabular} \\
\midrule
LoCoMo & Qwen3-8B & 105 & 0.53 & 0.61 & 0.79 & 0.54 & 0.58 & \begin{tabular}[t]{@{}c@{}}\textbf{+0.18}\\[-1pt]{\scriptsize [+0.07, +0.29]}\end{tabular} & \begin{tabular}[t]{@{}c@{}}\textbf{\textminus{}0.07}\\[-1pt]{\scriptsize [\textminus{}0.11, \textminus{}0.03]}\end{tabular} \\
 & Llama-3.1-8B & 102 & 0.81 & 0.71 & 0.82 & 0.79 & 0.82 & \begin{tabular}[t]{@{}c@{}}\textbf{+0.12}\\[-1pt]{\scriptsize [+0.03, +0.20]}\end{tabular} & \begin{tabular}[t]{@{}c@{}}+0.09\\[-1pt]{\scriptsize [+0.00, +0.17]}\end{tabular} \\
 & Mistral-7B & 36 & 0.72 & 0.61 & 0.64 & 0.78 & 0.72 & \begin{tabular}[t]{@{}c@{}}+0.03\\[-1pt]{\scriptsize [\textminus{}0.10, +0.18]}\end{tabular} & \begin{tabular}[t]{@{}c@{}}\textbf{+0.17}\\[-1pt]{\scriptsize [+0.05, +0.31]}\end{tabular} \\
\midrule
NQ-Open & Qwen3-8B & 146 & 0.15 & 0.18 & 0.71 & 0.15 & 0.14 & \begin{tabular}[t]{@{}c@{}}\textbf{+0.53}\\[-1pt]{\scriptsize [+0.44, +0.62]}\end{tabular} & \begin{tabular}[t]{@{}c@{}}\textminus{}0.03\\[-1pt]{\scriptsize [\textminus{}0.10, +0.03]}\end{tabular} \\
 & Llama-3.1-8B & 128 & 0.16 & 0.23 & 0.70 & 0.18 & 0.16 & \begin{tabular}[t]{@{}c@{}}\textbf{+0.46}\\[-1pt]{\scriptsize [+0.36, +0.56]}\end{tabular} & \begin{tabular}[t]{@{}c@{}}\textminus{}0.05\\[-1pt]{\scriptsize [\textminus{}0.13, +0.02]}\end{tabular} \\
 & Mistral-7B & 101 & 0.18 & 0.46 & 0.66 & 0.17 & 0.17 & \begin{tabular}[t]{@{}c@{}}\textbf{+0.21}\\[-1pt]{\scriptsize [+0.08, +0.35]}\end{tabular} & \begin{tabular}[t]{@{}c@{}}\textbf{\textminus{}0.29}\\[-1pt]{\scriptsize [\textminus{}0.39, \textminus{}0.19]}\end{tabular} \\
\midrule
Gorilla & Qwen3-8B & 89 & 0.00 & 0.02 & 0.43 & 0.01 & 0.02 & \begin{tabular}[t]{@{}c@{}}\textbf{+0.40}\\[-1pt]{\scriptsize [+0.29, +0.52]}\end{tabular} & \begin{tabular}[t]{@{}c@{}}\textminus{}0.01\\[-1pt]{\scriptsize [\textminus{}0.03, +0.00]}\end{tabular} \\
 & Llama-3.1-8B & 100 & 0.04 & 0.17 & 0.43 & 0.03 & 0.03 & \begin{tabular}[t]{@{}c@{}}\textbf{+0.26}\\[-1pt]{\scriptsize [+0.15, +0.38]}\end{tabular} & \begin{tabular}[t]{@{}c@{}}\textbf{\textminus{}0.14}\\[-1pt]{\scriptsize [\textminus{}0.21, \textminus{}0.07]}\end{tabular} \\
 & Mistral-7B & 108 & 0.06 & 0.32 & 0.53 & 0.06 & 0.06 & \begin{tabular}[t]{@{}c@{}}\textbf{+0.20}\\[-1pt]{\scriptsize [+0.08, +0.32]}\end{tabular} & \begin{tabular}[t]{@{}c@{}}\textbf{\textminus{}0.27}\\[-1pt]{\scriptsize [\textminus{}0.35, \textminus{}0.19]}\end{tabular} \\
\midrule
TriviaQA & Qwen3-8B & 151 & 0.27 & 0.31 & 0.80 & 0.25 & 0.30 & \begin{tabular}[t]{@{}c@{}}\textbf{+0.49}\\[-1pt]{\scriptsize [+0.40, +0.58]}\end{tabular} & \begin{tabular}[t]{@{}c@{}}\textminus{}0.06\\[-1pt]{\scriptsize [\textminus{}0.13, +0.01]}\end{tabular} \\
 & Llama-3.1-8B & 127 & 0.20 & 0.33 & 0.74 & 0.23 & 0.24 & \begin{tabular}[t]{@{}c@{}}\textbf{+0.41}\\[-1pt]{\scriptsize [+0.31, +0.51]}\end{tabular} & \begin{tabular}[t]{@{}c@{}}\textbf{\textminus{}0.10}\\[-1pt]{\scriptsize [\textminus{}0.17, \textminus{}0.03]}\end{tabular} \\
 & Mistral-7B & 117 & 0.25 & 0.55 & 0.71 & 0.24 & 0.26 & \begin{tabular}[t]{@{}c@{}}\textbf{+0.16}\\[-1pt]{\scriptsize [+0.03, +0.29]}\end{tabular} & \begin{tabular}[t]{@{}c@{}}\textbf{\textminus{}0.31}\\[-1pt]{\scriptsize [\textminus{}0.39, \textminus{}0.21]}\end{tabular} \\
\bottomrule
\end{tabular}
\caption{KV replacement under SIM. Columns give retention with no replacement ($R$), after replacing the target chunk (Target KV), all distractor chunks (All distr.), the single distractor most similar to the target (1 similar) and one random distractor (1 random) with their full-prefill KV. The last two columns are paired differences. The random distractor coincides with the most similar one on 0--6\% of questions.}
\label{tab:replacement}
\end{table*}

\begin{table*}[t]
\centering
\footnotesize
\begin{tabular}{llccccc}
\toprule
Dataset & Model & $N$ & \begin{tabular}[b]{@{}c@{}}Target KV\\$-$ $R$\end{tabular} & \begin{tabular}[b]{@{}c@{}}All distr.\\$-$ $R$\end{tabular} & \begin{tabular}[b]{@{}c@{}}1 similar\\$-$ $R$\end{tabular} & \begin{tabular}[b]{@{}c@{}}1 random\\$-$ $R$\end{tabular} \\
\midrule
LoCoMo & Qwen3-8B & 105 & \begin{tabular}[t]{@{}c@{}}\textbf{+0.08}\\[-1pt]{\scriptsize [+0.03, +0.12]}\end{tabular} & \begin{tabular}[t]{@{}c@{}}\textbf{+0.26}\\[-1pt]{\scriptsize [+0.17, +0.35]}\end{tabular} & \begin{tabular}[t]{@{}c@{}}+0.01\\[-1pt]{\scriptsize [+0.00, +0.03]}\end{tabular} & \begin{tabular}[t]{@{}c@{}}\textbf{+0.05}\\[-1pt]{\scriptsize [+0.01, +0.08]}\end{tabular} \\
 & Llama-3.1-8B & 102 & \begin{tabular}[t]{@{}c@{}}\textbf{\textminus{}0.11}\\[-1pt]{\scriptsize [\textminus{}0.16, \textminus{}0.04]}\end{tabular} & \begin{tabular}[t]{@{}c@{}}+0.01\\[-1pt]{\scriptsize [\textminus{}0.07, +0.07]}\end{tabular} & \begin{tabular}[t]{@{}c@{}}\textminus{}0.02\\[-1pt]{\scriptsize [\textminus{}0.05, +0.02]}\end{tabular} & \begin{tabular}[t]{@{}c@{}}+0.01\\[-1pt]{\scriptsize [\textminus{}0.02, +0.04]}\end{tabular} \\
 & Mistral-7B & 36 & \begin{tabular}[t]{@{}c@{}}\textminus{}0.11\\[-1pt]{\scriptsize [\textminus{}0.26, +0.00]}\end{tabular} & \begin{tabular}[t]{@{}c@{}}\textbf{\textminus{}0.08}\\[-1pt]{\scriptsize [\textminus{}0.19, \textminus{}0.02]}\end{tabular} & \begin{tabular}[t]{@{}c@{}}+0.06\\[-1pt]{\scriptsize [+0.00, +0.12]}\end{tabular} & \begin{tabular}[t]{@{}c@{}}+0.00\\[-1pt]{\scriptsize [\textminus{}0.09, +0.07]}\end{tabular} \\
\midrule
NQ-Open & Qwen3-8B & 146 & \begin{tabular}[t]{@{}c@{}}+0.03\\[-1pt]{\scriptsize [\textminus{}0.03, +0.10]}\end{tabular} & \begin{tabular}[t]{@{}c@{}}\textbf{+0.56}\\[-1pt]{\scriptsize [+0.48, +0.64]}\end{tabular} & \begin{tabular}[t]{@{}c@{}}+0.00\\[-1pt]{\scriptsize [\textminus{}0.02, +0.02]}\end{tabular} & \begin{tabular}[t]{@{}c@{}}\textminus{}0.01\\[-1pt]{\scriptsize [\textminus{}0.03, +0.01]}\end{tabular} \\
 & Llama-3.1-8B & 128 & \begin{tabular}[t]{@{}c@{}}+0.07\\[-1pt]{\scriptsize [+0.00, +0.15]}\end{tabular} & \begin{tabular}[t]{@{}c@{}}\textbf{+0.53}\\[-1pt]{\scriptsize [+0.44, +0.62]}\end{tabular} & \begin{tabular}[t]{@{}c@{}}+0.02\\[-1pt]{\scriptsize [\textminus{}0.01, +0.05]}\end{tabular} & \begin{tabular}[t]{@{}c@{}}\textminus{}0.01\\[-1pt]{\scriptsize [\textminus{}0.03, +0.02]}\end{tabular} \\
 & Mistral-7B & 101 & \begin{tabular}[t]{@{}c@{}}\textbf{+0.28}\\[-1pt]{\scriptsize [+0.18, +0.38]}\end{tabular} & \begin{tabular}[t]{@{}c@{}}\textbf{+0.49}\\[-1pt]{\scriptsize [+0.38, +0.59]}\end{tabular} & \begin{tabular}[t]{@{}c@{}}\textminus{}0.01\\[-1pt]{\scriptsize [\textminus{}0.06, +0.04]}\end{tabular} & \begin{tabular}[t]{@{}c@{}}\textminus{}0.01\\[-1pt]{\scriptsize [\textminus{}0.05, +0.02]}\end{tabular} \\
\midrule
Gorilla & Qwen3-8B & 89 & \begin{tabular}[t]{@{}c@{}}+0.02\\[-1pt]{\scriptsize [+0.00, +0.06]}\end{tabular} & \begin{tabular}[t]{@{}c@{}}\textbf{+0.43}\\[-1pt]{\scriptsize [+0.33, +0.53]}\end{tabular} & \begin{tabular}[t]{@{}c@{}}+0.01\\[-1pt]{\scriptsize [+0.00, +0.03]}\end{tabular} & \begin{tabular}[t]{@{}c@{}}+0.02\\[-1pt]{\scriptsize [+0.00, +0.06]}\end{tabular} \\
 & Llama-3.1-8B & 100 & \begin{tabular}[t]{@{}c@{}}\textbf{+0.13}\\[-1pt]{\scriptsize [+0.06, +0.20]}\end{tabular} & \begin{tabular}[t]{@{}c@{}}\textbf{+0.39}\\[-1pt]{\scriptsize [+0.29, +0.49]}\end{tabular} & \begin{tabular}[t]{@{}c@{}}\textminus{}0.01\\[-1pt]{\scriptsize [\textminus{}0.03, +0.00]}\end{tabular} & \begin{tabular}[t]{@{}c@{}}\textminus{}0.01\\[-1pt]{\scriptsize [\textminus{}0.04, +0.02]}\end{tabular} \\
 & Mistral-7B & 108 & \begin{tabular}[t]{@{}c@{}}\textbf{+0.27}\\[-1pt]{\scriptsize [+0.19, +0.35]}\end{tabular} & \begin{tabular}[t]{@{}c@{}}\textbf{+0.47}\\[-1pt]{\scriptsize [+0.37, +0.57]}\end{tabular} & \begin{tabular}[t]{@{}c@{}}+0.00\\[-1pt]{\scriptsize [+0.00, +0.00]}\end{tabular} & \begin{tabular}[t]{@{}c@{}}+0.01\\[-1pt]{\scriptsize [+0.00, +0.03]}\end{tabular} \\
\midrule
TriviaQA & Qwen3-8B & 151 & \begin{tabular}[t]{@{}c@{}}+0.04\\[-1pt]{\scriptsize [\textminus{}0.04, +0.12]}\end{tabular} & \begin{tabular}[t]{@{}c@{}}\textbf{+0.53}\\[-1pt]{\scriptsize [+0.44, +0.62]}\end{tabular} & \begin{tabular}[t]{@{}c@{}}\textminus{}0.02\\[-1pt]{\scriptsize [\textminus{}0.05, +0.01]}\end{tabular} & \begin{tabular}[t]{@{}c@{}}+0.03\\[-1pt]{\scriptsize [\textminus{}0.01, +0.06]}\end{tabular} \\
 & Llama-3.1-8B & 127 & \begin{tabular}[t]{@{}c@{}}\textbf{+0.13}\\[-1pt]{\scriptsize [+0.06, +0.19]}\end{tabular} & \begin{tabular}[t]{@{}c@{}}\textbf{+0.54}\\[-1pt]{\scriptsize [+0.43, +0.63]}\end{tabular} & \begin{tabular}[t]{@{}c@{}}+0.02\\[-1pt]{\scriptsize [\textminus{}0.02, +0.06]}\end{tabular} & \begin{tabular}[t]{@{}c@{}}\textbf{+0.04}\\[-1pt]{\scriptsize [+0.01, +0.08]}\end{tabular} \\
 & Mistral-7B & 117 & \begin{tabular}[t]{@{}c@{}}\textbf{+0.30}\\[-1pt]{\scriptsize [+0.21, +0.39]}\end{tabular} & \begin{tabular}[t]{@{}c@{}}\textbf{+0.46}\\[-1pt]{\scriptsize [+0.36, +0.56]}\end{tabular} & \begin{tabular}[t]{@{}c@{}}\textminus{}0.01\\[-1pt]{\scriptsize [\textminus{}0.04, +0.02]}\end{tabular} & \begin{tabular}[t]{@{}c@{}}+0.01\\[-1pt]{\scriptsize [\textminus{}0.03, +0.05]}\end{tabular} \\
\bottomrule
\end{tabular}
\caption{KV replacement under SIM: change in retention against no replacement ($R$), with 95\% paired bootstrap intervals.}
\label{tab:replacement_ci}
\end{table*}

\paragraph{Similarity term on Gorilla.} On Gorilla, retention under SIM is slightly higher than under RAND (by at most 0.06), which makes the similarity term significantly negative for two models. Because retention under RAND is zero on Gorilla, the three distractor sources cannot show an additional loss from similar API documents on this dataset.

\paragraph{Existing repairs.} Table~\ref{tab:repair} reports existing repairs applied to the SIM condition. Encoding every chunk after 512 tokens of in-domain text matches or beats encoding it after the SurroKV Wikipedia surrogate of the same length in eight of twelve settings. The strongest repairs combine a surrogate prefix with 10\% online recomputation (SurroKV's full method, or the best token-level selector on top of the in-domain prefix). In each setting, the stronger of the two raises retention on LoCoMo close to full prefill (0.91--0.97) and on TriviaQA to 0.68--0.78, but leaves retention at 0.41--0.63 on NQ-Open and at 0.33--0.39 on Gorilla.

\paragraph{Allocation and detection details.} On LoCoMo, Target+4 does not change retention significantly, and recomputing the retrieved session lowers retention for Qwen3-8B (\textminus{}0.10 at 5\%). For the detectors on Gorilla, the question is the user request alone (Appendix~\ref{app:data}). The Gorilla development settings contain only 11--16 questions that full prefill answers, so the estimates of detector accuracy on Gorilla are coarse. On questions where the retriever does not rank the target first, recomputing the retrieved chunk together with its four most similar chunks (Retrieved+4) gives higher retention than recomputing the retrieved chunk alone, significantly in seven of nine settings at 10\%, but it never significantly beats the token-level selector (Table~\ref{tab:miss}). At 5\%, a gate that applies Retrieved+4 only when the retriever is confident amounts to choosing one method per domain (Table~\ref{tab:hybrid}).

\paragraph{Universal term and KV replacement.} Replacing every distractor raises retention for Qwen3-8B on NQ-Open from 0.15 to 0.71. The universal term is largest for Gorilla with Mistral-7B (0.92) and on NQ-Open (0.43--0.72); in the Gorilla setting, 93\% of the wrong answers under OTHER describe the off-domain prose instead of naming an API. Replacing a single distractor, the most similar one or a random one, changes retention significantly in two of 24 cases (by at most 0.06; Tables~\ref{tab:replacement} and~\ref{tab:replacement_ci}), and most of the recoverable loss is restored only by replacing many distractors at once. On LoCoMo (dialogue memory), replacing all distractors raises retention significantly only for Qwen3-8B, and it lowers retention for Mistral-7B (\textminus{}0.08, [\textminus{}0.19, \textminus{}0.02]).

\begin{table*}[t]
\centering
\footnotesize
\setlength{\tabcolsep}{4pt}
\begin{tabular}{p{1.9cm}p{3.5cm}p{1.8cm}p{1.9cm}p{5.0cm}}
\toprule
Model, data & Question & Gold answer & Reuse answer & Similar passage (title: sentence) \\
\midrule
Qwen3-8B, NQ-Open & Who was the actor that played ben stone on law and order & Michael Moriarty & Philip Winchester & \emph{Peter Stone (Chicago and Law \& Order character)}: \ldots saw it develop and we saw the actor, it made sense. Peter Stone (Chicago and Law \& Order character) Peter Stone is a fictional character portrayed by Philip Winchester. \\
Llama-3.1-8B, NQ-Open & Who won the most medals in the 1924 winter olympics & Norway & Ville Ritola & \emph{Ville Ritola}: \ldots win the 3000 m team race. \\
Mistral-7B, NQ-Open & Dendrites and cell bodies are components of what type of matter found in the brain & gray & Neuronal matter & \emph{Neuron}: Other structural proteins that are important for neuronal function are actin and the tubulin of microtubules. \\
Qwen3-8B, TriviaQA & Liberty Bell was the theme tune for which British TV show? & Monty Python's Flying Circus & EastEnders & \emph{EastEnders theme tune}: \ldots the theme is still popular and in the national psyche. The theme tune was used in the 2012 Summer Olympics opening ceremony. \\
Llama-3.1-8B, TriviaQA & What Tory MP was found to have claimed unsuccessfully on his parliamentary expenses for a floating duck island? & Sir Peter Viggers & Gordon Brown & \emph{United Kingdom parliamentary expenses scandal}: Among those who had to repay expenses were Prime Minister Gordon Brown who claimed £12,415 for cleaning and gardening costs and Liberal Democrat leader Nick Clegg who was\ldots \\
Mistral-7B, TriviaQA & What was the character name of Commander Shore's daughter in the 60's series Stingray (voiced by Lois Maxwell) & Atlanta & Lois Lane & \emph{Lois Lane}: \ldots before was one of Luthor's robotic duplicates. \\
\bottomrule
\end{tabular}
\caption{For each model and open-domain QA dataset, the first question in \Qs{} (taking questions in identifier order, and the test set before the second set) that meets two conditions: its reuse answer under SIM is a short wrong answer (at most six words, with no reference to a document) that names an entity occurring in a similar passage but not in the target passage, and its reuse answer under RAND is correct. Long sentences are cut around that entity. Questions keep the capitalisation and punctuation of the datasets.}
\label{tab:examples}
\end{table*}

\begin{table*}[t]
\centering
\footnotesize
\setlength{\tabcolsep}{3pt}
\begin{tabular}{llcccccccc}
\toprule
Dataset & Model & $N$ & SIM & RAND & \begin{tabular}[b]{@{}c@{}}SIM $-$\\RAND\end{tabular} & \begin{tabular}[b]{@{}c@{}}Abstain\\SIM\end{tabular} & \begin{tabular}[b]{@{}c@{}}Abstain\\RAND\end{tabular} & \begin{tabular}[b]{@{}c@{}}Conditional\\(entity)\end{tabular} & \begin{tabular}[b]{@{}c@{}}Conditional\\(phrase)\end{tabular} \\
\midrule
NQ-Open & Qwen3-8B & 146 & 61.6 & 17.1 & \begin{tabular}[t]{@{}c@{}}\textbf{+44.5}\\[-1pt]{\scriptsize [+34.9, +53.4]}\end{tabular} & 4.0 & 25.0 & \begin{tabular}[t]{@{}c@{}}\textbf{+40.9}\\[-1pt]{\scriptsize [+27.5, +53.3]}\end{tabular} & \begin{tabular}[t]{@{}c@{}}\textbf{+47.6}\\[-1pt]{\scriptsize [+35.1, +60.1]}\end{tabular} \\
 & Llama-3.1-8B & 128 & 63.3 & 39.8 & \begin{tabular}[t]{@{}c@{}}\textbf{+23.4}\\[-1pt]{\scriptsize [+13.3, +33.6]}\end{tabular} & 0.0 & 14.9 & \begin{tabular}[t]{@{}c@{}}+6.8\\[-1pt]{\scriptsize [\textminus{}5.8, +19.1]}\end{tabular} & \begin{tabular}[t]{@{}c@{}}+8.3\\[-1pt]{\scriptsize [\textminus{}3.5, +19.8]}\end{tabular} \\
 & Mistral-7B & 101 & 58.4 & 12.9 & \begin{tabular}[t]{@{}c@{}}\textbf{+45.5}\\[-1pt]{\scriptsize [+33.7, +57.4]}\end{tabular} & 10.8 & 72.6 & \begin{tabular}[t]{@{}c@{}}+14.7\\[-1pt]{\scriptsize [\textminus{}7.8, +40.4]}\end{tabular} & \begin{tabular}[t]{@{}c@{}}\textbf{+32.8}\\[-1pt]{\scriptsize [+8.9, +56.8]}\end{tabular} \\
\midrule
TriviaQA & Qwen3-8B & 151 & 40.4 & 17.2 & \begin{tabular}[t]{@{}c@{}}\textbf{+23.2}\\[-1pt]{\scriptsize [+13.2, +32.5]}\end{tabular} & 8.2 & 17.9 & \begin{tabular}[t]{@{}c@{}}\textbf{+22.7}\\[-1pt]{\scriptsize [+7.9, +37.5]}\end{tabular} & \begin{tabular}[t]{@{}c@{}}\textbf{+31.2}\\[-1pt]{\scriptsize [+16.6, +44.7]}\end{tabular} \\
 & Llama-3.1-8B & 127 & 43.3 & 19.7 & \begin{tabular}[t]{@{}c@{}}\textbf{+23.6}\\[-1pt]{\scriptsize [+12.6, +33.1]}\end{tabular} & 3.0 & 9.8 & \begin{tabular}[t]{@{}c@{}}+10.7\\[-1pt]{\scriptsize [\textminus{}5.3, +25.4]}\end{tabular} & \begin{tabular}[t]{@{}c@{}}\textbf{+21.9}\\[-1pt]{\scriptsize [+7.7, +35.9]}\end{tabular} \\
 & Mistral-7B & 117 & 40.2 & 7.7 & \begin{tabular}[t]{@{}c@{}}\textbf{+32.5}\\[-1pt]{\scriptsize [+23.1, +41.9]}\end{tabular} & 26.1 & 59.3 & \begin{tabular}[t]{@{}c@{}}\textbf{+31.4}\\[-1pt]{\scriptsize [+8.5, +54.0]}\end{tabular} & \begin{tabular}[t]{@{}c@{}}\textbf{+40.7}\\[-1pt]{\scriptsize [+17.6, +62.8]}\end{tabular} \\
\bottomrule
\end{tabular}
\caption{Reuse answers under SIM and RAND that name an entity found in some distractor chunk but not in the target chunk. SIM and RAND: share of all \Qs{} questions (\%); SIM $-$ RAND: paired difference in percentage points with 95\% bootstrap interval. Abstain: share of abstentions among wrong reuse answers. Conditional (entity): the SIM $-$ RAND difference among wrong answers that are not abstentions. Conditional (phrase): the same difference when any 1--5-word phrase, rather than only an entity, counts as a match. Bold differences have intervals that exclude zero.}
\label{tab:competing}
\end{table*}

\paragraph{Competing answers.} RAND passages also contain many names, so the share of reuse answers that name a distractor entity under RAND serves as the baseline, rather than zero. Under RAND the model also abstains more often: abstentions make up 10--73\% of its wrong answers, against 0--26\% under SIM. When the comparison is restricted to wrong answers that are not abstentions, the difference remains positive in all six settings and is significant in three. Table~\ref{tab:examples} lists, for each model and open-domain QA dataset, one case like that of Figure~\ref{fig:schematic}, chosen by a fixed rule. Table~\ref{tab:competing} gives the share of all \Qs{} questions whose reuse answer names an entity that occurs in some distractor chunk but not in the target chunk, the share of abstentions among wrong answers, and the same comparison restricted to wrong answers that are not abstentions. Entities are capitalised spans (allowing short connectors such as ``of'' and ``the'') and numbers of at least three digits, excluding spans that occur in the question; spans are matched as token sequences. A broader rule that counts any 1--5-word phrase gives larger differences in the same direction (last column). Abstentions are counted separately, and Table~\ref{tab:abstain} compares their share with that among the errors of full prefill. On Gorilla, a wrong output counts as naming a distractor API when it contains the \texttt{api\_name} of a document other than the target. Under SIM and RAND this holds for 45--91\% of wrong reuse outputs, and the remaining wrong outputs either cite a document number or name no API; under OTHER, none of the wrong outputs names a distractor API.

\begin{table*}[t]
\centering
\footnotesize
\setlength{\tabcolsep}{3pt}
\begin{tabular}{llccccccccc}
\toprule
 & & & SIM & \multicolumn{2}{c}{Competing removed} & \multicolumn{2}{c}{Others removed} & \multicolumn{3}{c}{Gap closed} \\ \cmidrule(lr){4-4}\cmidrule(lr){5-6}\cmidrule(lr){7-8}\cmidrule(lr){9-11}
Dataset & Model & $N$ & Reuse & Full & Reuse & Full & Reuse & Competing & Others & \begin{tabular}[b]{@{}c@{}}Competing\\$-$ Others\end{tabular} \\
\midrule
NQ-Open & Qwen3-8B & 131 & 0.13 & 0.94 & 0.22 & 0.94 & 0.16 & \begin{tabular}[t]{@{}c@{}}\textbf{+0.15}\\[-1pt]{\scriptsize [+0.08, +0.22]}\end{tabular} & \begin{tabular}[t]{@{}c@{}}\textbf{+0.09}\\[-1pt]{\scriptsize [+0.04, +0.15]}\end{tabular} & \begin{tabular}[t]{@{}c@{}}+0.06\\[-1pt]{\scriptsize [\textminus{}0.01, +0.14]}\end{tabular} \\
 & Llama-3.1-8B & 109 & 0.15 & 0.98 & 0.34 & 0.97 & 0.11 & \begin{tabular}[t]{@{}c@{}}\textbf{+0.21}\\[-1pt]{\scriptsize [+0.12, +0.31]}\end{tabular} & \begin{tabular}[t]{@{}c@{}}\textminus{}0.01\\[-1pt]{\scriptsize [\textminus{}0.06, +0.04]}\end{tabular} & \begin{tabular}[t]{@{}c@{}}\textbf{+0.22}\\[-1pt]{\scriptsize [+0.13, +0.31]}\end{tabular} \\
 & Mistral-7B & 86 & 0.19 & 0.93 & 0.22 & 0.95 & 0.17 & \begin{tabular}[t]{@{}c@{}}\textbf{+0.10}\\[-1pt]{\scriptsize [+0.01, +0.20]}\end{tabular} & \begin{tabular}[t]{@{}c@{}}+0.03\\[-1pt]{\scriptsize [\textminus{}0.05, +0.12]}\end{tabular} & \begin{tabular}[t]{@{}c@{}}+0.07\\[-1pt]{\scriptsize [\textminus{}0.03, +0.17]}\end{tabular} \\
\midrule
TriviaQA & Qwen3-8B & 122 & 0.22 & 0.99 & 0.28 & 0.97 & 0.23 & \begin{tabular}[t]{@{}c@{}}+0.07\\[-1pt]{\scriptsize [+0.00, +0.13]}\end{tabular} & \begin{tabular}[t]{@{}c@{}}+0.04\\[-1pt]{\scriptsize [\textminus{}0.01, +0.10]}\end{tabular} & \begin{tabular}[t]{@{}c@{}}+0.02\\[-1pt]{\scriptsize [\textminus{}0.04, +0.10]}\end{tabular} \\
 & Llama-3.1-8B & 96 & 0.16 & 0.94 & 0.24 & 0.99 & 0.17 & \begin{tabular}[t]{@{}c@{}}\textbf{+0.15}\\[-1pt]{\scriptsize [+0.06, +0.24]}\end{tabular} & \begin{tabular}[t]{@{}c@{}}+0.02\\[-1pt]{\scriptsize [\textminus{}0.02, +0.06]}\end{tabular} & \begin{tabular}[t]{@{}c@{}}\textbf{+0.12}\\[-1pt]{\scriptsize [+0.04, +0.22]}\end{tabular} \\
 & Mistral-7B & 88 & 0.19 & 0.95 & 0.23 & 0.97 & 0.16 & \begin{tabular}[t]{@{}c@{}}+0.08\\[-1pt]{\scriptsize [+0.00, +0.16]}\end{tabular} & \begin{tabular}[t]{@{}c@{}}+0.00\\[-1pt]{\scriptsize [\textminus{}0.06, +0.06]}\end{tabular} & \begin{tabular}[t]{@{}c@{}}\textbf{+0.08}\\[-1pt]{\scriptsize [+0.01, +0.15]}\end{tabular} \\
\bottomrule
\end{tabular}
\caption{Competing-entity control on \Qs{} (questions with an answer type). In the SIM distractors, the named entities of the answer\textquoteright s type (Competing removed) or the same number of entities of other types (Others removed) are replaced by generic descriptors such as ``a person\textquotedblright, with the target chunk, the order and the headers unchanged. Reuse and Full: accuracy of reuse and of full prefill. SIM Reuse: accuracy of reuse on the unedited prompt, where full prefill answers every question. Gap closed: how much the gap between full prefill and reuse decreases relative to the unedited prompt. Differences are paired, with 95\% bootstrap intervals.}
\label{tab:compctl}
\end{table*}

\paragraph{Competing-entity control.} For Table~\ref{tab:compctl}, we edit only the bodies of the SIM distractor chunks, on the \Qs{} questions of NQ-Open and TriviaQA; the target chunk, the chunk order, the headers and the question are the same as in the main experiments. Named entities are found with spaCy (\texttt{en\_core\_web\_lg}) and grouped into persons, places, organisations, groups, works, events, languages, dates and quantities. An entity is a candidate when its text occurs neither in the target chunk nor in the question and neither contains nor is contained in a gold answer. The answer's type is the group of the gold answer's mention in the target chunk or, if there is no such mention, is inferred from the question word; the 138 of 770 \Qs{} questions without a type are left out. One edit replaces every candidate of the answer's type; the other replaces the same number of candidates of other types, drawn from the same chunk where possible. Each replaced entity is rewritten as a generic descriptor of its group (for instance ``a person'' or ``that year''), and the descriptor is chosen per model so that the token count of each chunk stays close to the original. On a sample of 40 questions checked by hand, the answer's type is right for 34, and 178 of 204 replaced entities have the right group on those questions. The run also encodes the unedited prompts of the questions with an answer type among the first ten of each setting, and all 49 of them reproduce the outputs of the main experiments. When the entities of the answer's type in the SIM distractors are replaced, reuse answers 22--34\% of the \Qs{} questions that have such a type instead of 13--22\%, while full prefill still answers 93--99\%. The gap narrows significantly in four of the six settings (three after Holm's correction), but in only one when the same number of entities of other types are replaced. Replacing the competing entities lowers the share of wrong reuse answers (abstentions excluded) that name a distractor entity from 64--87\% to 48--67\%, against 62--83\% when other entities are replaced. The difference between the two edits is positive everywhere but significant in only three (two after Holm's correction), so this control does not clearly separate the effect of the competing entities from that of editing. Leaving out the questions for which too few entities of other types were available makes the difference significant in four, and leaving out those whose token count changes by more than 1\% keeps it significant in three.

\begin{table*}[t]
\centering
\footnotesize
\begin{tabular}{llccccc}
\toprule
Dataset & Model & $N$ reuse & Reuse (\%) & $N$ full & Full prefill (\%) & Difference \\
\midrule
NQ-Open & Qwen3-8B & 124 & 4.0 & 218 & 1.4 & +2.7 [\textminus{}0.9, +6.3] \\
 & Llama-3.1-8B & 107 & 0.0 & 193 & 0.5 & \textminus{}0.5 [\textminus{}1.6, +0.0] \\
 & Mistral-7B & 83 & 10.8 & 235 & 6.8 & +4.0 [\textminus{}3.3, +11.6] \\
\midrule
TriviaQA & Qwen3-8B & 110 & 8.2 & 199 & 3.0 & +5.2 [+0.0, +10.7] \\
 & Llama-3.1-8B & 101 & 3.0 & 171 & 4.1 & \textminus{}1.1 [\textminus{}5.4, +3.4] \\
 & Mistral-7B & 88 & 26.1 & 222 & 8.6 & +17.6 [+8.0, +26.9] \\
\bottomrule
\end{tabular}
\caption{Share of abstentions among wrong outputs, for position-corrected reuse under SIM (on $Q^*$) and for full prefill under SIM (on the questions that full prefill answers wrongly).}
\label{tab:abstain}
\end{table*}

\begin{table*}[t]
\centering
\footnotesize
\begin{tabular}{llcccccc}
\toprule
Dataset & Model & $N$ & \begin{tabular}[b]{@{}c@{}}Token\\10\%\end{tabular} & Retrieved+4 & Retrieved & \begin{tabular}[b]{@{}c@{}}Retrieved+4\\$-$ Retrieved\end{tabular} & \begin{tabular}[b]{@{}c@{}}Retrieved+4\\$-$ Token\end{tabular} \\
\midrule
NQ-Open & Qwen3-8B & 66 & 0.24 & 0.26 & 0.14 & \begin{tabular}[t]{@{}c@{}}\textbf{+0.12}\\[-1pt]{\scriptsize [+0.03, +0.21]}\end{tabular} & \begin{tabular}[t]{@{}c@{}}+0.02\\[-1pt]{\scriptsize [\textminus{}0.09, +0.12]}\end{tabular} \\
 & Llama-3.1-8B & 68 & 0.31 & 0.21 & 0.16 & \begin{tabular}[t]{@{}c@{}}+0.04\\[-1pt]{\scriptsize [\textminus{}0.01, +0.12]}\end{tabular} & \begin{tabular}[t]{@{}c@{}}\textminus{}0.10\\[-1pt]{\scriptsize [\textminus{}0.22, +0.01]}\end{tabular} \\
 & Mistral-7B & 50 & 0.34 & 0.36 & 0.20 & \begin{tabular}[t]{@{}c@{}}\textbf{+0.16}\\[-1pt]{\scriptsize [+0.02, +0.30]}\end{tabular} & \begin{tabular}[t]{@{}c@{}}+0.02\\[-1pt]{\scriptsize [\textminus{}0.14, +0.16]}\end{tabular} \\
\midrule
Gorilla & Qwen3-8B & 42 & 0.14 & 0.12 & 0.00 & \begin{tabular}[t]{@{}c@{}}\textbf{+0.12}\\[-1pt]{\scriptsize [+0.02, +0.24]}\end{tabular} & \begin{tabular}[t]{@{}c@{}}\textminus{}0.02\\[-1pt]{\scriptsize [\textminus{}0.14, +0.10]}\end{tabular} \\
 & Llama-3.1-8B & 47 & 0.26 & 0.11 & 0.06 & \begin{tabular}[t]{@{}c@{}}+0.04\\[-1pt]{\scriptsize [\textminus{}0.02, +0.13]}\end{tabular} & \begin{tabular}[t]{@{}c@{}}\textbf{\textminus{}0.15}\\[-1pt]{\scriptsize [\textminus{}0.30, \textminus{}0.02]}\end{tabular} \\
 & Mistral-7B & 57 & 0.30 & 0.23 & 0.09 & \begin{tabular}[t]{@{}c@{}}\textbf{+0.14}\\[-1pt]{\scriptsize [+0.04, +0.25]}\end{tabular} & \begin{tabular}[t]{@{}c@{}}\textminus{}0.07\\[-1pt]{\scriptsize [\textminus{}0.19, +0.04]}\end{tabular} \\
\midrule
TriviaQA & Qwen3-8B & 54 & 0.41 & 0.35 & 0.24 & \begin{tabular}[t]{@{}c@{}}\textbf{+0.11}\\[-1pt]{\scriptsize [+0.02, +0.22]}\end{tabular} & \begin{tabular}[t]{@{}c@{}}\textminus{}0.06\\[-1pt]{\scriptsize [\textminus{}0.20, +0.09]}\end{tabular} \\
 & Llama-3.1-8B & 51 & 0.45 & 0.27 & 0.14 & \begin{tabular}[t]{@{}c@{}}\textbf{+0.14}\\[-1pt]{\scriptsize [+0.06, +0.24]}\end{tabular} & \begin{tabular}[t]{@{}c@{}}\textminus{}0.18\\[-1pt]{\scriptsize [\textminus{}0.33, +0.00]}\end{tabular} \\
 & Mistral-7B & 44 & 0.57 & 0.43 & 0.23 & \begin{tabular}[t]{@{}c@{}}\textbf{+0.20}\\[-1pt]{\scriptsize [+0.07, +0.34]}\end{tabular} & \begin{tabular}[t]{@{}c@{}}\textminus{}0.14\\[-1pt]{\scriptsize [\textminus{}0.32, +0.05]}\end{tabular} \\
\bottomrule
\end{tabular}
\caption{Questions on which the retriever misses the target chunk (10\% budget): retention of the token-level selector, of Retrieved+4 and of Retrieved (the retrieved chunk alone), with paired differences.}
\label{tab:miss}
\end{table*}

\begin{table*}[t]
\centering
\footnotesize
\begin{tabular}{llccccc}
\toprule
Dataset & Model & $N$ & \begin{tabular}[b]{@{}c@{}}Target\\$-$ Token\end{tabular} & \begin{tabular}[b]{@{}c@{}}\begin{tabular}[b]{@{}c@{}}Target\\+4\end{tabular}\\$-$ Token\end{tabular} & \begin{tabular}[b]{@{}c@{}}Retrieved\\$-$ Token\end{tabular} & \begin{tabular}[b]{@{}c@{}}\begin{tabular}[b]{@{}c@{}}Retrieved\\+4\end{tabular}\\$-$ Token\end{tabular} \\
\midrule
LoCoMo & Qwen3-8B & 105 & \begin{tabular}[t]{@{}c@{}}\textminus{}0.04\\[-1pt]{\scriptsize [\textminus{}0.13, +0.03]}\end{tabular} & \begin{tabular}[t]{@{}c@{}}\textminus{}0.06\\[-1pt]{\scriptsize [\textminus{}0.15, +0.01]}\end{tabular} & \begin{tabular}[t]{@{}c@{}}\textbf{\textminus{}0.10}\\[-1pt]{\scriptsize [\textminus{}0.19, \textminus{}0.03]}\end{tabular} & \begin{tabular}[t]{@{}c@{}}\textbf{\textminus{}0.10}\\[-1pt]{\scriptsize [\textminus{}0.20, \textminus{}0.04]}\end{tabular} \\
 & Llama-3.1-8B & 102 & \begin{tabular}[t]{@{}c@{}}+0.01\\[-1pt]{\scriptsize [\textminus{}0.06, +0.07]}\end{tabular} & \begin{tabular}[t]{@{}c@{}}+0.01\\[-1pt]{\scriptsize [\textminus{}0.06, +0.08]}\end{tabular} & \begin{tabular}[t]{@{}c@{}}\textminus{}0.01\\[-1pt]{\scriptsize [\textminus{}0.09, +0.07]}\end{tabular} & \begin{tabular}[t]{@{}c@{}}\textminus{}0.05\\[-1pt]{\scriptsize [\textminus{}0.13, +0.02]}\end{tabular} \\
 & Mistral-7B & 36 & \begin{tabular}[t]{@{}c@{}}+0.03\\[-1pt]{\scriptsize [\textminus{}0.09, +0.15]}\end{tabular} & \begin{tabular}[t]{@{}c@{}}\textminus{}0.03\\[-1pt]{\scriptsize [\textminus{}0.19, +0.13]}\end{tabular} & \begin{tabular}[t]{@{}c@{}}\textminus{}0.03\\[-1pt]{\scriptsize [\textminus{}0.14, +0.10]}\end{tabular} & \begin{tabular}[t]{@{}c@{}}+0.03\\[-1pt]{\scriptsize [\textminus{}0.12, +0.19]}\end{tabular} \\
\midrule
NQ-Open & Qwen3-8B & 146 & \begin{tabular}[t]{@{}c@{}}\textbf{+0.35}\\[-1pt]{\scriptsize [+0.27, +0.43]}\end{tabular} & \begin{tabular}[t]{@{}c@{}}\textbf{+0.29}\\[-1pt]{\scriptsize [+0.22, +0.38]}\end{tabular} & \begin{tabular}[t]{@{}c@{}}\textbf{+0.22}\\[-1pt]{\scriptsize [+0.14, +0.29]}\end{tabular} & \begin{tabular}[t]{@{}c@{}}\textbf{+0.21}\\[-1pt]{\scriptsize [+0.13, +0.28]}\end{tabular} \\
 & Llama-3.1-8B & 128 & \begin{tabular}[t]{@{}c@{}}\textbf{+0.15}\\[-1pt]{\scriptsize [+0.06, +0.23]}\end{tabular} & \begin{tabular}[t]{@{}c@{}}\textbf{+0.18}\\[-1pt]{\scriptsize [+0.09, +0.27]}\end{tabular} & \begin{tabular}[t]{@{}c@{}}+0.08\\[-1pt]{\scriptsize [+0.00, +0.16]}\end{tabular} & \begin{tabular}[t]{@{}c@{}}\textbf{+0.12}\\[-1pt]{\scriptsize [+0.04, +0.21]}\end{tabular} \\
 & Mistral-7B & 101 & \begin{tabular}[t]{@{}c@{}}\textbf{+0.31}\\[-1pt]{\scriptsize [+0.18, +0.43]}\end{tabular} & \begin{tabular}[t]{@{}c@{}}\textbf{+0.21}\\[-1pt]{\scriptsize [+0.09, +0.34]}\end{tabular} & \begin{tabular}[t]{@{}c@{}}\textbf{+0.14}\\[-1pt]{\scriptsize [+0.02, +0.25]}\end{tabular} & \begin{tabular}[t]{@{}c@{}}\textbf{+0.15}\\[-1pt]{\scriptsize [+0.04, +0.26]}\end{tabular} \\
\midrule
Gorilla & Qwen3-8B & 89 & \begin{tabular}[t]{@{}c@{}}\textbf{+0.17}\\[-1pt]{\scriptsize [+0.08, +0.26]}\end{tabular} & \begin{tabular}[t]{@{}c@{}}\textbf{+0.08}\\[-1pt]{\scriptsize [+0.02, +0.13]}\end{tabular} & \begin{tabular}[t]{@{}c@{}}+0.09\\[-1pt]{\scriptsize [+0.00, +0.18]}\end{tabular} & \begin{tabular}[t]{@{}c@{}}\textbf{+0.08}\\[-1pt]{\scriptsize [+0.02, +0.15]}\end{tabular} \\
 & Llama-3.1-8B & 100 & \begin{tabular}[t]{@{}c@{}}\textbf{+0.10}\\[-1pt]{\scriptsize [+0.01, +0.19]}\end{tabular} & \begin{tabular}[t]{@{}c@{}}+0.06\\[-1pt]{\scriptsize [\textminus{}0.02, +0.14]}\end{tabular} & \begin{tabular}[t]{@{}c@{}}+0.02\\[-1pt]{\scriptsize [\textminus{}0.07, +0.10]}\end{tabular} & \begin{tabular}[t]{@{}c@{}}+0.01\\[-1pt]{\scriptsize [\textminus{}0.07, +0.09]}\end{tabular} \\
 & Mistral-7B & 108 & \begin{tabular}[t]{@{}c@{}}\textbf{+0.49}\\[-1pt]{\scriptsize [+0.38, +0.59]}\end{tabular} & \begin{tabular}[t]{@{}c@{}}\textbf{+0.34}\\[-1pt]{\scriptsize [+0.24, +0.44]}\end{tabular} & \begin{tabular}[t]{@{}c@{}}\textbf{+0.19}\\[-1pt]{\scriptsize [+0.09, +0.30]}\end{tabular} & \begin{tabular}[t]{@{}c@{}}\textbf{+0.21}\\[-1pt]{\scriptsize [+0.11, +0.31]}\end{tabular} \\
\midrule
TriviaQA & Qwen3-8B & 151 & \begin{tabular}[t]{@{}c@{}}\textbf{+0.20}\\[-1pt]{\scriptsize [+0.11, +0.29]}\end{tabular} & \begin{tabular}[t]{@{}c@{}}\textbf{+0.23}\\[-1pt]{\scriptsize [+0.14, +0.32]}\end{tabular} & \begin{tabular}[t]{@{}c@{}}\textbf{+0.14}\\[-1pt]{\scriptsize [+0.06, +0.22]}\end{tabular} & \begin{tabular}[t]{@{}c@{}}\textbf{+0.17}\\[-1pt]{\scriptsize [+0.09, +0.25]}\end{tabular} \\
 & Llama-3.1-8B & 127 & \begin{tabular}[t]{@{}c@{}}\textbf{+0.11}\\[-1pt]{\scriptsize [+0.02, +0.20]}\end{tabular} & \begin{tabular}[t]{@{}c@{}}\textbf{+0.09}\\[-1pt]{\scriptsize [+0.01, +0.19]}\end{tabular} & \begin{tabular}[t]{@{}c@{}}\textminus{}0.02\\[-1pt]{\scriptsize [\textminus{}0.11, +0.08]}\end{tabular} & \begin{tabular}[t]{@{}c@{}}+0.02\\[-1pt]{\scriptsize [\textminus{}0.08, +0.11]}\end{tabular} \\
 & Mistral-7B & 117 & \begin{tabular}[t]{@{}c@{}}\textbf{+0.23}\\[-1pt]{\scriptsize [+0.15, +0.32]}\end{tabular} & \begin{tabular}[t]{@{}c@{}}\textbf{+0.18}\\[-1pt]{\scriptsize [+0.10, +0.26]}\end{tabular} & \begin{tabular}[t]{@{}c@{}}\textbf{+0.10}\\[-1pt]{\scriptsize [+0.02, +0.20]}\end{tabular} & \begin{tabular}[t]{@{}c@{}}\textbf{+0.11}\\[-1pt]{\scriptsize [+0.03, +0.20]}\end{tabular} \\
\bottomrule
\end{tabular}
\caption{Paired gains over the token-level selector at a 5\% budget, with 95\% bootstrap intervals (point estimates in Table~\ref{tab:alloc}).}
\label{tab:alloc_ci5}
\end{table*}

\begin{table*}[t]
\centering
\footnotesize
\begin{tabular}{llccccc}
\toprule
Dataset & Model & $N$ & \begin{tabular}[b]{@{}c@{}}Target\\$-$ Token\end{tabular} & \begin{tabular}[b]{@{}c@{}}\begin{tabular}[b]{@{}c@{}}Target\\+4\end{tabular}\\$-$ Token\end{tabular} & \begin{tabular}[b]{@{}c@{}}Retrieved\\$-$ Token\end{tabular} & \begin{tabular}[b]{@{}c@{}}\begin{tabular}[b]{@{}c@{}}Retrieved\\+4\end{tabular}\\$-$ Token\end{tabular} \\
\midrule
LoCoMo & Qwen3-8B & 105 & \begin{tabular}[t]{@{}c@{}}\textminus{}0.04\\[-1pt]{\scriptsize [\textminus{}0.14, +0.06]}\end{tabular} & \begin{tabular}[t]{@{}c@{}}\textminus{}0.04\\[-1pt]{\scriptsize [\textminus{}0.15, +0.05]}\end{tabular} & \begin{tabular}[t]{@{}c@{}}\textbf{\textminus{}0.10}\\[-1pt]{\scriptsize [\textminus{}0.20, \textminus{}0.01]}\end{tabular} & \begin{tabular}[t]{@{}c@{}}\textbf{\textminus{}0.09}\\[-1pt]{\scriptsize [\textminus{}0.18, \textminus{}0.01]}\end{tabular} \\
 & Llama-3.1-8B & 102 & \begin{tabular}[t]{@{}c@{}}+0.00\\[-1pt]{\scriptsize [\textminus{}0.06, +0.06]}\end{tabular} & \begin{tabular}[t]{@{}c@{}}\textminus{}0.02\\[-1pt]{\scriptsize [\textminus{}0.08, +0.04]}\end{tabular} & \begin{tabular}[t]{@{}c@{}}\textminus{}0.02\\[-1pt]{\scriptsize [\textminus{}0.08, +0.05]}\end{tabular} & \begin{tabular}[t]{@{}c@{}}\textminus{}0.05\\[-1pt]{\scriptsize [\textminus{}0.14, +0.04]}\end{tabular} \\
 & Mistral-7B & 36 & \begin{tabular}[t]{@{}c@{}}+0.00\\[-1pt]{\scriptsize [\textminus{}0.11, +0.12]}\end{tabular} & \begin{tabular}[t]{@{}c@{}}+0.00\\[-1pt]{\scriptsize [\textminus{}0.18, +0.16]}\end{tabular} & \begin{tabular}[t]{@{}c@{}}+0.00\\[-1pt]{\scriptsize [\textminus{}0.11, +0.11]}\end{tabular} & \begin{tabular}[t]{@{}c@{}}+0.08\\[-1pt]{\scriptsize [\textminus{}0.06, +0.24]}\end{tabular} \\
\midrule
NQ-Open & Qwen3-8B & 146 & \begin{tabular}[t]{@{}c@{}}\textbf{+0.27}\\[-1pt]{\scriptsize [+0.20, +0.36]}\end{tabular} & \begin{tabular}[t]{@{}c@{}}\textbf{+0.33}\\[-1pt]{\scriptsize [+0.25, +0.41]}\end{tabular} & \begin{tabular}[t]{@{}c@{}}\textbf{+0.14}\\[-1pt]{\scriptsize [+0.07, +0.23]}\end{tabular} & \begin{tabular}[t]{@{}c@{}}\textbf{+0.24}\\[-1pt]{\scriptsize [+0.16, +0.32]}\end{tabular} \\
 & Llama-3.1-8B & 128 & \begin{tabular}[t]{@{}c@{}}\textminus{}0.02\\[-1pt]{\scriptsize [\textminus{}0.12, +0.06]}\end{tabular} & \begin{tabular}[t]{@{}c@{}}+0.00\\[-1pt]{\scriptsize [\textminus{}0.09, +0.09]}\end{tabular} & \begin{tabular}[t]{@{}c@{}}\textbf{\textminus{}0.09}\\[-1pt]{\scriptsize [\textminus{}0.19, \textminus{}0.01]}\end{tabular} & \begin{tabular}[t]{@{}c@{}}\textminus{}0.07\\[-1pt]{\scriptsize [\textminus{}0.17, +0.02]}\end{tabular} \\
 & Mistral-7B & 101 & \begin{tabular}[t]{@{}c@{}}\textbf{+0.25}\\[-1pt]{\scriptsize [+0.14, +0.36]}\end{tabular} & \begin{tabular}[t]{@{}c@{}}\textbf{+0.25}\\[-1pt]{\scriptsize [+0.14, +0.37]}\end{tabular} & \begin{tabular}[t]{@{}c@{}}+0.08\\[-1pt]{\scriptsize [\textminus{}0.03, +0.20]}\end{tabular} & \begin{tabular}[t]{@{}c@{}}\textbf{+0.14}\\[-1pt]{\scriptsize [+0.02, +0.26]}\end{tabular} \\
\midrule
Gorilla & Qwen3-8B & 89 & \begin{tabular}[t]{@{}c@{}}+0.03\\[-1pt]{\scriptsize [\textminus{}0.06, +0.13]}\end{tabular} & \begin{tabular}[t]{@{}c@{}}+0.02\\[-1pt]{\scriptsize [\textminus{}0.06, +0.11]}\end{tabular} & \begin{tabular}[t]{@{}c@{}}\textminus{}0.04\\[-1pt]{\scriptsize [\textminus{}0.13, +0.06]}\end{tabular} & \begin{tabular}[t]{@{}c@{}}+0.00\\[-1pt]{\scriptsize [\textminus{}0.09, +0.09]}\end{tabular} \\
 & Llama-3.1-8B & 100 & \begin{tabular}[t]{@{}c@{}}\textminus{}0.09\\[-1pt]{\scriptsize [\textminus{}0.19, +0.00]}\end{tabular} & \begin{tabular}[t]{@{}c@{}}\textbf{\textminus{}0.11}\\[-1pt]{\scriptsize [\textminus{}0.20, \textminus{}0.02]}\end{tabular} & \begin{tabular}[t]{@{}c@{}}\textbf{\textminus{}0.17}\\[-1pt]{\scriptsize [\textminus{}0.27, \textminus{}0.08]}\end{tabular} & \begin{tabular}[t]{@{}c@{}}\textbf{\textminus{}0.17}\\[-1pt]{\scriptsize [\textminus{}0.27, \textminus{}0.08]}\end{tabular} \\
 & Mistral-7B & 108 & \begin{tabular}[t]{@{}c@{}}\textbf{+0.30}\\[-1pt]{\scriptsize [+0.19, +0.41]}\end{tabular} & \begin{tabular}[t]{@{}c@{}}\textbf{+0.23}\\[-1pt]{\scriptsize [+0.12, +0.35]}\end{tabular} & \begin{tabular}[t]{@{}c@{}}+0.00\\[-1pt]{\scriptsize [\textminus{}0.11, +0.11]}\end{tabular} & \begin{tabular}[t]{@{}c@{}}+0.09\\[-1pt]{\scriptsize [\textminus{}0.01, +0.19]}\end{tabular} \\
\midrule
TriviaQA & Qwen3-8B & 151 & \begin{tabular}[t]{@{}c@{}}+0.03\\[-1pt]{\scriptsize [\textminus{}0.05, +0.11]}\end{tabular} & \begin{tabular}[t]{@{}c@{}}+0.08\\[-1pt]{\scriptsize [+0.00, +0.16]}\end{tabular} & \begin{tabular}[t]{@{}c@{}}\textminus{}0.03\\[-1pt]{\scriptsize [\textminus{}0.11, +0.05]}\end{tabular} & \begin{tabular}[t]{@{}c@{}}+0.03\\[-1pt]{\scriptsize [\textminus{}0.05, +0.11]}\end{tabular} \\
 & Llama-3.1-8B & 127 & \begin{tabular}[t]{@{}c@{}}+0.02\\[-1pt]{\scriptsize [\textminus{}0.08, +0.11]}\end{tabular} & \begin{tabular}[t]{@{}c@{}}+0.04\\[-1pt]{\scriptsize [\textminus{}0.05, +0.13]}\end{tabular} & \begin{tabular}[t]{@{}c@{}}\textbf{\textminus{}0.11}\\[-1pt]{\scriptsize [\textminus{}0.20, \textminus{}0.02]}\end{tabular} & \begin{tabular}[t]{@{}c@{}}\textminus{}0.06\\[-1pt]{\scriptsize [\textminus{}0.16, +0.03]}\end{tabular} \\
 & Mistral-7B & 117 & \begin{tabular}[t]{@{}c@{}}\textbf{+0.15}\\[-1pt]{\scriptsize [+0.06, +0.24]}\end{tabular} & \begin{tabular}[t]{@{}c@{}}\textbf{+0.13}\\[-1pt]{\scriptsize [+0.04, +0.21]}\end{tabular} & \begin{tabular}[t]{@{}c@{}}+0.03\\[-1pt]{\scriptsize [\textminus{}0.08, +0.13]}\end{tabular} & \begin{tabular}[t]{@{}c@{}}+0.07\\[-1pt]{\scriptsize [\textminus{}0.03, +0.16]}\end{tabular} \\
\bottomrule
\end{tabular}
\caption{Paired gains over the token-level selector at a 10\% budget, with 95\% bootstrap intervals (point estimates in Table~\ref{tab:alloc}).}
\label{tab:alloc_ci10}
\end{table*}

\begin{table*}[t]
\centering
\footnotesize
\begin{tabular}{llccccc}
\toprule
Dataset & Model & $N$ & \begin{tabular}[b]{@{}c@{}}Target+4 $-$\\Target (5\%)\end{tabular} & \begin{tabular}[b]{@{}c@{}}Target+4 $-$\\Target (10\%)\end{tabular} & \begin{tabular}[b]{@{}c@{}}Target+4 $-$\\random (10\%)\end{tabular} & \begin{tabular}[b]{@{}c@{}}Similar\\$-$ Token (10\%)\end{tabular} \\
\midrule
LoCoMo & Qwen3-8B & 105 & \begin{tabular}[t]{@{}c@{}}\textminus{}0.02\\[-1pt]{\scriptsize [\textminus{}0.06, +0.04]}\end{tabular} & \begin{tabular}[t]{@{}c@{}}+0.00\\[-1pt]{\scriptsize [\textminus{}0.05, +0.05]}\end{tabular} & \begin{tabular}[t]{@{}c@{}}\textminus{}0.01\\[-1pt]{\scriptsize [\textminus{}0.07, +0.03]}\end{tabular} & \begin{tabular}[t]{@{}c@{}}\textbf{\textminus{}0.14}\\[-1pt]{\scriptsize [\textminus{}0.21, \textminus{}0.07]}\end{tabular} \\
 & Llama-3.1-8B & 102 & \begin{tabular}[t]{@{}c@{}}+0.00\\[-1pt]{\scriptsize [\textminus{}0.05, +0.05]}\end{tabular} & \begin{tabular}[t]{@{}c@{}}\textminus{}0.02\\[-1pt]{\scriptsize [\textminus{}0.06, +0.02]}\end{tabular} & \begin{tabular}[t]{@{}c@{}}\textbf{\textminus{}0.05}\\[-1pt]{\scriptsize [\textminus{}0.08, \textminus{}0.02]}\end{tabular} & \begin{tabular}[t]{@{}c@{}}\textbf{\textminus{}0.13}\\[-1pt]{\scriptsize [\textminus{}0.21, \textminus{}0.04]}\end{tabular} \\
 & Mistral-7B & 36 & \begin{tabular}[t]{@{}c@{}}\textminus{}0.06\\[-1pt]{\scriptsize [\textminus{}0.17, +0.05]}\end{tabular} & \begin{tabular}[t]{@{}c@{}}+0.00\\[-1pt]{\scriptsize [\textminus{}0.13, +0.10]}\end{tabular} & \begin{tabular}[t]{@{}c@{}}+0.00\\[-1pt]{\scriptsize [\textminus{}0.13, +0.10]}\end{tabular} & \begin{tabular}[t]{@{}c@{}}\textminus{}0.03\\[-1pt]{\scriptsize [\textminus{}0.13, +0.06]}\end{tabular} \\
\midrule
NQ-Open & Qwen3-8B & 146 & \begin{tabular}[t]{@{}c@{}}\textminus{}0.05\\[-1pt]{\scriptsize [\textminus{}0.12, +0.02]}\end{tabular} & \begin{tabular}[t]{@{}c@{}}+0.05\\[-1pt]{\scriptsize [+0.00, +0.11]}\end{tabular} & \begin{tabular}[t]{@{}c@{}}+0.05\\[-1pt]{\scriptsize [\textminus{}0.01, +0.11]}\end{tabular} & \begin{tabular}[t]{@{}c@{}}\textbf{\textminus{}0.09}\\[-1pt]{\scriptsize [\textminus{}0.14, \textminus{}0.03]}\end{tabular} \\
 & Llama-3.1-8B & 128 & \begin{tabular}[t]{@{}c@{}}+0.03\\[-1pt]{\scriptsize [\textminus{}0.02, +0.09]}\end{tabular} & \begin{tabular}[t]{@{}c@{}}+0.02\\[-1pt]{\scriptsize [\textminus{}0.03, +0.08]}\end{tabular} & \begin{tabular}[t]{@{}c@{}}\textminus{}0.03\\[-1pt]{\scriptsize [\textminus{}0.10, +0.04]}\end{tabular} & \begin{tabular}[t]{@{}c@{}}\textbf{\textminus{}0.20}\\[-1pt]{\scriptsize [\textminus{}0.29, \textminus{}0.12]}\end{tabular} \\
 & Mistral-7B & 101 & \begin{tabular}[t]{@{}c@{}}\textbf{\textminus{}0.10}\\[-1pt]{\scriptsize [\textminus{}0.19, \textminus{}0.01]}\end{tabular} & \begin{tabular}[t]{@{}c@{}}+0.00\\[-1pt]{\scriptsize [\textminus{}0.07, +0.07]}\end{tabular} & \begin{tabular}[t]{@{}c@{}}\textminus{}0.01\\[-1pt]{\scriptsize [\textminus{}0.09, +0.07]}\end{tabular} & \begin{tabular}[t]{@{}c@{}}\textbf{\textminus{}0.16}\\[-1pt]{\scriptsize [\textminus{}0.26, \textminus{}0.06]}\end{tabular} \\
\midrule
Gorilla & Qwen3-8B & 89 & \begin{tabular}[t]{@{}c@{}}\textbf{\textminus{}0.09}\\[-1pt]{\scriptsize [\textminus{}0.17, \textminus{}0.01]}\end{tabular} & \begin{tabular}[t]{@{}c@{}}\textminus{}0.01\\[-1pt]{\scriptsize [\textminus{}0.09, +0.07]}\end{tabular} & \begin{tabular}[t]{@{}c@{}}\textminus{}0.04\\[-1pt]{\scriptsize [\textminus{}0.12, +0.03]}\end{tabular} & \begin{tabular}[t]{@{}c@{}}\textbf{\textminus{}0.16}\\[-1pt]{\scriptsize [\textminus{}0.24, \textminus{}0.08]}\end{tabular} \\
 & Llama-3.1-8B & 100 & \begin{tabular}[t]{@{}c@{}}\textminus{}0.04\\[-1pt]{\scriptsize [\textminus{}0.10, +0.02]}\end{tabular} & \begin{tabular}[t]{@{}c@{}}\textminus{}0.02\\[-1pt]{\scriptsize [\textminus{}0.08, +0.04]}\end{tabular} & \begin{tabular}[t]{@{}c@{}}\textminus{}0.04\\[-1pt]{\scriptsize [\textminus{}0.12, +0.04]}\end{tabular} & \begin{tabular}[t]{@{}c@{}}\textbf{\textminus{}0.29}\\[-1pt]{\scriptsize [\textminus{}0.39, \textminus{}0.20]}\end{tabular} \\
 & Mistral-7B & 108 & \begin{tabular}[t]{@{}c@{}}\textbf{\textminus{}0.15}\\[-1pt]{\scriptsize [\textminus{}0.23, \textminus{}0.07]}\end{tabular} & \begin{tabular}[t]{@{}c@{}}\textminus{}0.06\\[-1pt]{\scriptsize [\textminus{}0.16, +0.02]}\end{tabular} & \begin{tabular}[t]{@{}c@{}}+0.07\\[-1pt]{\scriptsize [\textminus{}0.01, +0.17]}\end{tabular} & \begin{tabular}[t]{@{}c@{}}\textbf{\textminus{}0.24}\\[-1pt]{\scriptsize [\textminus{}0.33, \textminus{}0.16]}\end{tabular} \\
\midrule
TriviaQA & Qwen3-8B & 151 & \begin{tabular}[t]{@{}c@{}}+0.03\\[-1pt]{\scriptsize [\textminus{}0.03, +0.09]}\end{tabular} & \begin{tabular}[t]{@{}c@{}}+0.05\\[-1pt]{\scriptsize [+0.00, +0.10]}\end{tabular} & \begin{tabular}[t]{@{}c@{}}+0.00\\[-1pt]{\scriptsize [\textminus{}0.05, +0.05]}\end{tabular} & \begin{tabular}[t]{@{}c@{}}\textbf{\textminus{}0.23}\\[-1pt]{\scriptsize [\textminus{}0.32, \textminus{}0.15]}\end{tabular} \\
 & Llama-3.1-8B & 127 & \begin{tabular}[t]{@{}c@{}}\textminus{}0.02\\[-1pt]{\scriptsize [\textminus{}0.08, +0.05]}\end{tabular} & \begin{tabular}[t]{@{}c@{}}+0.02\\[-1pt]{\scriptsize [\textminus{}0.04, +0.09]}\end{tabular} & \begin{tabular}[t]{@{}c@{}}\textminus{}0.02\\[-1pt]{\scriptsize [\textminus{}0.08, +0.05]}\end{tabular} & \begin{tabular}[t]{@{}c@{}}\textbf{\textminus{}0.32}\\[-1pt]{\scriptsize [\textminus{}0.41, \textminus{}0.24]}\end{tabular} \\
 & Mistral-7B & 117 & \begin{tabular}[t]{@{}c@{}}\textminus{}0.05\\[-1pt]{\scriptsize [\textminus{}0.12, +0.02]}\end{tabular} & \begin{tabular}[t]{@{}c@{}}\textminus{}0.03\\[-1pt]{\scriptsize [\textminus{}0.09, +0.03]}\end{tabular} & \begin{tabular}[t]{@{}c@{}}\textbf{\textminus{}0.07}\\[-1pt]{\scriptsize [\textminus{}0.13, \textminus{}0.01]}\end{tabular} & \begin{tabular}[t]{@{}c@{}}\textbf{\textminus{}0.31}\\[-1pt]{\scriptsize [\textminus{}0.39, \textminus{}0.21]}\end{tabular} \\
\bottomrule
\end{tabular}
\caption{Controls for the chunk-level allocation (paired differences in retention on $Q^*$). Target+4 $-$ Target: the four most similar chunks added to the target, at 5\% and 10\%. Target+4 $-$ random: the target with its four most similar chunks against the target with four random chunks (10\%). Similar $-$ Token: the four most similar chunks without the target against the token-level selector (10\%).}
\label{tab:alloc_controls}
\end{table*}

\begin{figure*}[t]
\centering
\includegraphics[width=\textwidth]{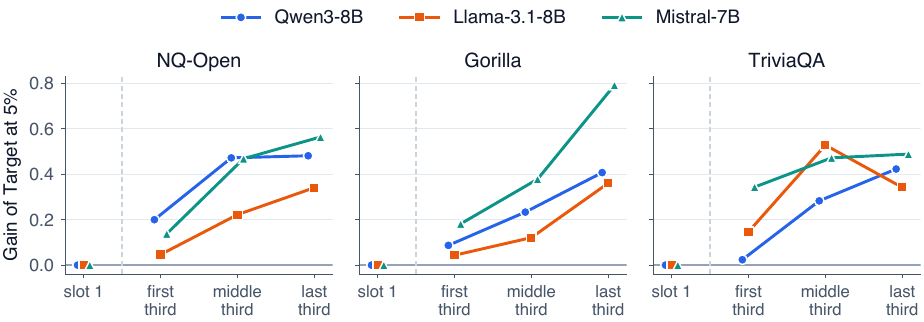}
\caption{Change in retention on \Qs{} from recomputing the known target chunk at 5\% instead of applying no repair, by the slot of the target (Table~\ref{tab:position}). In slot 1 the recomputed KV equals the reused KV. Targets in later slots miss more of their context under reuse, and recomputing them raises retention more on average.}
\label{fig:position}
\end{figure*}

\paragraph{Allocation controls.} Table~\ref{tab:alloc_controls} adds two controls. Recomputing the four most similar chunks without the target is significantly worse than the token-level selector in eleven of twelve settings. Replacing these four chunks in Target+4 with four random chunks changes retention significantly in two settings, both times in favour of the random chunks. At 5\%, Target+4 shares the budget among five chunks, so the target is no longer guaranteed to be recomputed in full. On Gorilla, Target+4 at 10\% changes retention relative to the token-level selector by +0.23 for Mistral-7B and by \textminus{}0.11 for Llama-3.1-8B, so we report it separately for each model.
\paragraph{The first slot.} By Proposition~\ref{prop:layer}, recomputing a target in the first slot reproduces the KV that it already has, and the outputs of all 46 such questions are unchanged. Figure~\ref{fig:position} shows the change in retention from recomputing the target when the target is in a later slot.

\begin{table*}[t]
\centering
\footnotesize
\begin{tabular}{llccccccc}
\toprule
 & & & \multicolumn{2}{c}{First third} & \multicolumn{2}{c}{Middle third} & \multicolumn{2}{c}{Last third} \\ \cmidrule(lr){4-5}\cmidrule(lr){6-7}\cmidrule(lr){8-9}
Dataset & Model & Slot 1 & $N$ & Gain & $N$ & Gain & $N$ & Gain \\
\midrule
NQ-Open & Qwen3-8B & 6 & 50 & +0.20 & 36 & +0.47 & 54 & +0.48 \\
 & Llama-3.1-8B & 5 & 43 & +0.05 & 36 & +0.22 & 44 & +0.34 \\
 & Mistral-7B & 1 & 29 & +0.14 & 32 & +0.47 & 39 & +0.56 \\
\midrule
Gorilla & Qwen3-8B & 9 & 23 & +0.09 & 30 & +0.23 & 27 & +0.41 \\
 & Llama-3.1-8B & 8 & 23 & +0.04 & 33 & +0.12 & 36 & +0.36 \\
 & Mistral-7B & 7 & 11 & +0.18 & 37 & +0.38 & 53 & +0.79 \\
\midrule
TriviaQA & Qwen3-8B & 4 & 42 & +0.02 & 53 & +0.28 & 52 & +0.42 \\
 & Llama-3.1-8B & 2 & 34 & +0.15 & 53 & +0.53 & 38 & +0.34 \\
 & Mistral-7B & 4 & 32 & +0.34 & 36 & +0.47 & 45 & +0.49 \\
\bottomrule
\end{tabular}
\caption{Change in retention on \Qs{} from recomputing the target chunk (Target, 5\%) instead of applying no repair, by the slot of the target in the prompt. Slot 1: the number of questions whose target is the first chunk after the instruction prefix; for these questions the recomputed KV equals the reused KV and the output does not change. The other questions are split into thirds of the prompt by the slot of the target; $N$ gives their number and Gain the change in retention.}
\label{tab:position}
\end{table*}

\begin{table*}[t]
\centering
\footnotesize
\setlength{\tabcolsep}{5pt}
\begin{tabular}{llccccccc}
\toprule
Dataset & Model & $N$ & \begin{tabular}[b]{@{}c@{}}Token 5\%\\(own rule)\end{tabular} & \begin{tabular}[b]{@{}c@{}}Token 5\%\\(ProphetKV)\end{tabular} & \begin{tabular}[b]{@{}c@{}}Target\\$-$ Token\end{tabular} & \begin{tabular}[b]{@{}c@{}}Target+4\\$-$ Token\end{tabular} & \begin{tabular}[b]{@{}c@{}}Retrieved\\$-$ Token\end{tabular} & \begin{tabular}[b]{@{}c@{}}Retrieved\\+4 $-$ Token\end{tabular} \\
\midrule
LoCoMo & Mistral-7B & 36 & 0.78 & 0.69 & \begin{tabular}[t]{@{}c@{}}\textbf{+0.11}\\[-1pt]{\scriptsize [+0.02, +0.24]}\end{tabular} & \begin{tabular}[t]{@{}c@{}}+0.08\\[-1pt]{\scriptsize [\textminus{}0.06, +0.24]}\end{tabular} & \begin{tabular}[t]{@{}c@{}}+0.06\\[-1pt]{\scriptsize [\textminus{}0.05, +0.16]}\end{tabular} & \begin{tabular}[t]{@{}c@{}}+0.08\\[-1pt]{\scriptsize [\textminus{}0.05, +0.23]}\end{tabular} \\
Gorilla & Qwen3-8B & 89 & 0.06 & 0.09 & \begin{tabular}[t]{@{}c@{}}\textbf{+0.13}\\[-1pt]{\scriptsize [+0.04, +0.22]}\end{tabular} & \begin{tabular}[t]{@{}c@{}}\textbf{+0.15}\\[-1pt]{\scriptsize [+0.07, +0.24]}\end{tabular} & \begin{tabular}[t]{@{}c@{}}+0.06\\[-1pt]{\scriptsize [\textminus{}0.02, +0.13]}\end{tabular} & \begin{tabular}[t]{@{}c@{}}\textbf{+0.12}\\[-1pt]{\scriptsize [+0.04, +0.20]}\end{tabular} \\
Gorilla & Llama-3.1-8B & 100 & 0.12 & 0.07 & \begin{tabular}[t]{@{}c@{}}\textbf{+0.16}\\[-1pt]{\scriptsize [+0.08, +0.25]}\end{tabular} & \begin{tabular}[t]{@{}c@{}}\textbf{+0.12}\\[-1pt]{\scriptsize [+0.04, +0.20]}\end{tabular} & \begin{tabular}[t]{@{}c@{}}\textbf{+0.08}\\[-1pt]{\scriptsize [+0.01, +0.16]}\end{tabular} & \begin{tabular}[t]{@{}c@{}}+0.07\\[-1pt]{\scriptsize [\textminus{}0.01, +0.14]}\end{tabular} \\
Gorilla & Mistral-7B & 108 & 0.10 & 0.12 & \begin{tabular}[t]{@{}c@{}}\textbf{+0.46}\\[-1pt]{\scriptsize [+0.35, +0.56]}\end{tabular} & \begin{tabular}[t]{@{}c@{}}\textbf{+0.35}\\[-1pt]{\scriptsize [+0.24, +0.46]}\end{tabular} & \begin{tabular}[t]{@{}c@{}}\textbf{+0.18}\\[-1pt]{\scriptsize [+0.08, +0.27]}\end{tabular} & \begin{tabular}[t]{@{}c@{}}\textbf{+0.19}\\[-1pt]{\scriptsize [+0.10, +0.29]}\end{tabular} \\
TriviaQA & Mistral-7B & 117 & 0.44 & 0.38 & \begin{tabular}[t]{@{}c@{}}\textbf{+0.30}\\[-1pt]{\scriptsize [+0.21, +0.38]}\end{tabular} & \begin{tabular}[t]{@{}c@{}}\textbf{+0.24}\\[-1pt]{\scriptsize [+0.15, +0.32]}\end{tabular} & \begin{tabular}[t]{@{}c@{}}\textbf{+0.17}\\[-1pt]{\scriptsize [+0.09, +0.26]}\end{tabular} & \begin{tabular}[t]{@{}c@{}}\textbf{+0.18}\\[-1pt]{\scriptsize [+0.09, +0.26]}\end{tabular} \\
\bottomrule
\end{tabular}
\caption{The 5\% allocation of Table~\ref{tab:alloc} with ProphetKV as the token-level rule, in the settings where the rule selected on development questions is not ProphetKV. Token (own rule) repeats the baseline of Table~\ref{tab:alloc}; the gains (with 95\% bootstrap intervals) are over the ProphetKV baseline. A chosen chunk that exceeds the budget has its tokens ranked by ProphetKV here; on LoCoMo the session exceeds it on 32 of the 36 questions. Target and Retrieved nevertheless give the same scores as in Table~\ref{tab:alloc}, except for 3 scores, each on a question where a chosen chunk exceeds the budget.}
\label{tab:uniform}
\end{table*}

\begin{table*}[t]
\centering
\footnotesize
\begin{tabular}{llccccccc}
\toprule
 & & & \multicolumn{3}{c}{5\% budget} & \multicolumn{3}{c}{10\% budget} \\ \cmidrule(lr){4-6}\cmidrule(lr){7-9}
Dataset & Model & $N$ & \begin{tabular}[b]{@{}c@{}}Retrieved\\+4 (\%)\end{tabular} & \begin{tabular}[b]{@{}c@{}}Gated $-$\\Token\end{tabular} & \begin{tabular}[b]{@{}c@{}}Gated $-$\\Retrieved+4\end{tabular} & \begin{tabular}[b]{@{}c@{}}Retrieved\\+4 (\%)\end{tabular} & \begin{tabular}[b]{@{}c@{}}Gated $-$\\Token\end{tabular} & \begin{tabular}[b]{@{}c@{}}Gated $-$\\Retrieved+4\end{tabular} \\
\midrule
LoCoMo & Qwen3-8B & 105 & 0 & \begin{tabular}[t]{@{}c@{}}+0.00\\[-1pt]{\scriptsize [+0.00, +0.00]}\end{tabular} & \begin{tabular}[t]{@{}c@{}}\textbf{+0.10}\\[-1pt]{\scriptsize [+0.04, +0.20]}\end{tabular} & 1 & \begin{tabular}[t]{@{}c@{}}\textminus{}0.01\\[-1pt]{\scriptsize [\textminus{}0.03, +0.00]}\end{tabular} & \begin{tabular}[t]{@{}c@{}}\textbf{+0.08}\\[-1pt]{\scriptsize [+0.01, +0.16]}\end{tabular} \\
 & Llama-3.1-8B & 102 & 0 & \begin{tabular}[t]{@{}c@{}}+0.00\\[-1pt]{\scriptsize [+0.00, +0.00]}\end{tabular} & \begin{tabular}[t]{@{}c@{}}+0.05\\[-1pt]{\scriptsize [\textminus{}0.02, +0.13]}\end{tabular} & 1 & \begin{tabular}[t]{@{}c@{}}+0.00\\[-1pt]{\scriptsize [+0.00, +0.00]}\end{tabular} & \begin{tabular}[t]{@{}c@{}}+0.05\\[-1pt]{\scriptsize [\textminus{}0.04, +0.14]}\end{tabular} \\
 & Mistral-7B & 36 & 0 & \begin{tabular}[t]{@{}c@{}}+0.00\\[-1pt]{\scriptsize [+0.00, +0.00]}\end{tabular} & \begin{tabular}[t]{@{}c@{}}\textminus{}0.03\\[-1pt]{\scriptsize [\textminus{}0.19, +0.12]}\end{tabular} & 3 & \begin{tabular}[t]{@{}c@{}}+0.00\\[-1pt]{\scriptsize [+0.00, +0.00]}\end{tabular} & \begin{tabular}[t]{@{}c@{}}\textminus{}0.08\\[-1pt]{\scriptsize [\textminus{}0.24, +0.06]}\end{tabular} \\
\midrule
NQ-Open & Qwen3-8B & 146 & 79 & \begin{tabular}[t]{@{}c@{}}\textbf{+0.17}\\[-1pt]{\scriptsize [+0.11, +0.24]}\end{tabular} & \begin{tabular}[t]{@{}c@{}}\textminus{}0.03\\[-1pt]{\scriptsize [\textminus{}0.08, +0.00]}\end{tabular} & 84 & \begin{tabular}[t]{@{}c@{}}\textbf{+0.23}\\[-1pt]{\scriptsize [+0.16, +0.32]}\end{tabular} & \begin{tabular}[t]{@{}c@{}}\textminus{}0.01\\[-1pt]{\scriptsize [\textminus{}0.03, +0.01]}\end{tabular} \\
 & Llama-3.1-8B & 128 & 77 & \begin{tabular}[t]{@{}c@{}}\textbf{+0.11}\\[-1pt]{\scriptsize [+0.02, +0.20]}\end{tabular} & \begin{tabular}[t]{@{}c@{}}\textminus{}0.02\\[-1pt]{\scriptsize [\textminus{}0.04, +0.00]}\end{tabular} & 84 & \begin{tabular}[t]{@{}c@{}}\textminus{}0.08\\[-1pt]{\scriptsize [\textminus{}0.16, +0.01]}\end{tabular} & \begin{tabular}[t]{@{}c@{}}\textminus{}0.01\\[-1pt]{\scriptsize [\textminus{}0.05, +0.03]}\end{tabular} \\
 & Mistral-7B & 101 & 79 & \begin{tabular}[t]{@{}c@{}}\textbf{+0.13}\\[-1pt]{\scriptsize [+0.03, +0.23]}\end{tabular} & \begin{tabular}[t]{@{}c@{}}\textminus{}0.02\\[-1pt]{\scriptsize [\textminus{}0.08, +0.03]}\end{tabular} & 84 & \begin{tabular}[t]{@{}c@{}}\textbf{+0.12}\\[-1pt]{\scriptsize [+0.02, +0.23]}\end{tabular} & \begin{tabular}[t]{@{}c@{}}\textminus{}0.02\\[-1pt]{\scriptsize [\textminus{}0.07, +0.03]}\end{tabular} \\
\midrule
Gorilla & Qwen3-8B & 89 & 93 & \begin{tabular}[t]{@{}c@{}}\textbf{+0.07}\\[-1pt]{\scriptsize [+0.01, +0.12]}\end{tabular} & \begin{tabular}[t]{@{}c@{}}\textminus{}0.01\\[-1pt]{\scriptsize [\textminus{}0.03, +0.00]}\end{tabular} & 25 & \begin{tabular}[t]{@{}c@{}}\textminus{}0.01\\[-1pt]{\scriptsize [\textminus{}0.06, +0.03]}\end{tabular} & \begin{tabular}[t]{@{}c@{}}\textminus{}0.01\\[-1pt]{\scriptsize [\textminus{}0.09, +0.07]}\end{tabular} \\
 & Llama-3.1-8B & 100 & 91 & \begin{tabular}[t]{@{}c@{}}+0.00\\[-1pt]{\scriptsize [\textminus{}0.08, +0.08]}\end{tabular} & \begin{tabular}[t]{@{}c@{}}\textminus{}0.01\\[-1pt]{\scriptsize [\textminus{}0.03, +0.00]}\end{tabular} & 23 & \begin{tabular}[t]{@{}c@{}}\textbf{\textminus{}0.07}\\[-1pt]{\scriptsize [\textminus{}0.13, \textminus{}0.03]}\end{tabular} & \begin{tabular}[t]{@{}c@{}}\textbf{+0.10}\\[-1pt]{\scriptsize [+0.02, +0.18]}\end{tabular} \\
 & Mistral-7B & 108 & 93 & \begin{tabular}[t]{@{}c@{}}\textbf{+0.20}\\[-1pt]{\scriptsize [+0.10, +0.31]}\end{tabular} & \begin{tabular}[t]{@{}c@{}}\textminus{}0.01\\[-1pt]{\scriptsize [\textminus{}0.03, +0.00]}\end{tabular} & 21 & \begin{tabular}[t]{@{}c@{}}\textbf{+0.08}\\[-1pt]{\scriptsize [+0.02, +0.15]}\end{tabular} & \begin{tabular}[t]{@{}c@{}}\textminus{}0.01\\[-1pt]{\scriptsize [\textminus{}0.08, +0.07]}\end{tabular} \\
\midrule
TriviaQA & Qwen3-8B & 151 & 87 & \begin{tabular}[t]{@{}c@{}}\textbf{+0.15}\\[-1pt]{\scriptsize [+0.07, +0.23]}\end{tabular} & \begin{tabular}[t]{@{}c@{}}\textminus{}0.02\\[-1pt]{\scriptsize [\textminus{}0.05, +0.01]}\end{tabular} & 48 & \begin{tabular}[t]{@{}c@{}}\textminus{}0.03\\[-1pt]{\scriptsize [\textminus{}0.09, +0.03]}\end{tabular} & \begin{tabular}[t]{@{}c@{}}\textminus{}0.05\\[-1pt]{\scriptsize [\textminus{}0.11, +0.01]}\end{tabular} \\
 & Llama-3.1-8B & 127 & 87 & \begin{tabular}[t]{@{}c@{}}+0.01\\[-1pt]{\scriptsize [\textminus{}0.08, +0.09]}\end{tabular} & \begin{tabular}[t]{@{}c@{}}\textminus{}0.01\\[-1pt]{\scriptsize [\textminus{}0.05, +0.02]}\end{tabular} & 47 & \begin{tabular}[t]{@{}c@{}}\textbf{\textminus{}0.06}\\[-1pt]{\scriptsize [\textminus{}0.13, \textminus{}0.01]}\end{tabular} & \begin{tabular}[t]{@{}c@{}}+0.00\\[-1pt]{\scriptsize [\textminus{}0.07, +0.07]}\end{tabular} \\
 & Mistral-7B & 117 & 91 & \begin{tabular}[t]{@{}c@{}}\textbf{+0.09}\\[-1pt]{\scriptsize [+0.01, +0.16]}\end{tabular} & \begin{tabular}[t]{@{}c@{}}\textminus{}0.03\\[-1pt]{\scriptsize [\textminus{}0.07, +0.01]}\end{tabular} & 54 & \begin{tabular}[t]{@{}c@{}}+0.03\\[-1pt]{\scriptsize [\textminus{}0.03, +0.10]}\end{tabular} & \begin{tabular}[t]{@{}c@{}}\textminus{}0.03\\[-1pt]{\scriptsize [\textminus{}0.11, +0.03]}\end{tabular} \\
\bottomrule
\end{tabular}
\caption{A gate that recomputes Retrieved+4 when the cosine similarity of the retriever\textquoteright s top-ranked chunk exceeds that of its second-ranked chunk by at least a threshold, and uses the token-level selector otherwise. The questions are split into two halves (by conversation for LoCoMo). A threshold fitted on each half, pooled over the three models of a dataset, is applied to the other half, so every question is evaluated once. Retrieved+4 (\%): share of questions sent to Retrieved+4; differences are paired, with 95\% bootstrap intervals that keep the gate\textquoteright s decisions fixed.}
\label{tab:hybrid}
\end{table*}

\begin{table*}[t]
\centering
\footnotesize
\setlength{\tabcolsep}{3.5pt}
\begin{tabular}{lcccccccc}
\toprule
 & \multicolumn{4}{c}{TTFT, ms: median (p95)} & \multicolumn{4}{c}{Ratio to full prefill} \\ \cmidrule(lr){2-5}\cmidrule(lr){6-9}
Condition & \begin{tabular}[b]{@{}c@{}}LoCoMo\\(10.8K)\end{tabular} & \begin{tabular}[b]{@{}c@{}}NQ-Open\\(4.6K)\end{tabular} & \begin{tabular}[b]{@{}c@{}}Gorilla\\(5.7K)\end{tabular} & \begin{tabular}[b]{@{}c@{}}TriviaQA\\(4.7K)\end{tabular} & LoCoMo & NQ-Open & Gorilla & TriviaQA \\
\midrule
Full prefill & 2554 (3136) & 840 (898) & 1072 (1938) & 842 (959) & 1.00 & 1.00 & 1.00 & 1.00 \\
Reuse & 94 (120) & 53 (66) & 58 (81) & 53 (57) & 0.04 & 0.06 & 0.06 & 0.06 \\
Token 5\% & 309 (363) & 162 (198) & 129 (189) & 165 (176) & 0.12 & 0.19 & 0.12 & 0.19 \\
Target 5\% & 268 (324) & 92 (114) & 103 (149) & 94 (104) & 0.11 & 0.11 & 0.09 & 0.11 \\
Retrieved 5\% & 282 (352) & 105 (130) & 117 (165) & 108 (114) & 0.11 & 0.13 & 0.11 & 0.13 \\
Retrieved+4 5\% & 316 (368) & 169 (183) & 137 (190) & 173 (186) & 0.12 & 0.20 & 0.13 & 0.20 \\
Token 10\% & 415 (510) & 198 (237) & 173 (263) & 200 (215) & 0.16 & 0.23 & 0.16 & 0.23 \\
Retrieved+4 10\% & 417 (491) & 202 (237) & 177 (252) & 206 (223) & 0.17 & 0.24 & 0.16 & 0.24 \\
\bottomrule
\end{tabular}
\caption{Time to first token (TTFT) with Qwen3-8B in our engine on one RTX 4090 D, measured on the first 50 \Qs{} questions of each dataset (taken from the second question set for NQ-Open and Gorilla). The time for each question is the median of three runs; the table gives the median over questions and, in parentheses, the 95th percentile (p95). The median context length of these questions is given in parentheses below each dataset name. Reuse assembles stored chunk caches (RoPE shift and concatenation); the repairs add selection (question-attention scoring where the rule needs it, and BGE-M3 question encoding and chunk ranking for Retrieved) and recomputation. All conditions use the same attention kernels; an optimised serving system would change the absolute times.}
\label{tab:timing}
\end{table*}

\begin{table*}[t]
\centering
\scriptsize
\setlength{\tabcolsep}{3pt}
\begin{tabular}{llccccccccc}
\toprule
 & & & & & \multicolumn{6}{c}{Retrieved chunk, 5\% budget} \\ \cmidrule(lr){6-11}
Dataset & Model & $N$ & \begin{tabular}[b]{@{}c@{}}Hit\\(\%)\end{tabular} & \begin{tabular}[b]{@{}c@{}}Token\\5\%\end{tabular} & \begin{tabular}[b]{@{}c@{}}In place\\$-$ Token\end{tabular} & \begin{tabular}[b]{@{}c@{}}Moved\\$-$ Token\end{tabular} & \begin{tabular}[b]{@{}c@{}}Moved only,\\no recomp.\end{tabular} & \begin{tabular}[b]{@{}c@{}}Moved $-$\\in place\end{tabular} & \begin{tabular}[b]{@{}c@{}}Moved $-$ Token\\(hits)\end{tabular} & \begin{tabular}[b]{@{}c@{}}Moved $-$ Token\\(misses)\end{tabular} \\
\midrule
NQ-Open & Qwen3-8B & 146 & 55 & 0.16 & \begin{tabular}[t]{@{}c@{}}\textbf{+0.22}\\[-1pt]{\scriptsize [+0.14, +0.29]}\end{tabular} & \begin{tabular}[t]{@{}c@{}}\textbf{+0.40}\\[-1pt]{\scriptsize [+0.31, +0.49]}\end{tabular} & \begin{tabular}[t]{@{}c@{}}+0.00\\[-1pt]{\scriptsize [\textminus{}0.02, +0.02]}\end{tabular} & \begin{tabular}[t]{@{}c@{}}\textbf{+0.18}\\[-1pt]{\scriptsize [+0.11, +0.25]}\end{tabular} & \begin{tabular}[t]{@{}c@{}}\textbf{+0.75}\\[-1pt]{\scriptsize [+0.65, +0.85]}\end{tabular} & \begin{tabular}[t]{@{}c@{}}\textminus{}0.03\\[-1pt]{\scriptsize [\textminus{}0.12, +0.06]}\end{tabular} \\
 & Llama-3.1-8B & 128 & 47 & 0.21 & \begin{tabular}[t]{@{}c@{}}+0.08\\[-1pt]{\scriptsize [+0.00, +0.16]}\end{tabular} & \begin{tabular}[t]{@{}c@{}}\textbf{+0.25}\\[-1pt]{\scriptsize [+0.15, +0.35]}\end{tabular} & \begin{tabular}[t]{@{}c@{}}+0.04\\[-1pt]{\scriptsize [\textminus{}0.01, +0.09]}\end{tabular} & \begin{tabular}[t]{@{}c@{}}\textbf{+0.17}\\[-1pt]{\scriptsize [+0.10, +0.24]}\end{tabular} & \begin{tabular}[t]{@{}c@{}}\textbf{+0.63}\\[-1pt]{\scriptsize [+0.50, +0.75]}\end{tabular} & \begin{tabular}[t]{@{}c@{}}\textminus{}0.09\\[-1pt]{\scriptsize [\textminus{}0.18, +0.00]}\end{tabular} \\
 & Mistral-7B & 101 & 50 & 0.28 & \begin{tabular}[t]{@{}c@{}}\textbf{+0.14}\\[-1pt]{\scriptsize [+0.02, +0.25]}\end{tabular} & \begin{tabular}[t]{@{}c@{}}\textbf{+0.27}\\[-1pt]{\scriptsize [+0.14, +0.39]}\end{tabular} & \begin{tabular}[t]{@{}c@{}}\textbf{+0.09}\\[-1pt]{\scriptsize [+0.02, +0.16]}\end{tabular} & \begin{tabular}[t]{@{}c@{}}\textbf{+0.13}\\[-1pt]{\scriptsize [+0.03, +0.22]}\end{tabular} & \begin{tabular}[t]{@{}c@{}}\textbf{+0.65}\\[-1pt]{\scriptsize [+0.49, +0.78]}\end{tabular} & \begin{tabular}[t]{@{}c@{}}\textminus{}0.12\\[-1pt]{\scriptsize [\textminus{}0.26, +0.00]}\end{tabular} \\
\midrule
Gorilla & Qwen3-8B & 89 & 53 & 0.06 & \begin{tabular}[t]{@{}c@{}}+0.09\\[-1pt]{\scriptsize [+0.00, +0.18]}\end{tabular} & \begin{tabular}[t]{@{}c@{}}\textbf{+0.21}\\[-1pt]{\scriptsize [+0.12, +0.30]}\end{tabular} & \begin{tabular}[t]{@{}c@{}}+0.02\\[-1pt]{\scriptsize [+0.00, +0.06]}\end{tabular} & \begin{tabular}[t]{@{}c@{}}\textbf{+0.12}\\[-1pt]{\scriptsize [+0.03, +0.22]}\end{tabular} & \begin{tabular}[t]{@{}c@{}}\textbf{+0.43}\\[-1pt]{\scriptsize [+0.28, +0.57]}\end{tabular} & \begin{tabular}[t]{@{}c@{}}\textminus{}0.02\\[-1pt]{\scriptsize [\textminus{}0.07, +0.00]}\end{tabular} \\
 & Llama-3.1-8B & 100 & 53 & 0.12 & \begin{tabular}[t]{@{}c@{}}+0.02\\[-1pt]{\scriptsize [\textminus{}0.07, +0.10]}\end{tabular} & \begin{tabular}[t]{@{}c@{}}\textbf{+0.15}\\[-1pt]{\scriptsize [+0.06, +0.24]}\end{tabular} & \begin{tabular}[t]{@{}c@{}}+0.00\\[-1pt]{\scriptsize [\textminus{}0.04, +0.04]}\end{tabular} & \begin{tabular}[t]{@{}c@{}}\textbf{+0.13}\\[-1pt]{\scriptsize [+0.04, +0.22]}\end{tabular} & \begin{tabular}[t]{@{}c@{}}\textbf{+0.34}\\[-1pt]{\scriptsize [+0.23, +0.47]}\end{tabular} & \begin{tabular}[t]{@{}c@{}}\textminus{}0.06\\[-1pt]{\scriptsize [\textminus{}0.17, +0.02]}\end{tabular} \\
 & Mistral-7B & 108 & 47 & 0.10 & \begin{tabular}[t]{@{}c@{}}\textbf{+0.19}\\[-1pt]{\scriptsize [+0.09, +0.30]}\end{tabular} & \begin{tabular}[t]{@{}c@{}}\textbf{+0.32}\\[-1pt]{\scriptsize [+0.21, +0.44]}\end{tabular} & \begin{tabular}[t]{@{}c@{}}+0.01\\[-1pt]{\scriptsize [\textminus{}0.03, +0.05]}\end{tabular} & \begin{tabular}[t]{@{}c@{}}\textbf{+0.13}\\[-1pt]{\scriptsize [+0.04, +0.22]}\end{tabular} & \begin{tabular}[t]{@{}c@{}}\textbf{+0.84}\\[-1pt]{\scriptsize [+0.75, +0.94]}\end{tabular} & \begin{tabular}[t]{@{}c@{}}\textbf{\textminus{}0.14}\\[-1pt]{\scriptsize [\textminus{}0.23, \textminus{}0.05]}\end{tabular} \\
\midrule
TriviaQA & Qwen3-8B & 151 & 64 & 0.32 & \begin{tabular}[t]{@{}c@{}}\textbf{+0.14}\\[-1pt]{\scriptsize [+0.06, +0.22]}\end{tabular} & \begin{tabular}[t]{@{}c@{}}\textbf{+0.28}\\[-1pt]{\scriptsize [+0.19, +0.36]}\end{tabular} & \begin{tabular}[t]{@{}c@{}}\textbf{\textminus{}0.05}\\[-1pt]{\scriptsize [\textminus{}0.09, \textminus{}0.01]}\end{tabular} & \begin{tabular}[t]{@{}c@{}}\textbf{+0.14}\\[-1pt]{\scriptsize [+0.07, +0.22]}\end{tabular} & \begin{tabular}[t]{@{}c@{}}\textbf{+0.46}\\[-1pt]{\scriptsize [+0.36, +0.57]}\end{tabular} & \begin{tabular}[t]{@{}c@{}}\textminus{}0.06\\[-1pt]{\scriptsize [\textminus{}0.15, +0.04]}\end{tabular} \\
 & Llama-3.1-8B & 127 & 60 & 0.46 & \begin{tabular}[t]{@{}c@{}}\textminus{}0.02\\[-1pt]{\scriptsize [\textminus{}0.11, +0.08]}\end{tabular} & \begin{tabular}[t]{@{}c@{}}\textbf{+0.14}\\[-1pt]{\scriptsize [+0.04, +0.24]}\end{tabular} & \begin{tabular}[t]{@{}c@{}}\textbf{+0.13}\\[-1pt]{\scriptsize [+0.06, +0.20]}\end{tabular} & \begin{tabular}[t]{@{}c@{}}\textbf{+0.16}\\[-1pt]{\scriptsize [+0.09, +0.23]}\end{tabular} & \begin{tabular}[t]{@{}c@{}}\textbf{+0.39}\\[-1pt]{\scriptsize [+0.28, +0.51]}\end{tabular} & \begin{tabular}[t]{@{}c@{}}\textbf{\textminus{}0.24}\\[-1pt]{\scriptsize [\textminus{}0.37, \textminus{}0.10]}\end{tabular} \\
 & Mistral-7B & 117 & 62 & 0.44 & \begin{tabular}[t]{@{}c@{}}\textbf{+0.10}\\[-1pt]{\scriptsize [+0.02, +0.20]}\end{tabular} & \begin{tabular}[t]{@{}c@{}}\textbf{+0.17}\\[-1pt]{\scriptsize [+0.07, +0.27]}\end{tabular} & \begin{tabular}[t]{@{}c@{}}+0.02\\[-1pt]{\scriptsize [\textminus{}0.04, +0.09]}\end{tabular} & \begin{tabular}[t]{@{}c@{}}\textbf{+0.07}\\[-1pt]{\scriptsize [+0.01, +0.14]}\end{tabular} & \begin{tabular}[t]{@{}c@{}}\textbf{+0.42}\\[-1pt]{\scriptsize [+0.32, +0.53]}\end{tabular} & \begin{tabular}[t]{@{}c@{}}\textbf{\textminus{}0.25}\\[-1pt]{\scriptsize [\textminus{}0.39, \textminus{}0.11]}\end{tabular} \\
\bottomrule
\end{tabular}
\caption{Recomputing a chunk at the 5\% budget after moving it to the last slot before the question (headers renumbered, chunks encoded alone as before). In place: the chunk recomputed where the retriever found it (Retrieved in Table~\ref{tab:alloc}). Moved only: the reordered cache without recomputation, against the unrepaired cache. Hits and misses: questions on which the retriever does or does not rank the target first. Paired differences with 95\% bootstrap intervals.}
\label{tab:reorder}
\end{table*}

\begin{table*}[t]
\centering
\footnotesize
\setlength{\tabcolsep}{3pt}
\begin{tabular}{llcccccccc}
\toprule
 & & & \multicolumn{3}{c}{Retention, 5\% budget} & & \multicolumn{3}{c}{Chunk moved $-$ Token moved} \\ \cmidrule(lr){4-6}\cmidrule(lr){8-10}
Dataset & Model & $N$ & \begin{tabular}[b]{@{}c@{}}Token\\original\end{tabular} & \begin{tabular}[b]{@{}c@{}}Token\\moved\end{tabular} & \begin{tabular}[b]{@{}c@{}}Chunk\\moved\end{tabular} & \begin{tabular}[b]{@{}c@{}}Token moved\\$-$ Token\end{tabular} & All & Hits & Misses \\
\midrule
NQ-Open & Qwen3-8B & 146 & 0.16 & 0.27 & 0.56 & \begin{tabular}[t]{@{}c@{}}\textbf{+0.11}\\[-1pt]{\scriptsize [+0.05, +0.17]}\end{tabular} & \begin{tabular}[t]{@{}c@{}}\textbf{+0.29}\\[-1pt]{\scriptsize [+0.21, +0.37]}\end{tabular} & \begin{tabular}[t]{@{}c@{}}\textbf{+0.56}\\[-1pt]{\scriptsize [+0.45, +0.68]}\end{tabular} & \begin{tabular}[t]{@{}c@{}}\textminus{}0.05\\[-1pt]{\scriptsize [\textminus{}0.12, +0.03]}\end{tabular} \\
 & Llama-3.1-8B & 128 & 0.21 & 0.46 & 0.46 & \begin{tabular}[t]{@{}c@{}}\textbf{+0.25}\\[-1pt]{\scriptsize [+0.16, +0.34]}\end{tabular} & \begin{tabular}[t]{@{}c@{}}+0.00\\[-1pt]{\scriptsize [\textminus{}0.09, +0.09]}\end{tabular} & \begin{tabular}[t]{@{}c@{}}+0.08\\[-1pt]{\scriptsize [\textminus{}0.05, +0.22]}\end{tabular} & \begin{tabular}[t]{@{}c@{}}\textminus{}0.07\\[-1pt]{\scriptsize [\textminus{}0.18, +0.03]}\end{tabular} \\
 & Mistral-7B & 101 & 0.28 & 0.51 & 0.54 & \begin{tabular}[t]{@{}c@{}}\textbf{+0.24}\\[-1pt]{\scriptsize [+0.13, +0.34]}\end{tabular} & \begin{tabular}[t]{@{}c@{}}+0.03\\[-1pt]{\scriptsize [\textminus{}0.08, +0.13]}\end{tabular} & \begin{tabular}[t]{@{}c@{}}\textbf{+0.22}\\[-1pt]{\scriptsize [+0.08, +0.35]}\end{tabular} & \begin{tabular}[t]{@{}c@{}}\textbf{\textminus{}0.16}\\[-1pt]{\scriptsize [\textminus{}0.30, \textminus{}0.02]}\end{tabular} \\
\midrule
Gorilla & Qwen3-8B & 89 & 0.06 & 0.06 & 0.27 & \begin{tabular}[t]{@{}c@{}}+0.00\\[-1pt]{\scriptsize [\textminus{}0.03, +0.03]}\end{tabular} & \begin{tabular}[t]{@{}c@{}}\textbf{+0.21}\\[-1pt]{\scriptsize [+0.12, +0.31]}\end{tabular} & \begin{tabular}[t]{@{}c@{}}\textbf{+0.45}\\[-1pt]{\scriptsize [+0.30, +0.60]}\end{tabular} & \begin{tabular}[t]{@{}c@{}}\textminus{}0.05\\[-1pt]{\scriptsize [\textminus{}0.12, +0.00]}\end{tabular} \\
 & Llama-3.1-8B & 100 & 0.12 & 0.34 & 0.27 & \begin{tabular}[t]{@{}c@{}}\textbf{+0.22}\\[-1pt]{\scriptsize [+0.12, +0.32]}\end{tabular} & \begin{tabular}[t]{@{}c@{}}\textminus{}0.07\\[-1pt]{\scriptsize [\textminus{}0.16, +0.02]}\end{tabular} & \begin{tabular}[t]{@{}c@{}}\textminus{}0.09\\[-1pt]{\scriptsize [\textminus{}0.25, +0.06]}\end{tabular} & \begin{tabular}[t]{@{}c@{}}\textminus{}0.04\\[-1pt]{\scriptsize [\textminus{}0.13, +0.04]}\end{tabular} \\
 & Mistral-7B & 108 & 0.10 & 0.12 & 0.43 & \begin{tabular}[t]{@{}c@{}}+0.02\\[-1pt]{\scriptsize [\textminus{}0.03, +0.07]}\end{tabular} & \begin{tabular}[t]{@{}c@{}}\textbf{+0.31}\\[-1pt]{\scriptsize [+0.19, +0.42]}\end{tabular} & \begin{tabular}[t]{@{}c@{}}\textbf{+0.82}\\[-1pt]{\scriptsize [+0.73, +0.92]}\end{tabular} & \begin{tabular}[t]{@{}c@{}}\textbf{\textminus{}0.16}\\[-1pt]{\scriptsize [\textminus{}0.26, \textminus{}0.07]}\end{tabular} \\
\midrule
TriviaQA & Qwen3-8B & 151 & 0.32 & 0.52 & 0.60 & \begin{tabular}[t]{@{}c@{}}\textbf{+0.19}\\[-1pt]{\scriptsize [+0.12, +0.26]}\end{tabular} & \begin{tabular}[t]{@{}c@{}}\textbf{+0.09}\\[-1pt]{\scriptsize [+0.02, +0.15]}\end{tabular} & \begin{tabular}[t]{@{}c@{}}\textbf{+0.15}\\[-1pt]{\scriptsize [+0.07, +0.25]}\end{tabular} & \begin{tabular}[t]{@{}c@{}}\textminus{}0.04\\[-1pt]{\scriptsize [\textminus{}0.13, +0.07]}\end{tabular} \\
 & Llama-3.1-8B & 127 & 0.46 & 0.61 & 0.60 & \begin{tabular}[t]{@{}c@{}}\textbf{+0.16}\\[-1pt]{\scriptsize [+0.06, +0.24]}\end{tabular} & \begin{tabular}[t]{@{}c@{}}\textminus{}0.02\\[-1pt]{\scriptsize [\textminus{}0.09, +0.06]}\end{tabular} & \begin{tabular}[t]{@{}c@{}}+0.09\\[-1pt]{\scriptsize [+0.00, +0.18]}\end{tabular} & \begin{tabular}[t]{@{}c@{}}\textbf{\textminus{}0.18}\\[-1pt]{\scriptsize [\textminus{}0.29, \textminus{}0.08]}\end{tabular} \\
 & Mistral-7B & 117 & 0.44 & 0.57 & 0.62 & \begin{tabular}[t]{@{}c@{}}\textbf{+0.13}\\[-1pt]{\scriptsize [+0.04, +0.21]}\end{tabular} & \begin{tabular}[t]{@{}c@{}}+0.04\\[-1pt]{\scriptsize [\textminus{}0.03, +0.12]}\end{tabular} & \begin{tabular}[t]{@{}c@{}}\textbf{+0.16}\\[-1pt]{\scriptsize [+0.08, +0.26]}\end{tabular} & \begin{tabular}[t]{@{}c@{}}\textbf{\textminus{}0.16}\\[-1pt]{\scriptsize [\textminus{}0.30, \textminus{}0.05]}\end{tabular} \\
\bottomrule
\end{tabular}
\caption{Token-level selection after the move. On the prompt with the retriever\textquoteright s top chunk moved to the last slot (Table~\ref{tab:reorder}), Token moved spends the 5\% budget with the setting\textquoteright s token-level rule over all chunks, and Chunk moved recomputes the moved chunk. Token: the token-level selector on the original prompt. Paired differences with 95\% bootstrap intervals.}
\label{tab:reorder_tok}
\end{table*}

\begin{table*}[t]
\centering
\scriptsize
\setlength{\tabcolsep}{1.4pt}
\begin{tabular}{llccccccccccc}
\toprule
 & & & \multicolumn{2}{c}{Hit (\%)} & \multicolumn{4}{c}{Retention} & \multicolumn{2}{c}{Chunk moved $-$ InfoFlow style} & & \\ \cmidrule(lr){4-5}\cmidrule(lr){6-9}\cmidrule(lr){10-11}
Dataset & Model & $N$ & Retrieved & InfoFlow & \begin{tabular}[b]{@{}c@{}}Token\\5\%\end{tabular} & \begin{tabular}[b]{@{}c@{}}InfoFlow\\5\%\end{tabular} & \begin{tabular}[b]{@{}c@{}}InfoFlow\\15\%\end{tabular} & \begin{tabular}[b]{@{}c@{}}Chunk\\moved 5\%\end{tabular} & at 5\% & at 15\% & \begin{tabular}[b]{@{}c@{}}InfoFlow 5\%\\$-$ Token\end{tabular} & \begin{tabular}[b]{@{}c@{}}InfoFlow chunk\\$-$ Chunk moved\end{tabular} \\
\midrule
NQ-Open & Qwen3-8B & 146 & 55 & 52 & 0.16 & 0.40 & 0.66 & 0.56 & \begin{tabular}[t]{@{}c@{}}\textbf{+0.16}\\[-1pt]{\scriptsize [+0.07, +0.26]}\end{tabular} & \begin{tabular}[t]{@{}c@{}}\textminus{}0.10\\[-1pt]{\scriptsize [\textminus{}0.20, +0.00]}\end{tabular} & \begin{tabular}[t]{@{}c@{}}\textbf{+0.23}\\[-1pt]{\scriptsize [+0.16, +0.32]}\end{tabular} & \begin{tabular}[t]{@{}c@{}}\textminus{}0.01\\[-1pt]{\scriptsize [\textminus{}0.11, +0.08]}\end{tabular} \\
 & Llama-3.1-8B & 128 & 47 & 39 & 0.21 & 0.40 & 0.59 & 0.46 & \begin{tabular}[t]{@{}c@{}}+0.06\\[-1pt]{\scriptsize [\textminus{}0.04, +0.16]}\end{tabular} & \begin{tabular}[t]{@{}c@{}}\textbf{\textminus{}0.12}\\[-1pt]{\scriptsize [\textminus{}0.22, \textminus{}0.02]}\end{tabular} & \begin{tabular}[t]{@{}c@{}}\textbf{+0.19}\\[-1pt]{\scriptsize [+0.10, +0.27]}\end{tabular} & \begin{tabular}[t]{@{}c@{}}\textminus{}0.08\\[-1pt]{\scriptsize [\textminus{}0.17, +0.02]}\end{tabular} \\
 & Mistral-7B & 101 & 50 & 43 & 0.28 & 0.38 & 0.53 & 0.54 & \begin{tabular}[t]{@{}c@{}}\textbf{+0.17}\\[-1pt]{\scriptsize [+0.02, +0.31]}\end{tabular} & \begin{tabular}[t]{@{}c@{}}+0.01\\[-1pt]{\scriptsize [\textminus{}0.13, +0.16]}\end{tabular} & \begin{tabular}[t]{@{}c@{}}+0.10\\[-1pt]{\scriptsize [+0.00, +0.20]}\end{tabular} & \begin{tabular}[t]{@{}c@{}}\textminus{}0.07\\[-1pt]{\scriptsize [\textminus{}0.20, +0.07]}\end{tabular} \\
\midrule
Gorilla & Qwen3-8B & 89 & 53 & 24 & 0.06 & 0.10 & 0.28 & 0.27 & \begin{tabular}[t]{@{}c@{}}\textbf{+0.17}\\[-1pt]{\scriptsize [+0.07, +0.27]}\end{tabular} & \begin{tabular}[t]{@{}c@{}}\textminus{}0.01\\[-1pt]{\scriptsize [\textminus{}0.11, +0.09]}\end{tabular} & \begin{tabular}[t]{@{}c@{}}+0.04\\[-1pt]{\scriptsize [\textminus{}0.02, +0.11]}\end{tabular} & \begin{tabular}[t]{@{}c@{}}\textbf{\textminus{}0.13}\\[-1pt]{\scriptsize [\textminus{}0.24, \textminus{}0.03]}\end{tabular} \\
 & Llama-3.1-8B & 100 & 53 & 15 & 0.12 & 0.13 & 0.27 & 0.27 & \begin{tabular}[t]{@{}c@{}}\textbf{+0.14}\\[-1pt]{\scriptsize [+0.05, +0.23]}\end{tabular} & \begin{tabular}[t]{@{}c@{}}+0.00\\[-1pt]{\scriptsize [\textminus{}0.11, +0.11]}\end{tabular} & \begin{tabular}[t]{@{}c@{}}+0.01\\[-1pt]{\scriptsize [\textminus{}0.08, +0.10]}\end{tabular} & \begin{tabular}[t]{@{}c@{}}\textbf{\textminus{}0.15}\\[-1pt]{\scriptsize [\textminus{}0.24, \textminus{}0.06]}\end{tabular} \\
 & Mistral-7B & 108 & 47 & 12 & 0.10 & 0.08 & 0.14 & 0.43 & \begin{tabular}[t]{@{}c@{}}\textbf{+0.34}\\[-1pt]{\scriptsize [+0.25, +0.44]}\end{tabular} & \begin{tabular}[t]{@{}c@{}}\textbf{+0.29}\\[-1pt]{\scriptsize [+0.19, +0.40]}\end{tabular} & \begin{tabular}[t]{@{}c@{}}\textminus{}0.02\\[-1pt]{\scriptsize [\textminus{}0.09, +0.06]}\end{tabular} & \begin{tabular}[t]{@{}c@{}}\textbf{\textminus{}0.30}\\[-1pt]{\scriptsize [\textminus{}0.39, \textminus{}0.20]}\end{tabular} \\
\midrule
TriviaQA & Qwen3-8B & 151 & 64 & 51 & 0.32 & 0.60 & 0.78 & 0.60 & \begin{tabular}[t]{@{}c@{}}+0.00\\[-1pt]{\scriptsize [\textminus{}0.09, +0.09]}\end{tabular} & \begin{tabular}[t]{@{}c@{}}\textbf{\textminus{}0.18}\\[-1pt]{\scriptsize [\textminus{}0.25, \textminus{}0.11]}\end{tabular} & \begin{tabular}[t]{@{}c@{}}\textbf{+0.28}\\[-1pt]{\scriptsize [+0.20, +0.36]}\end{tabular} & \begin{tabular}[t]{@{}c@{}}\textbf{\textminus{}0.09}\\[-1pt]{\scriptsize [\textminus{}0.17, \textminus{}0.01]}\end{tabular} \\
 & Llama-3.1-8B & 127 & 60 & 37 & 0.46 & 0.51 & 0.66 & 0.60 & \begin{tabular}[t]{@{}c@{}}+0.09\\[-1pt]{\scriptsize [\textminus{}0.01, +0.18]}\end{tabular} & \begin{tabular}[t]{@{}c@{}}\textminus{}0.06\\[-1pt]{\scriptsize [\textminus{}0.16, +0.03]}\end{tabular} & \begin{tabular}[t]{@{}c@{}}+0.06\\[-1pt]{\scriptsize [\textminus{}0.03, +0.14]}\end{tabular} & \begin{tabular}[t]{@{}c@{}}\textbf{\textminus{}0.16}\\[-1pt]{\scriptsize [\textminus{}0.25, \textminus{}0.06]}\end{tabular} \\
 & Mistral-7B & 117 & 62 & 29 & 0.44 & 0.43 & 0.62 & 0.62 & \begin{tabular}[t]{@{}c@{}}\textbf{+0.19}\\[-1pt]{\scriptsize [+0.09, +0.29]}\end{tabular} & \begin{tabular}[t]{@{}c@{}}\textminus{}0.01\\[-1pt]{\scriptsize [\textminus{}0.10, +0.09]}\end{tabular} & \begin{tabular}[t]{@{}c@{}}\textminus{}0.02\\[-1pt]{\scriptsize [\textminus{}0.09, +0.06]}\end{tabular} & \begin{tabular}[t]{@{}c@{}}\textbf{\textminus{}0.25}\\[-1pt]{\scriptsize [\textminus{}0.35, \textminus{}0.15]}\end{tabular} \\
\bottomrule
\end{tabular}
\caption{Comparison with InfoFlow-KV-style reordering \citep{infoflowkv}, re-implemented in our engine at the 5\% budget. InfoFlow style: tokens are scored by the L2 norm of question attention (layers 22--25) on the cache with chunk-local positions; chunks are ordered by the fraction of their tokens among the top 5\%, with the highest fraction last; tokens are scored again on the reordered prompt; and the top 5\% are recomputed. Token 5\%: the token-level selector at 5\% on the original prompt. Chunk moved: the retriever\textquoteright s top chunk moved last and recomputed (Table~\ref{tab:reorder}). InfoFlow chunk: the last chunk of the InfoFlow order recomputed with the setting\textquoteright s rule. InfoFlow 15\%: the same procedure at its default budget of 15\% in both passes. Hit: the share of questions on which the chunk placed last at 5\% is the target. Differences are paired, with 95\% bootstrap intervals.}
\label{tab:infoflow}
\end{table*}

\paragraph{Moving the chosen chunk last.} We move either the retriever's top chunk (chosen in the original order) or the target to the last slot. All headers are renumbered by position, every chunk is encoded alone as before, and the moved chunk is recomputed under the 5\% budget with the setting's rule. Table~\ref{tab:reorder} gives the results for the retrieved chunk, and the results for the moved target are given below. Table~\ref{tab:reorder_tok} spends the same 5\% budget with the setting's token-level rule on the same reordered prompts. Recomputing the moved chunk uses 48--67\% as many tokens as this rule, as one contiguous block after the reused chunks. Because our headers carry position numbers, they are renumbered in the reordered prompts and every chunk is encoded alone again. With headers that carry no position, the stored caches could be reused in the new order without re-encoding.

Whether the moved chunk beats the selector on the reordered prompt depends on whether the retriever found the target. On hits, the questions where the retriever ranks the target first, recomputing the moved chunk gives significantly higher retention than the selector in six of the nine settings. On misses it gives lower retention everywhere, significantly so in four. On Gorilla with Qwen3-8B and Mistral-7B, the setting's rule is EPIC, whose selection does not depend on the order, so the gains of the moved chunk there are measured against EPIC. With ProphetKV as the token-level rule instead, the move raises the retention of the selector by +0.20 and +0.16 over ProphetKV on the original prompt, and the recomputed moved chunk beats this selector on the reordered prompt with Mistral-7B (+0.15 [+0.06, +0.23]) but not with Qwen3-8B (\textminus{}0.02 [\textminus{}0.11, +0.08]).

The moved target gains 0.18--0.41 over the target recomputed in place. Without recomputation, a late slot can help by itself \citep{lostmiddle}. Moving the target without recomputing it raises retention significantly, by 0.09 to 0.17, only with Llama-3.1-8B and Mistral-7B on NQ-Open and with Llama-3.1-8B on TriviaQA. Moving the retrieved chunk without recomputing it changes retention by \textminus{}0.05 to +0.13. On hits this is a significant gain in only two settings, and on misses never a significant loss.

Reordering also changes full prefill, which answers 79--98\% of \Qs{} on the reordered prompts, while retention is still measured against full prefill in the original order. Restricting the comparison to the questions that full prefill also answers in the moved order (83--98\% of \Qs{}) gives the same counts as on all of \Qs{}: the moved chunk significantly beats the selector in nine settings (by 0.12 to 0.40), the move significantly raises the retention of the selector in seven, and on the reordered prompt the moved chunk is significantly better than selected tokens in four settings and significantly worse in none.

\paragraph{InfoFlow-style reordering.} \citet{infoflowkv} also move chunks toward the question. We re-implemented the reordering procedure of their released code (commit 725b75c) in our engine (Table~\ref{tab:infoflow}). It scores each context token by the L2 norm of the question tokens' head-averaged attention on it in layers 22--25, the default of their run script for Qwen and Llama models, and we use the same layers for Mistral-7B. The first scoring pass reads the cache with chunk-local positions, as their paper (its HL--TP setting) and code do, so that no chunk is favoured for its current slot. The procedure then orders the chunks by the fraction of their tokens that are among the top-scoring tokens and places the chunk with the highest fraction last. Finally, it scores the tokens again on the reordered prompt and recomputes the top-scoring tokens. Their paper describes the score as a column sum of prompt attention; we follow the code. We use our 5\% budget (their default is 15\%) and our question suffix as the query. The comparison matches the token budget, not latency.

At this budget, moving the retriever's top chunk last and recomputing it beats the procedure significantly in six of the nine settings, by 0.14 to 0.34, and five of these differences remain significant after Holm's correction. In the other three, Llama-3.1-8B on NQ-Open and TriviaQA and Qwen3-8B on TriviaQA, the differences are +0.00 to +0.09, so the moved chunk is never significantly worse. The procedure itself improves on the token-level selector significantly in three settings, and on the same scoring applied without reordering in six. At its default budget of 15\% in both passes, the procedure recomputes 4.5--6.2 times as many tokens as the moved chunk (medians), and its retention rises by 0.06 to 0.26 over the 5\% version, significantly in eight of the nine. Even with this larger budget, the procedure is significantly better than the moved chunk at 5\% in only two settings and significantly worse in one (one of each after Holm's correction).

The difference between the two methods has two parts: which chunk is placed last, and whether all of its tokens are recomputed. The chunk that the procedure places last is the target on 12--52\% of the questions, against 47--64\% for the retriever's top chunk, and this share is significantly lower in six settings. Recomputing all tokens of that last chunk gives lower retention than recomputing the retriever's chunk everywhere, significantly so in six of the nine. On the questions where both orders put the target last (10 to 55 per setting), recomputing all tokens of the last chunk gives significantly different retention under the two orders in only one setting. On these questions, recomputing all tokens of the moved chunk still beats recomputing the procedure's selected tokens significantly in six settings, by 0.11 to 0.70. On the prompts with the retriever's chunk moved last, the token-level rules based on question attention spend 13--24\% of the budget on the moved chunk (one of 30 or 31 chunks), whereas EPIC spends about 3\%.

The order also affects full prefill, which answers 58--95\% of \Qs{} in the InfoFlow order and 83--98\% with the retriever's chunk moved. On the questions that full prefill answers in both reordered prompts (52--93\% of \Qs{}), the moved chunk still scores higher than the procedure in eight of the nine settings and the same in the remaining one (+0.00 to +0.55), significantly higher in six. Recomputing all tokens of the last chunk of the InfoFlow order again gives lower retention than recomputing the moved chunk in all nine settings, significantly so in six.

\begin{table*}[t]
\centering
\footnotesize
\setlength{\tabcolsep}{1.9pt}
\begin{tabular}{llcccccccccc}
\toprule
Dataset & Model & $N$ & \begin{tabular}[b]{@{}c@{}}Hit (\%)\\top-1/3\end{tabular} & \begin{tabular}[b]{@{}c@{}}$R_{\mathrm{hit}}$\\$R_{\mathrm{miss}}$\end{tabular} & \begin{tabular}[b]{@{}c@{}}$k{=}1$\\chunk\end{tabular} & \begin{tabular}[b]{@{}c@{}}$k{=}3$\\chunks\end{tabular} & \begin{tabular}[b]{@{}c@{}}$k{=}3$\\tokens\end{tabular} & \begin{tabular}[b]{@{}c@{}}$k{=}3-k{=}1$\\chunks\end{tabular} & \begin{tabular}[b]{@{}c@{}}Predicted\\$k{=}3-k{=}1$\end{tabular} & \begin{tabular}[b]{@{}c@{}}Chunks $-$\\tokens, $k{=}3$\end{tabular} & \begin{tabular}[b]{@{}c@{}}$k{=}3$ chunks $-$\\Token, orig.\end{tabular} \\
\midrule
NQ-Open & Qwen3-8B & 146 & 55/75 & \begin{tabular}[t]{@{}c@{}}0.93\\0.12\end{tabular} & 0.56 & 0.67 & 0.50 & \begin{tabular}[t]{@{}c@{}}\textbf{+0.11}\\[-1pt]{\scriptsize [+0.04, +0.18]}\end{tabular} & +0.16 & \begin{tabular}[t]{@{}c@{}}\textbf{+0.17}\\[-1pt]{\scriptsize [+0.09, +0.25]}\end{tabular} & \begin{tabular}[t]{@{}c@{}}\textbf{+0.43}\\[-1pt]{\scriptsize [+0.34, +0.53]}\end{tabular} \\
 & Llama-3.1-8B & 128 & 47/70 & \begin{tabular}[t]{@{}c@{}}0.83\\0.13\end{tabular} & 0.46 & 0.52 & 0.65 & \begin{tabular}[t]{@{}c@{}}+0.06\\[-1pt]{\scriptsize [\textminus{}0.02, +0.14]}\end{tabular} & +0.16 & \begin{tabular}[t]{@{}c@{}}\textbf{\textminus{}0.12}\\[-1pt]{\scriptsize [\textminus{}0.20, \textminus{}0.05]}\end{tabular} & \begin{tabular}[t]{@{}c@{}}\textbf{+0.14}\\[-1pt]{\scriptsize [+0.04, +0.23]}\end{tabular} \\
 & Mistral-7B & 101 & 50/71 & \begin{tabular}[t]{@{}c@{}}0.92\\0.16\end{tabular} & 0.54 & 0.66 & 0.66 & \begin{tabular}[t]{@{}c@{}}\textbf{+0.12}\\[-1pt]{\scriptsize [+0.03, +0.21]}\end{tabular} & +0.16 & \begin{tabular}[t]{@{}c@{}}+0.00\\[-1pt]{\scriptsize [\textminus{}0.09, +0.09]}\end{tabular} & \begin{tabular}[t]{@{}c@{}}\textbf{+0.33}\\[-1pt]{\scriptsize [+0.21, +0.46]}\end{tabular} \\
\midrule
TriviaQA & Qwen3-8B & 151 & 64/85 & \begin{tabular}[t]{@{}c@{}}0.84\\0.19\end{tabular} & 0.60 & 0.74 & 0.69 & \begin{tabular}[t]{@{}c@{}}\textbf{+0.13}\\[-1pt]{\scriptsize [+0.07, +0.20]}\end{tabular} & +0.14 & \begin{tabular}[t]{@{}c@{}}+0.05\\[-1pt]{\scriptsize [\textminus{}0.03, +0.12]}\end{tabular} & \begin{tabular}[t]{@{}c@{}}\textbf{+0.25}\\[-1pt]{\scriptsize [+0.15, +0.34]}\end{tabular} \\
 & Llama-3.1-8B & 127 & 60/79 & \begin{tabular}[t]{@{}c@{}}0.89\\0.16\end{tabular} & 0.60 & 0.69 & 0.74 & \begin{tabular}[t]{@{}c@{}}\textbf{+0.09}\\[-1pt]{\scriptsize [+0.01, +0.17]}\end{tabular} & +0.14 & \begin{tabular}[t]{@{}c@{}}\textminus{}0.06\\[-1pt]{\scriptsize [\textminus{}0.13, +0.02]}\end{tabular} & \begin{tabular}[t]{@{}c@{}}\textbf{+0.13}\\[-1pt]{\scriptsize [+0.03, +0.24]}\end{tabular} \\
 & Mistral-7B & 117 & 62/85 & \begin{tabular}[t]{@{}c@{}}0.88\\0.18\end{tabular} & 0.62 & 0.77 & 0.74 & \begin{tabular}[t]{@{}c@{}}\textbf{+0.15}\\[-1pt]{\scriptsize [+0.09, +0.22]}\end{tabular} & +0.15 & \begin{tabular}[t]{@{}c@{}}+0.03\\[-1pt]{\scriptsize [\textminus{}0.04, +0.09]}\end{tabular} & \begin{tabular}[t]{@{}c@{}}\textbf{+0.25}\\[-1pt]{\scriptsize [+0.15, +0.35]}\end{tabular} \\
\midrule
Gorilla & Qwen3-8B & 89 & 53/74 & \begin{tabular}[t]{@{}c@{}}0.47\\0.05\end{tabular} & 0.28 & 0.44 & 0.20 & \begin{tabular}[t]{@{}c@{}}\textbf{+0.16}\\[-1pt]{\scriptsize [+0.07, +0.26]}\end{tabular} & +0.09 & \begin{tabular}[t]{@{}c@{}}\textbf{+0.24}\\[-1pt]{\scriptsize [+0.12, +0.36]}\end{tabular} & \begin{tabular}[t]{@{}c@{}}\textbf{+0.25}\\[-1pt]{\scriptsize [+0.15, +0.36]}\end{tabular} \\
 & Llama-3.1-8B & 100 & 53/69 & \begin{tabular}[t]{@{}c@{}}0.47\\0.04\end{tabular} & 0.27 & 0.37 & 0.51 & \begin{tabular}[t]{@{}c@{}}+0.10\\[-1pt]{\scriptsize [+0.00, +0.20]}\end{tabular} & +0.07 & \begin{tabular}[t]{@{}c@{}}\textbf{\textminus{}0.14}\\[-1pt]{\scriptsize [\textminus{}0.23, \textminus{}0.04]}\end{tabular} & \begin{tabular}[t]{@{}c@{}}+0.06\\[-1pt]{\scriptsize [\textminus{}0.05, +0.17]}\end{tabular} \\
 & Mistral-7B & 108 & 47/72 & \begin{tabular}[t]{@{}c@{}}0.90\\0.00\end{tabular} & 0.43 & 0.56 & 0.28 & \begin{tabular}[t]{@{}c@{}}\textbf{+0.13}\\[-1pt]{\scriptsize [+0.04, +0.22]}\end{tabular} & +0.23 & \begin{tabular}[t]{@{}c@{}}\textbf{+0.28}\\[-1pt]{\scriptsize [+0.17, +0.39]}\end{tabular} & \begin{tabular}[t]{@{}c@{}}\textbf{+0.26}\\[-1pt]{\scriptsize [+0.13, +0.39]}\end{tabular} \\
\bottomrule
\end{tabular}
\caption{Moving the retriever\textquoteright s top-$k$ chunks to the slots before the question, in reverse order of rank (top-1 last), and recomputing them at the 10\% budget. Hit: share of questions whose target is the top-1 chunk or among the top three. $R_{\mathrm{hit}}$ (upper) and $R_{\mathrm{miss}}$ (lower): retention of the moved top-1 chunk at 5\% on the questions where it is and is not the target (Table~\ref{tab:reorder}), the values the prediction uses. Chunks: the $k$ moved chunks recomputed with the setting\textquoteright s rule; tokens: the setting\textquoteright s token-level rule over all chunks on the same reordered prompt. Predicted: the gain predicted from the hit rates and from the retention of the moved top-1 chunk on hits and misses (Appendix~\ref{app:topk}). Token, orig.: the token-level selector at 10\% on the original prompt. Paired 95\% bootstrap intervals; bold intervals exclude zero.}
\label{tab:topk}
\end{table*}

\paragraph{Moving the top three chunks.}
\label{app:topk}
Let $h_k$ be the share of questions whose target is among the top $k$ chunks, and $R_{\mathrm{hit}}$ and $R_{\mathrm{miss}}$ the retention of the moved top chunk at 5\% on the questions where it is and is not the target. If moving $k$ chunks gives retention $R_{\mathrm{hit}}$ when the target is among them and $R_{\mathrm{miss}}$ otherwise, the gain of moving three chunks over one is $(h_3-h_1)(R_{\mathrm{hit}}-R_{\mathrm{miss}})$. The predictions in Table~\ref{tab:topk} use the values of $R_{\mathrm{hit}}$ and $R_{\mathrm{miss}}$ listed there, which come from the experiments of Table~\ref{tab:reorder}, and range from +0.07 to +0.23. The chunks are ranked by the BGE-M3 cosine similarity with the question alone. The top three are moved to the last three slots in reverse order of rank, so that the top one is last; the other chunks keep their order, the headers are renumbered, and every chunk is encoded alone as before. With one chunk moved, the prompt is the same as in Table~\ref{tab:reorder}, and the results at 5\% are those of Tables~\ref{tab:reorder} and~\ref{tab:reorder_tok}. Full prefill of the reordered prompts answers 68--97\% of \Qs{}, and retention is still measured against full prefill in the original order.

When the moved chunks hold more tokens than the budget, the tokens within them are ranked by the setting's rule. At 10\%, the three chunks fit entirely within the budget on 41--73\% of the questions. With one chunk moved and a 10\% budget, only that chunk is recomputed and the rest of the budget is left unspent, as for Target and Retrieved in Table~\ref{tab:alloc}. Its scores then equal those at 5\% in Table~\ref{tab:reorder} on every question except one, on Gorilla with Qwen3-8B, where the moved chunk exceeds the 5\% budget.

The prediction lies within the interval of the measured gain in seven of the nine settings and above it for NQ-Open with Llama-3.1-8B and Gorilla with Mistral-7B. On questions where the top chunk is not the target but one of the three moved chunks is, the gain is 0.43 to 0.62. Moving more chunks helps even at 5\%, although two moved chunks then fit entirely within the budget on at most 43\% of the questions and three on at most 4\%. Moving two or three chunks beats moving one everywhere, significantly in four and five settings, respectively.

Against the other methods, the three moved chunks recomputed at 10\% are significantly better than InfoFlow-style reordering at its default 15\% in three settings and never significantly worse. At 5\%, they beat the token-level selector on the same reordered prompt significantly in six and never lose to it significantly, whereas the single moved chunk does so in four. At 10\%, the single moved chunk beats the selector on its own reordered prompt significantly in two settings but loses significantly in four (\textminus{}0.21 to +0.13). The two losses from recomputing the three chunks at 10\% with Llama-3.1-8B remain on the questions whose target is among the three chunks and where all three fit within the budget: \textminus{}0.18 [\textminus{}0.28, \textminus{}0.08] on NQ-Open and \textminus{}0.19 [\textminus{}0.33, \textminus{}0.04] on Gorilla. This is consistent with Table~\ref{tab:alloc}, where recomputing the known target at 10\% does not beat the selector with Llama-3.1-8B on these two datasets.

\begin{table*}[t]
\centering
\footnotesize
\setlength{\tabcolsep}{2.5pt}
\begin{tabular}{llcccccccc}
\toprule
 & & & \multicolumn{2}{c}{Hit top-1/top-3 (\%)} & \multicolumn{2}{c}{\begin{tabular}[b]{@{}c@{}}Top-1 moved,\\retention at 10\%\end{tabular}} & \multicolumn{2}{c}{\begin{tabular}[b]{@{}c@{}}Top-3 $-$ top-1\\moved at 10\%\end{tabular}} & \\ \cmidrule(lr){4-5}\cmidrule(lr){6-7}\cmidrule(lr){8-9}
Dataset & Model & $N$ & BGE-M3 & BM25 & Predicted & Measured & Predicted & Measured & \begin{tabular}[b]{@{}c@{}}Top-1 moved,\\BM25 $-$ BGE\end{tabular} \\
\midrule
NQ-Open & Qwen3-8B & 146 & 55/75 & 23/47 & 0.31 & \begin{tabular}[t]{@{}c@{}}0.32\\[-1pt]{\scriptsize [+0.25, +0.40]}\end{tabular} & +0.19 & \begin{tabular}[t]{@{}c@{}}\textbf{+0.18}\\[-1pt]{\scriptsize [+0.12, +0.26]}\end{tabular} & \begin{tabular}[t]{@{}c@{}}\textbf{\textminus{}0.24}\\[-1pt]{\scriptsize [\textminus{}0.33, \textminus{}0.15]}\end{tabular} \\
 & Llama-3.1-8B & 128 & 47/70 & 23/46 & 0.29 & \begin{tabular}[t]{@{}c@{}}0.28\\[-1pt]{\scriptsize [+0.20, +0.35]}\end{tabular} & +0.16 & \begin{tabular}[t]{@{}c@{}}\textbf{+0.19}\\[-1pt]{\scriptsize [+0.11, +0.27]}\end{tabular} & \begin{tabular}[t]{@{}c@{}}\textbf{\textminus{}0.18}\\[-1pt]{\scriptsize [\textminus{}0.27, \textminus{}0.09]}\end{tabular} \\
 & Mistral-7B & 101 & 50/71 & 25/41 & 0.35 & \begin{tabular}[t]{@{}c@{}}0.38\\[-1pt]{\scriptsize [+0.28, +0.48]}\end{tabular} & +0.12 & \begin{tabular}[t]{@{}c@{}}+0.07\\[-1pt]{\scriptsize [\textminus{}0.02, +0.15]}\end{tabular} & \begin{tabular}[t]{@{}c@{}}\textbf{\textminus{}0.17}\\[-1pt]{\scriptsize [\textminus{}0.28, \textminus{}0.05]}\end{tabular} \\
\midrule
Gorilla & Qwen3-8B & 89 & 53/74 & 54/72 & 0.27 & \begin{tabular}[t]{@{}c@{}}0.31\\[-1pt]{\scriptsize [+0.22, +0.42]}\end{tabular} & +0.08 & \begin{tabular}[t]{@{}c@{}}\textbf{+0.11}\\[-1pt]{\scriptsize [+0.01, +0.21]}\end{tabular} & \begin{tabular}[t]{@{}c@{}}+0.03\\[-1pt]{\scriptsize [\textminus{}0.06, +0.12]}\end{tabular} \\
 & Llama-3.1-8B & 100 & 53/69 & 52/67 & 0.27 & \begin{tabular}[t]{@{}c@{}}0.28\\[-1pt]{\scriptsize [+0.19, +0.37]}\end{tabular} & +0.06 & \begin{tabular}[t]{@{}c@{}}+0.07\\[-1pt]{\scriptsize [\textminus{}0.02, +0.16]}\end{tabular} & \begin{tabular}[t]{@{}c@{}}+0.01\\[-1pt]{\scriptsize [\textminus{}0.06, +0.09]}\end{tabular} \\
 & Mistral-7B & 108 & 47/72 & 49/72 & 0.44 & \begin{tabular}[t]{@{}c@{}}0.46\\[-1pt]{\scriptsize [+0.37, +0.56]}\end{tabular} & +0.21 & \begin{tabular}[t]{@{}c@{}}+0.06\\[-1pt]{\scriptsize [\textminus{}0.01, +0.15]}\end{tabular} & \begin{tabular}[t]{@{}c@{}}+0.04\\[-1pt]{\scriptsize [\textminus{}0.06, +0.12]}\end{tabular} \\
\midrule
TriviaQA & Qwen3-8B & 151 & 64/85 & 65/83 & 0.61 & \begin{tabular}[t]{@{}c@{}}0.65\\[-1pt]{\scriptsize [+0.57, +0.72]}\end{tabular} & +0.12 & \begin{tabular}[t]{@{}c@{}}\textbf{+0.11}\\[-1pt]{\scriptsize [+0.04, +0.17]}\end{tabular} & \begin{tabular}[t]{@{}c@{}}+0.05\\[-1pt]{\scriptsize [\textminus{}0.02, +0.11]}\end{tabular} \\
 & Llama-3.1-8B & 127 & 60/79 & 60/76 & 0.60 & \begin{tabular}[t]{@{}c@{}}0.59\\[-1pt]{\scriptsize [+0.50, +0.67]}\end{tabular} & +0.12 & \begin{tabular}[t]{@{}c@{}}+0.08\\[-1pt]{\scriptsize [\textminus{}0.01, +0.16]}\end{tabular} & \begin{tabular}[t]{@{}c@{}}\textminus{}0.01\\[-1pt]{\scriptsize [\textminus{}0.09, +0.08]}\end{tabular} \\
 & Mistral-7B & 117 & 62/85 & 67/83 & 0.65 & \begin{tabular}[t]{@{}c@{}}0.67\\[-1pt]{\scriptsize [+0.59, +0.75]}\end{tabular} & +0.11 & \begin{tabular}[t]{@{}c@{}}\textbf{+0.08}\\[-1pt]{\scriptsize [+0.03, +0.13]}\end{tabular} & \begin{tabular}[t]{@{}c@{}}+0.05\\[-1pt]{\scriptsize [\textminus{}0.03, +0.14]}\end{tabular} \\
\bottomrule
\end{tabular}
\caption{Moving the top chunks of a second retriever, BM25. Predicted values use the BM25 hit rates $h_1$ (top-1) and $h_3$ (top-3) and the retention of the moved BGE-M3 chunk at 5\% on the questions where it is and is not the target ($R_{\mathrm{hit}}$ and $R_{\mathrm{miss}}$, listed in Table~\ref{tab:topk}): top-1 retention $h_1R_{\mathrm{hit}}+(1-h_1)R_{\mathrm{miss}}$, and the gain of moving three chunks at 10\%, $(h_3-h_1)(R_{\mathrm{hit}}-R_{\mathrm{miss}})$. Measured values: the retention of the moved BM25 top chunk when all its tokens are recomputed at the 10\% budget, and the paired gain of the three moved chunks at 10\%, with 95\% bootstrap intervals. Top-1 moved, BM25 $-$ BGE: the paired difference in retention between the moved BM25 and BGE-M3 top chunks. Bold differences have intervals that exclude zero.}
\label{tab:topkbm25}
\end{table*}

\begin{figure*}[t]
\centering
\includegraphics[width=\textwidth]{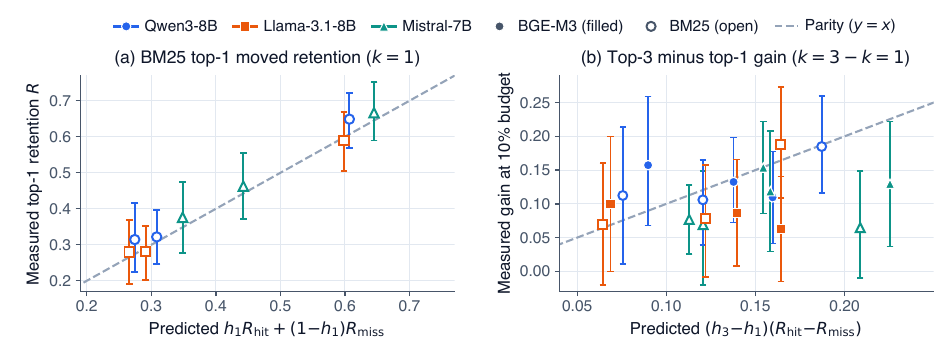}
\caption{Predicted and measured effects of moving a retriever's top chunks last at the 10\% budget, on NQ-Open, Gorilla and TriviaQA. (a) Retention of the moved BM25 top chunk against $h_1R_{\mathrm{hit}}+(1-h_1)R_{\mathrm{miss}}$, where $h_1$ is the BM25 top-1 hit rate and $R_{\mathrm{hit}}$ and $R_{\mathrm{miss}}$ are the BGE-M3 values from Table~\ref{tab:reorder}. (b) Gain of moving the top three chunks over the top one against $(h_3-h_1)(R_{\mathrm{hit}}-R_{\mathrm{miss}})$, where $h_3$ is the top-3 hit rate, for BGE-M3 (filled; Table~\ref{tab:topk}) and BM25 (open; Table~\ref{tab:topkbm25}). Bars are 95\% paired bootstrap intervals, and the dashed line is $y=x$.}
\label{fig:topkpred}
\end{figure*}

\paragraph{A second retriever.} Table~\ref{tab:topkbm25} repeats these experiments with BM25 \citep{robertson2009bm25} in place of BGE-M3. We use Okapi BM25 with $k_1=1.5$ and $b=0.75$ over the chunk texts without their position headers, take document frequencies from all chunks of the dataset, and break ties by position. Prompts, rules and budgets are unchanged. On NQ-Open, BM25 ranks the target first on 23--25\% of \Qs{}, against 47--55\% for BGE-M3; on Gorilla and TriviaQA, the two hit rates are close. From the BM25 hit rates and the $R_{\mathrm{hit}}$ and $R_{\mathrm{miss}}$ of BGE-M3, we predict the retention of the moved BM25 chunk, $h_1R_{\mathrm{hit}}+(1-h_1)R_{\mathrm{miss}}$, and the gain of moving three chunks at 10\%, as in Table~\ref{tab:topk}. The first prediction lies within the interval of the measured retention in every setting, and the second lies within the interval of the measured gain in all but one, Gorilla with Mistral-7B, where the prediction for BGE-M3 is also above the measured gain. On NQ-Open, where the hit rate halves, the moved BM25 chunk changes retention by \textminus{}0.24 to \textminus{}0.17 relative to the moved BGE-M3 chunk, significantly with all three models, as the lower hit rate predicts. In none of the nine settings is the moved BM25 chunk significantly better than the moved BGE-M3 chunk. The gain of moving a retriever's chunk thus follows the retriever's hit rate rather than a property of BGE-M3. Figure~\ref{fig:topkpred} plots both predictions against the measurements.

\paragraph{Time to first token.} Table~\ref{tab:timing} reports the time to first token of each condition with Qwen3-8B in our engine, where every condition runs through the same attention kernels. Reuse reaches the first token in 4--6\% of the full-prefill time, and the repairs do so in 9--24\%. At 5\%, Retrieved+4 takes a median of 6--9\,ms per question longer than the token-level selector, which is the cost of encoding the question with BGE-M3 and ranking the chunks. Recomputing the retrieved chunk alone is 11--57\,ms faster than the token-level selector. On open-domain QA and API documentation, the chunk usually holds fewer tokens than the budget and needs no attention scoring, whereas on LoCoMo the session usually fills the budget and its tokens are still ranked.

\paragraph{A fixed token-level rule.} The token-level baseline uses the rule chosen on development questions for each setting. Table~\ref{tab:uniform} repeats the 5\% allocation with ProphetKV as the rule in the five settings where another rule was chosen: Gorilla with all three models, TriviaQA with Mistral-7B and LoCoMo with Mistral-7B. Recomputing the target alone beats the ProphetKV baseline in all five, by 0.11 to 0.46, so it beats the ProphetKV baseline on every open-domain QA and API-documentation setting. On LoCoMo with Mistral-7B, by contrast, the target alone does not beat the setting's own rule (Table~\ref{tab:alloc}). There, ProphetKV has a retention of 0.69, against 0.78 for question attention, so the result on LoCoMo depends on which token-level rule serves as the baseline.

\begin{table*}[t]
\centering
\scriptsize
\setlength{\tabcolsep}{2.5pt}
\begin{tabular}{llccccccccc}
\toprule
 & & & \multicolumn{3}{c}{Loss and two terms} & \multicolumn{2}{c}{5\%, original order} & \multicolumn{3}{c}{Retrieved chunk moved last} \\ \cmidrule(lr){4-6}\cmidrule(lr){7-8}\cmidrule(lr){9-11}
Dataset & Model & $N$ & Loss & Same & Sim. & \begin{tabular}[b]{@{}c@{}}Target\\$-$ Token\end{tabular} & \begin{tabular}[b]{@{}c@{}}Retrieved\\$-$ Token\end{tabular} & \begin{tabular}[b]{@{}c@{}}Chunk moved\\$-$ Token\end{tabular} & \begin{tabular}[b]{@{}c@{}}Token moved\\$-$ Token\end{tabular} & \begin{tabular}[b]{@{}c@{}}Chunk moved $-$\\Token moved\end{tabular} \\
\midrule
LoCoMo & 14B & 74 & \begin{tabular}[t]{@{}c@{}}\textbf{+0.36}\\[-1pt]{\scriptsize [+0.21, +0.51]}\end{tabular} & \begin{tabular}[t]{@{}c@{}}\textbf{+0.20}\\[-1pt]{\scriptsize [+0.08, +0.31]}\end{tabular} & \begin{tabular}[t]{@{}c@{}}+0.03\\[-1pt]{\scriptsize [\textminus{}0.07, +0.12]}\end{tabular} & \begin{tabular}[t]{@{}c@{}}\textbf{+0.14}\\[-1pt]{\scriptsize [+0.05, +0.22]}\end{tabular} & \begin{tabular}[t]{@{}c@{}}+0.05\\[-1pt]{\scriptsize [\textminus{}0.05, +0.17]}\end{tabular} & -- & -- & -- \\
 & 32B & 67 & \begin{tabular}[t]{@{}c@{}}\textbf{+0.46}\\[-1pt]{\scriptsize [+0.35, +0.54]}\end{tabular} & \begin{tabular}[t]{@{}c@{}}\textbf{+0.16}\\[-1pt]{\scriptsize [+0.02, +0.30]}\end{tabular} & \begin{tabular}[t]{@{}c@{}}\textbf{+0.18}\\[-1pt]{\scriptsize [+0.05, +0.31]}\end{tabular} & \begin{tabular}[t]{@{}c@{}}+0.07\\[-1pt]{\scriptsize [\textminus{}0.01, +0.16]}\end{tabular} & \begin{tabular}[t]{@{}c@{}}+0.00\\[-1pt]{\scriptsize [\textminus{}0.04, +0.04]}\end{tabular} & -- & -- & -- \\
 & 8B & 105 & \textbf{+0.47} & \textbf{+0.19} & \textbf{+0.16} & \textminus{}0.04 & \textbf{\textminus{}0.10} & -- & -- & -- \\
\midrule
NQ-Open & 14B & 150 & \begin{tabular}[t]{@{}c@{}}\textbf{+0.79}\\[-1pt]{\scriptsize [+0.72, +0.85]}\end{tabular} & \begin{tabular}[t]{@{}c@{}}\textbf{+0.17}\\[-1pt]{\scriptsize [+0.09, +0.25]}\end{tabular} & \begin{tabular}[t]{@{}c@{}}+0.08\\[-1pt]{\scriptsize [\textminus{}0.01, +0.17]}\end{tabular} & \begin{tabular}[t]{@{}c@{}}\textbf{+0.22}\\[-1pt]{\scriptsize [+0.13, +0.30]}\end{tabular} & \begin{tabular}[t]{@{}c@{}}\textbf{+0.13}\\[-1pt]{\scriptsize [+0.05, +0.21]}\end{tabular} & \begin{tabular}[t]{@{}c@{}}\textbf{+0.34}\\[-1pt]{\scriptsize [+0.24, +0.43]}\end{tabular} & \begin{tabular}[t]{@{}c@{}}\textbf{+0.24}\\[-1pt]{\scriptsize [+0.16, +0.32]}\end{tabular} & \begin{tabular}[t]{@{}c@{}}\textbf{+0.10}\\[-1pt]{\scriptsize [+0.03, +0.17]}\end{tabular} \\
 & 32B & 138 & \begin{tabular}[t]{@{}c@{}}\textbf{+0.78}\\[-1pt]{\scriptsize [+0.71, +0.85]}\end{tabular} & \begin{tabular}[t]{@{}c@{}}\textbf{+0.37}\\[-1pt]{\scriptsize [+0.28, +0.46]}\end{tabular} & \begin{tabular}[t]{@{}c@{}}+0.08\\[-1pt]{\scriptsize [\textminus{}0.01, +0.18]}\end{tabular} & \begin{tabular}[t]{@{}c@{}}\textbf{+0.38}\\[-1pt]{\scriptsize [+0.28, +0.46]}\end{tabular} & \begin{tabular}[t]{@{}c@{}}\textbf{+0.23}\\[-1pt]{\scriptsize [+0.14, +0.31]}\end{tabular} & \begin{tabular}[t]{@{}c@{}}\textbf{+0.35}\\[-1pt]{\scriptsize [+0.25, +0.44]}\end{tabular} & \begin{tabular}[t]{@{}c@{}}\textbf{+0.25}\\[-1pt]{\scriptsize [+0.17, +0.33]}\end{tabular} & \begin{tabular}[t]{@{}c@{}}\textbf{+0.09}\\[-1pt]{\scriptsize [+0.01, +0.17]}\end{tabular} \\
 & 8B & 146 & \textbf{+0.85} & \textbf{+0.23} & \textbf{+0.19} & \textbf{+0.35} & \textbf{+0.22} & \textbf{+0.40} & \textbf{+0.11} & \textbf{+0.29} \\
\midrule
Gorilla & 14B & 111 & \begin{tabular}[t]{@{}c@{}}\textbf{+0.90}\\[-1pt]{\scriptsize [+0.85, +0.95]}\end{tabular} & \begin{tabular}[t]{@{}c@{}}\textbf{+0.58}\\[-1pt]{\scriptsize [+0.49, +0.68]}\end{tabular} & \begin{tabular}[t]{@{}c@{}}\textminus{}0.10$^\dagger$\\[-1pt]{\scriptsize [\textminus{}0.15, \textminus{}0.05]}\end{tabular} & \begin{tabular}[t]{@{}c@{}}\textbf{+0.45}\\[-1pt]{\scriptsize [+0.36, +0.54]}\end{tabular} & \begin{tabular}[t]{@{}c@{}}\textbf{+0.28}\\[-1pt]{\scriptsize [+0.18, +0.38]}\end{tabular} & \begin{tabular}[t]{@{}c@{}}\textbf{+0.38}\\[-1pt]{\scriptsize [+0.28, +0.49]}\end{tabular} & \begin{tabular}[t]{@{}c@{}}\textbf{+0.17}\\[-1pt]{\scriptsize [+0.09, +0.25]}\end{tabular} & \begin{tabular}[t]{@{}c@{}}\textbf{+0.21}\\[-1pt]{\scriptsize [+0.13, +0.30]}\end{tabular} \\
 & 32B & 122 & \begin{tabular}[t]{@{}c@{}}\textbf{+0.93}\\[-1pt]{\scriptsize [+0.88, +0.97]}\end{tabular} & \begin{tabular}[t]{@{}c@{}}\textbf{+0.90}\\[-1pt]{\scriptsize [+0.84, +0.95]}\end{tabular} & \begin{tabular}[t]{@{}c@{}}\textminus{}0.07$^\dagger$\\[-1pt]{\scriptsize [\textminus{}0.12, \textminus{}0.03]}\end{tabular} & \begin{tabular}[t]{@{}c@{}}\textbf{+0.43}\\[-1pt]{\scriptsize [+0.34, +0.52]}\end{tabular} & \begin{tabular}[t]{@{}c@{}}\textbf{+0.21}\\[-1pt]{\scriptsize [+0.11, +0.30]}\end{tabular} & \begin{tabular}[t]{@{}c@{}}\textbf{+0.30}\\[-1pt]{\scriptsize [+0.19, +0.40]}\end{tabular} & \begin{tabular}[t]{@{}c@{}}\textbf{+0.15}\\[-1pt]{\scriptsize [+0.05, +0.25]}\end{tabular} & \begin{tabular}[t]{@{}c@{}}\textbf{+0.15}\\[-1pt]{\scriptsize [+0.08, +0.22]}\end{tabular} \\
 & 8B & 89 & \textbf{+1.00} & \textbf{+0.82} & +0.00$^\dagger$ & \textbf{+0.17} & +0.09 & \textbf{+0.21} & +0.00 & \textbf{+0.21} \\
\midrule
TriviaQA & 14B & 157 & \begin{tabular}[t]{@{}c@{}}\textbf{+0.77}\\[-1pt]{\scriptsize [+0.71, +0.83]}\end{tabular} & \begin{tabular}[t]{@{}c@{}}\textbf{+0.32}\\[-1pt]{\scriptsize [+0.24, +0.40]}\end{tabular} & \begin{tabular}[t]{@{}c@{}}\textbf{+0.20}\\[-1pt]{\scriptsize [+0.11, +0.29]}\end{tabular} & \begin{tabular}[t]{@{}c@{}}\textbf{+0.20}\\[-1pt]{\scriptsize [+0.13, +0.28]}\end{tabular} & \begin{tabular}[t]{@{}c@{}}\textbf{+0.10}\\[-1pt]{\scriptsize [+0.03, +0.18]}\end{tabular} & \begin{tabular}[t]{@{}c@{}}\textbf{+0.24}\\[-1pt]{\scriptsize [+0.14, +0.34]}\end{tabular} & \begin{tabular}[t]{@{}c@{}}\textbf{+0.22}\\[-1pt]{\scriptsize [+0.15, +0.30]}\end{tabular} & \begin{tabular}[t]{@{}c@{}}+0.01\\[-1pt]{\scriptsize [\textminus{}0.05, +0.08]}\end{tabular} \\
 & 32B & 123 & \begin{tabular}[t]{@{}c@{}}\textbf{+0.72}\\[-1pt]{\scriptsize [+0.64, +0.80]}\end{tabular} & \begin{tabular}[t]{@{}c@{}}\textbf{+0.24}\\[-1pt]{\scriptsize [+0.15, +0.33]}\end{tabular} & \begin{tabular}[t]{@{}c@{}}\textbf{+0.25}\\[-1pt]{\scriptsize [+0.16, +0.34]}\end{tabular} & \begin{tabular}[t]{@{}c@{}}\textbf{+0.25}\\[-1pt]{\scriptsize [+0.15, +0.36]}\end{tabular} & \begin{tabular}[t]{@{}c@{}}\textbf{+0.15}\\[-1pt]{\scriptsize [+0.05, +0.25]}\end{tabular} & \begin{tabular}[t]{@{}c@{}}\textbf{+0.22}\\[-1pt]{\scriptsize [+0.11, +0.32]}\end{tabular} & \begin{tabular}[t]{@{}c@{}}\textbf{+0.24}\\[-1pt]{\scriptsize [+0.16, +0.33]}\end{tabular} & \begin{tabular}[t]{@{}c@{}}\textminus{}0.02\\[-1pt]{\scriptsize [\textminus{}0.10, +0.06]}\end{tabular} \\
 & 8B & 151 & \textbf{+0.73} & \textbf{+0.25} & \textbf{+0.17} & \textbf{+0.20} & \textbf{+0.14} & \textbf{+0.28} & \textbf{+0.19} & \textbf{+0.09} \\
\bottomrule
\end{tabular}
\caption{Qwen3-14B (bf16, with its layers split over two RTX 4090 D) and Qwen3-32B (bf16, on one RTX PRO 6000) on the test sets, with the second question sets for NQ-Open and Gorilla, compared with Qwen3-8B. Loss, Same and Sim.: the loss and the same-category and similarity terms, as in Table~\ref{tab:main_decomposition}; allocation columns as in Table~\ref{tab:alloc}; moved-chunk columns as in Tables~\ref{tab:reorder} and~\ref{tab:reorder_tok}. The token-level rule is ProphetKV in every Qwen3-14B and Qwen3-32B setting; for Qwen3-8B it is the development-selected rule, which differs from ProphetKV only on Gorilla. The five replication criteria cover the loss, the same-category and similarity terms, Target $-$ Token and Chunk moved $-$ Token; the last two columns are reported in addition. Retrieved+4 was also evaluated but is not shown; it beats Token by +0.12 to +0.27 on open-domain QA and API documentation. Qwen3-14B and Qwen3-32B: paired differences with 95\% bootstrap intervals; Qwen3-8B: point estimates. Bold: the interval excludes zero. $^\dagger$Retention under RAND is zero, so the term cannot be positive; these intervals are not set in bold.}
\label{tab:larger}
\end{table*}

\paragraph{Larger models.}
\label{app:larger}
We repeated the main conditions with Qwen3-14B and Qwen3-32B (Table~\ref{tab:larger}). These experiments cover the four test sets, with the second question sets of NQ-Open and Gorilla added by the rule of Appendix~\ref{app:data}, which gives 67--157 \Qs{} questions per dataset and model. They include the three distractor sources and, on \Qs{}, the token-level selector, Target, Retrieved and Retrieved+4 at 5\% with ProphetKV as the rule. On open-domain QA and API documentation, they also include the two moved-chunk conditions of Tables~\ref{tab:reorder} and~\ref{tab:reorder_tok}.

For Qwen3-14B, the 40 layers are split over two GPUs, and each layer's cache is kept on the GPU that holds that layer. When Qwen3-8B is split in the same way, it reproduces all 260 outputs of its single-GPU runs on 20 NQ-Open questions. Qwen3-32B runs on one RTX PRO 6000. Run on that GPU, Qwen3-8B gives the same score as on the RTX 4090 for 247 of the 249 outputs that both runs produce on 20 NQ-Open questions, but the same text for only 210 of them. Greedy decoding turns small numerical differences between the two GPU architectures into different wordings. Each model runs all its conditions on one kind of GPU, so these differences do not favour any condition.

Both models reproduce all five results that we test for replication. The loss is significant on all four datasets (0.36--0.93), and the same-category term is significantly positive on all four. The similarity term is positive on both open-domain QA datasets but significant only on TriviaQA. On NQ-Open, it is +0.08 [\textminus{}0.01, +0.17] with Qwen3-14B and +0.08 [\textminus{}0.01, +0.18] with Qwen3-32B. On Gorilla, retention under RAND is again zero. Recomputing the target beats the token-level selector at 5\% on all three open-domain QA and API-documentation datasets (0.20 to 0.45), and so does recomputing the retrieved chunk after moving it last (0.22 to 0.38). Beyond these five results, the move also raises the retention of the token-level selector on all three datasets (0.15 to 0.25). On the reordered prompt, recomputing the moved chunk instead of selected tokens is significantly better on NQ-Open and Gorilla but not on TriviaQA, where the differences are +0.01 with Qwen3-14B and \textminus{}0.02 with Qwen3-32B.

On LoCoMo the two models differ. Recomputing the target session beats the selector by +0.14 [+0.05, +0.22] with Qwen3-14B, and by +0.07 [\textminus{}0.01, +0.16], not significantly, with Qwen3-32B. With the same rule, Qwen3-8B shows no gain (\textminus{}0.04), whereas Mistral-7B does (+0.11, Table~\ref{tab:uniform}). The whole session fits within the budget on 22\% and 21\% of these questions, about as often as for the smaller models, and the bootstrap intervals resample only eight conversations. Recomputing the retrieved session gains +0.05 [\textminus{}0.05, +0.17] with Qwen3-14B and +0.00 [\textminus{}0.04, +0.04] with Qwen3-32B.

\begin{table*}[t]
\centering
\footnotesize
\setlength{\tabcolsep}{2pt}
\begin{tabular}{llcccccccccccc}
\toprule
 & & & \multicolumn{5}{c}{Retention} & \multicolumn{2}{c}{KV dist.} & \multicolumn{4}{c}{Attention on target} \\ \cmidrule(lr){4-8}\cmidrule(lr){9-10}\cmidrule(lr){11-14}
Dataset & Model & $N$ & None & \begin{tabular}[b]{@{}c@{}}Target\\KV\end{tabular} & Recomp. & \begin{tabular}[b]{@{}c@{}}Recomp.\\on full\end{tabular} & \begin{tabular}[b]{@{}c@{}}Recomp. $-$\\Target KV\end{tabular} & None & Recomp. & None & \begin{tabular}[b]{@{}c@{}}Target\\KV\end{tabular} & Rec. & Full \\
\midrule
NQ-Open & Qwen3-8B & 40 & 0.15 & 0.15 & 0.53 & 0.97 & \begin{tabular}[t]{@{}c@{}}\textbf{+0.38}\\[-1pt]{\scriptsize [+0.23, +0.53]}\end{tabular} & 0.36 & 0.26 & 0.035 & 0.021 & 0.050 & 0.060 \\
NQ-Open & Llama-3.1-8B & 30 & 0.20 & 0.20 & 0.33 & 1.00 & \begin{tabular}[t]{@{}c@{}}\textbf{+0.13}\\[-1pt]{\scriptsize [+0.03, +0.27]}\end{tabular} & 0.40 & 0.28 & 0.029 & 0.020 & 0.025 & 0.064 \\
NQ-Open & Mistral-7B & 40 & 0.23 & 0.42 & 0.62 & 0.97 & \begin{tabular}[t]{@{}c@{}}\textbf{+0.20}\\[-1pt]{\scriptsize [+0.05, +0.35]}\end{tabular} & 0.40 & 0.27 & 0.031 & 0.019 & 0.030 & 0.059 \\
Gorilla & Qwen3-8B & 40 & 0.00 & 0.05 & 0.25 & 0.95 & \begin{tabular}[t]{@{}c@{}}\textbf{+0.20}\\[-1pt]{\scriptsize [+0.10, +0.33]}\end{tabular} & 0.42 & 0.30 & 0.040 & 0.027 & 0.065 & 0.053 \\
Gorilla & Mistral-7B & 40 & 0.05 & 0.38 & 0.55 & 1.00 & \begin{tabular}[t]{@{}c@{}}\textbf{+0.17}\\[-1pt]{\scriptsize [+0.03, +0.33]}\end{tabular} & 0.45 & 0.29 & 0.044 & 0.034 & 0.045 & 0.068 \\
TriviaQA & Qwen3-8B & 40 & 0.35 & 0.30 & 0.50 & 1.00 & \begin{tabular}[t]{@{}c@{}}\textbf{+0.20}\\[-1pt]{\scriptsize [+0.07, +0.33]}\end{tabular} & 0.34 & 0.23 & 0.040 & 0.026 & 0.061 & 0.076 \\
TriviaQA & Mistral-7B & 40 & 0.25 & 0.50 & 0.65 & 0.97 & \begin{tabular}[t]{@{}c@{}}\textbf{+0.15}\\[-1pt]{\scriptsize [+0.03, +0.28]}\end{tabular} & 0.39 & 0.25 & 0.048 & 0.040 & 0.058 & 0.092 \\
\bottomrule
\end{tabular}
\caption{Repairing the target chunk in the reused SIM cache, for the first 40 \Qs{} questions (30 for Llama-3.1-8B) in identifier order whose target is not in the first slot. Retention with the target unrepaired (None), with its full-prefill KV inserted (Target KV), recomputed against the reused cache (Recomp.), and recomputed against the full-prefill cache (Recomp.\ on full; by causality it equals full prefill up to numerical differences). KV dist.: median relative $L_2$ distance of the target KV from its full-prefill value. Attention: the share of the question\textquoteright s attention mass on the target chunk, aggregated as for the detector $d_1$ (Table~\ref{tab:detect}), for the three caches (None, Target KV, and Rec.\ for Recomp.) and for full prefill (Full).}
\label{tab:compat}
\end{table*}

\begin{table*}[t]
\centering
\footnotesize
\setlength{\tabcolsep}{2.5pt}
\begin{tabular}{llccccccccc}
\toprule
 & & & \multicolumn{2}{c}{Earlier chunks reused} & \multicolumn{2}{c}{Earlier chunks full} & & & & \\ \cmidrule(lr){4-5}\cmidrule(lr){6-7}
Dataset & Model & $N$ & \begin{tabular}[b]{@{}c@{}}Target\\reused\end{tabular} & \begin{tabular}[b]{@{}c@{}}Target\\KV\end{tabular} & \begin{tabular}[b]{@{}c@{}}Target\\reused\end{tabular} & \begin{tabular}[b]{@{}c@{}}Target\\KV\end{tabular} & \begin{tabular}[b]{@{}c@{}}Target\\recomp.\end{tabular} & \begin{tabular}[b]{@{}c@{}}Full target\\effect (reused)\end{tabular} & \begin{tabular}[b]{@{}c@{}}Full target\\effect (full)\end{tabular} & Interaction \\
\midrule
NQ-Open & Qwen3-8B & 40 & 0.15 & 0.15 & 0.28 & 0.55 & 0.53 & \begin{tabular}[t]{@{}c@{}}+0.00\\[-1pt]{\scriptsize [\textminus{}0.12, +0.12]}\end{tabular} & \begin{tabular}[t]{@{}c@{}}\textbf{+0.28}\\[-1pt]{\scriptsize [+0.12, +0.42]}\end{tabular} & \begin{tabular}[t]{@{}c@{}}\textbf{+0.28}\\[-1pt]{\scriptsize [+0.15, +0.40]}\end{tabular} \\
NQ-Open & Llama-3.1-8B & 30 & 0.20 & 0.20 & 0.27 & 0.43 & 0.33 & \begin{tabular}[t]{@{}c@{}}+0.00\\[-1pt]{\scriptsize [\textminus{}0.17, +0.17]}\end{tabular} & \begin{tabular}[t]{@{}c@{}}+0.17\\[-1pt]{\scriptsize [+0.00, +0.33]}\end{tabular} & \begin{tabular}[t]{@{}c@{}}+0.17\\[-1pt]{\scriptsize [\textminus{}0.03, +0.37]}\end{tabular} \\
NQ-Open & Mistral-7B & 40 & 0.23 & 0.42 & 0.33 & 0.55 & 0.62 & \begin{tabular}[t]{@{}c@{}}\textbf{+0.20}\\[-1pt]{\scriptsize [+0.03, +0.40]}\end{tabular} & \begin{tabular}[t]{@{}c@{}}\textbf{+0.23}\\[-1pt]{\scriptsize [+0.10, +0.35]}\end{tabular} & \begin{tabular}[t]{@{}c@{}}+0.03\\[-1pt]{\scriptsize [\textminus{}0.12, +0.17]}\end{tabular} \\
Gorilla & Qwen3-8B & 40 & 0.00 & 0.05 & 0.15 & 0.45 & 0.25 & \begin{tabular}[t]{@{}c@{}}+0.05\\[-1pt]{\scriptsize [+0.00, +0.12]}\end{tabular} & \begin{tabular}[t]{@{}c@{}}\textbf{+0.30}\\[-1pt]{\scriptsize [+0.17, +0.45]}\end{tabular} & \begin{tabular}[t]{@{}c@{}}\textbf{+0.25}\\[-1pt]{\scriptsize [+0.12, +0.38]}\end{tabular} \\
Gorilla & Mistral-7B & 40 & 0.05 & 0.38 & 0.07 & 0.60 & 0.55 & \begin{tabular}[t]{@{}c@{}}\textbf{+0.33}\\[-1pt]{\scriptsize [+0.17, +0.47]}\end{tabular} & \begin{tabular}[t]{@{}c@{}}\textbf{+0.53}\\[-1pt]{\scriptsize [+0.38, +0.68]}\end{tabular} & \begin{tabular}[t]{@{}c@{}}\textbf{+0.20}\\[-1pt]{\scriptsize [+0.03, +0.35]}\end{tabular} \\
TriviaQA & Qwen3-8B & 40 & 0.35 & 0.30 & 0.42 & 0.45 & 0.50 & \begin{tabular}[t]{@{}c@{}}\textminus{}0.05\\[-1pt]{\scriptsize [\textminus{}0.20, +0.10]}\end{tabular} & \begin{tabular}[t]{@{}c@{}}+0.03\\[-1pt]{\scriptsize [\textminus{}0.10, +0.15]}\end{tabular} & \begin{tabular}[t]{@{}c@{}}+0.07\\[-1pt]{\scriptsize [\textminus{}0.05, +0.20]}\end{tabular} \\
TriviaQA & Mistral-7B & 40 & 0.25 & 0.50 & 0.42 & 0.65 & 0.65 & \begin{tabular}[t]{@{}c@{}}\textbf{+0.25}\\[-1pt]{\scriptsize [+0.07, +0.40]}\end{tabular} & \begin{tabular}[t]{@{}c@{}}\textbf{+0.23}\\[-1pt]{\scriptsize [+0.07, +0.38]}\end{tabular} & \begin{tabular}[t]{@{}c@{}}\textminus{}0.03\\[-1pt]{\scriptsize [\textminus{}0.20, +0.17]}\end{tabular} \\
\bottomrule
\end{tabular}
\caption{Inserting the target\textquoteright s full-prefill KV with the chunks before it reused or restored to their full-prefill KV (the chunks after the target stay reused), on the questions of Table~\ref{tab:compat}. Earlier chunks full: the chunks before the target restored to their full-prefill KV. Target recomp.: the target recomputed against the reused cache. Full target effect: retention with the target\textquoteright s full-prefill KV inserted minus retention with the target reused, when the earlier chunks are reused (reused) or restored to their full-prefill KV (full). Effects are paired differences with 95\% bootstrap intervals; Interaction is the second effect minus the first.}
\label{tab:compat2}
\end{table*}

\paragraph{Full-prefill KV versus recomputation.}
\label{app:compat}
Inserting the target's full-prefill KV into the reused cache (Section~\ref{sec:replace}) and recomputing the target against the reused cache (Section~\ref{sec:alloc}) both restore the same chunk, yet the latter recovers more. Table~\ref{tab:compat} compares them on the same questions in seven settings. As a reference, recomputing the target against the full-prefill cache gives the full-prefill score on 265 of 270 questions and the same output text on 244, and its KV lies within a median relative distance of 0.02 of the full-prefill KV, an order of magnitude below the distances reported next. On these questions, the inserted full-prefill KV changes retention significantly in three of the seven settings, and on all of \Qs{} it raises retention significantly in six of the twelve settings (Section~\ref{sec:replace}). Recomputation, in contrast, gains 0.13--0.38 over the inserted KV, significantly in every setting, and it is better on 60 questions and worse on four. The recomputed target stays far from its full-prefill KV (median relative distance 0.23--0.30, against 0.34--0.45 for the unrepaired target), so closeness to the full-prefill KV does not explain the gain.

One explanation is that the recomputed target is encoded against the same reused chunks that the question sees when it reads the target, whereas the full-prefill KV of the target was encoded against different KV for those chunks. Table~\ref{tab:compat2} tests this explanation by also restoring the full-prefill KV of every chunk before the target, while the chunks after it stay reused. It uses the same questions, which include three Mistral-7B settings where the inserted KV helps on all of \Qs{}. On 270 \Qs{} questions in seven settings, inserting the target's full-prefill KV raises retention by +0.11 [+0.06, +0.17] when the earlier chunks stay reused and by +0.25 [+0.20, +0.31] when their full-prefill KV is restored too. With the earlier chunks reused, the inserted target KV raises retention significantly only in the three Mistral-7B settings, and with them restored it does so in five. The interaction is positive in six of the seven and significant in three, and pooled over all seven it is +0.14 [+0.08, +0.20].

The models differ in how much the inserted KV helps on its own. With Qwen3-8B and Llama-3.1-8B (four settings), it changes retention by \textminus{}0.05 to +0.05, and it gains more when the earlier chunks are also restored, in all four settings (significantly in two). With Mistral-7B, it helps on its own (+0.20 to +0.33), and restoring the earlier chunks adds a significant further gain only on Gorilla. Recomputing the target against the restored earlier chunks reproduces the inserted full-prefill target, as causality requires. It gives the same score on 269 of 270 questions, with a median relative KV distance of at most 0.02. Attention mass does not follow these effects. Restoring the earlier chunks raises the question's attention on the target from 0.03--0.05 to 0.04--0.08 of its mass, and adding the target's full-prefill KV lowers it again to 0.03--0.06 while retention rises.

\paragraph{Copied and recomputed states in other settings.} Two studies of stale caches find little difference between copying full-prefill states into a cache and recomputing them in place. After a document edit, \citet{contiguity} find that the two recover the answer margin to within 0.02 of each other when the repaired positions form a contiguous window starting at the edit, whereas copying recovers far more for the scattered positions where copying has the largest effect. SAFE-KV keeps 99.8\% of the full-prefill answers on HotpotQA with copied suffix states and 100.0\% with recomputed ones, at an operating point that re-encodes 94.8\% of the context \citep{safekv}. In both studies, the copied states are read next to states whose context is close to their own: the stale text differs from the edited text in a span of at most six tokens, and the copied suffix covers most of the prompt. In our setting, the target is read next to chunks that were each encoded alone and are not restored, and Table~\ref{tab:compat2} shows that the inserted KV gains more when the chunks before it are restored. Our explanation in terms of contextual compatibility is limited to this configuration, in which the target's full-prefill KV is inserted while the chunks before it stay reused.

\end{document}